\documentclass[10pt,twocolumn,letterpaper]{article}

\newif\ifcompactsubmission
\compactsubmissiontrue
\newif\ifincludeappendix
\includeappendixtrue          %
\newif\ifprintbibliography
\printbibliographytrue

\newif\ifwacvreview
\wacvreviewfalse              %
\newif\ifwacvpagenumbers
\wacvpagenumbersfalse         %
\newif\ifwacvalgorithmstrack
\wacvalgorithmstrackfalse
\newif\ifwacvapplicationstrack
\wacvapplicationstrackfalse
\newif\ifwacvdatasetstrack
\wacvdatasetstracktrue

\newcommand{\paperid}{2597} %
\newcommand{\papertitle}{CrowdCC: Jointly Benchmarking Corruption Robustness
and Deployment Quantization in Crowd Counting}
\newcommand{\paperauthors}{%
  S Mahmudul Hasan \qquad Lu Peng \\
  Tulane University \\
  {\tt\small \{shasan1, lpeng3\}@tulane.edu}%
}

\newif\ifincludeacknowledgments
\includeacknowledgmentstrue

\newcommand{\suite}{CrowdCC\xspace}
\newcommand{\suitelong}{Crowd Counting Corruptions\xspace}

\makeatletter
\def\input@path{{venues/wacv-2027/template/}}
\makeatother

\ifwacvreview
  \ifwacvalgorithmstrack
    \usepackage[review,algorithms]{wacv}
  \else\ifwacvapplicationstrack
    \usepackage[review,applications]{wacv}
  \else\ifwacvdatasetstrack
    \usepackage[review,datasets]{wacv}
  \else
    \PackageError{paper-config}{Select a WACV review track in config.tex}{}
  \fi\fi\fi
\else
  \ifwacvpagenumbers
    \usepackage[pagenumbers]{wacv}
  \else
    \usepackage{wacv}
  \fi
\fi

\usepackage{array}      %
\usepackage{booktabs}
\usepackage{multirow}
\usepackage{amsmath}
\usepackage{amssymb}
\usepackage{graphicx}
\usepackage{xspace}
\usepackage{siunitx}
\usepackage{tikz}
\usepackage{pgfplots}
\pgfplotsset{compat=1.18}
\usetikzlibrary{positioning,fit,backgrounds,arrows.meta,patterns}
\usepgfplotslibrary{groupplots}

\tikzset{every picture/.append style={scale=0.94,transform shape}}

\definecolor{worsecol}{HTML}{B3261E}
\definecolor{bettercol}{HTML}{1B6E3C}
\definecolor{fp32col}{HTML}{4C6EF5}
\definecolor{fp16col}{HTML}{12A594}
\definecolor{int8col}{HTML}{E8590C}
\definecolor{rulecol}{HTML}{888888}

\definecolor{dsA}{HTML}{2B6CB0}
\definecolor{dsB}{HTML}{D97706}
\definecolor{dsC}{HTML}{2F855A}
\definecolor{dsD}{HTML}{9B2C6F}

\definecolor{grpA}{HTML}{1F77B4}
\definecolor{grpB}{HTML}{D62728}
\definecolor{grpC}{HTML}{2CA02C}
\definecolor{grpD}{HTML}{9467BD}
\definecolor{grpE}{HTML}{8C564B}
\definecolor{grpF}{HTML}{E377C2}

\newcommand{\worse}[1]{\textcolor{worsecol}{#1}}
\newcommand{\better}[1]{\textcolor{bettercol}{#1}}

\newcommand{\drd}{\ensuremath{\Delta_{\mathrm{RD}}}\xspace}
\newcommand{\mcmae}{\textrm{mcMAE}\xspace}

\makeatletter
\renewcommand\paragraph{\@startsection{paragraph}{4}{\z@}%
  {1.5ex plus .5ex minus .2ex}%
  {-1em}%
  {\normalfont\normalsize\bfseries}}
\makeatother

\definecolor{wacvblue}{rgb}{0.21,0.49,0.74}
\usepackage[pagebackref,breaklinks,colorlinks,allcolors=wacvblue]{hyperref}

\usepackage[accsupp]{axessibility}  %

\def\wacvPaperID{\paperid}
\def\confName{WACV}
\def\confYear{2027}

\title{\papertitle}
\author{\paperauthors}

\begin{document}
\maketitle

\begin{abstract}

Crowd-counting models deployed on edge devices face two challenges: inputs are degraded along the capture and transmission pipeline, while models are often quantized to satisfy resource constraints of the edge. Yet corruption robustness and deployment quantization are typically evaluated separately. We introduce \suite, a controlled benchmark that jointly evaluates these effects across 23 deployment-motivated corruptions at three severity levels, four crowd-counting datasets, and 26 models. We evaluate TensorRT engines at FP32, FP16, and post-training INT8 precision on an A100 GPU, alongside throughput and energy measurements on corresponding Jetson Orin~NX configurations. The corruptions span six families, with head annotations adjusted when geometry or visibility changes. Our results show that capture-chain corruptions consistently cause the largest degradation across evaluated models, while crowd density affects absolute and relative errors differently. Accuracy on clean images does not reliably predict performance on the corresponding corrupted images. INT8 changes corruption robustness more than FP16, without a consistent direction, and does not consistently improve latency or energy efficiency. INT8 inference efficiency is strongly associated with reformat overhead in the compiled engine, with slowdowns also observed under limited INT8 coverage. Precision changes corrupted-error rankings and robustness--cost frontiers, highlighting the need to consider corruption robustness and deployment cost jointly when selecting models and precision. The code is available at \url{https://github.com/Tulane-Computer-Architecture-Systems/CrowdCC-Release}.

\end{abstract}

\section{Introduction}
\label{sec:intro}

Crowd counting supports applications such as video surveillance, traffic management, and monitoring of large public gatherings~\cite{zhang2015crossscene,xiong2017spatiotemporal,shi2019perspective}. Increasingly, these systems operate near cameras on resource-constrained hardware~\cite{khan2023lcdnet,gu2023npu,khan2025edgekd}. Deployment changes both the input and the model: images are degraded by weather, optics, exposure, sensor noise, compression, and transmission artifacts, while models are compiled and executed at reduced numerical precision. Evaluation on clean images at full precision therefore captures only part of deployed performance.

Robustness evaluation ~\cite{qi2026quantum} is further complicated by the spatial nature of crowd counting. Unlike standard classification corruption benchmarks, which generally retain the original labels, corruptions such as cropping, viewpoint changes, and occlusion can move or remove annotated heads, altering the valid counting target. Evaluating these images against unchanged annotations can therefore conflate model errors with annotation mismatches. Even with consistent annotations, counting errors can arise from both over- and underestimation and vary substantially with crowd size and dataset scale. A meaningful benchmark must therefore account for changes in the counting target and distinguish absolute corrupted error from degradation relative to the corresponding clean baseline.

Existing work has largely examined these issues separately. Corruption benchmarks are well established for classification and dense vision~\cite{hendrycks2019imagenetc,kamann2021segrobustness,kong2023robodepth,kar2022threedcc}, while crowd-counting studies have investigated adverse weather, domain shift, adversarial perturbations, and compression individually~\cite{sindagi2022jhucrowdpp,huang2023awccnet,wu2021advcrowd,peng2024single,bakhtiarnia2022compression}. Efficient crowd-counting models, meanwhile, are evaluated primarily in terms of clean accuracy, model complexity, and latency~\cite{wang2020mobilecount,khan2023lcdnet,khan2025edgekd}. Together, these studies leave open how corruption robustness, numerical precision, and deployment cost interact to influence model selection under compiled inference.

To address this gap, we introduce \suite{} (\suitelong), a controlled, deployment-oriented corruption benchmark comprising 23 corruptions at three calibrated severity levels across ShanghaiTech~A/B, UCF-QNRF, and JHU-Crowd++~\cite{zhang2016single,idrees2018qnrf,sindagi2022jhucrowdpp}. The corruptions span six families, with annotations updated when geometry or visibility changes. We evaluate 26 crowd-counting models at FP32, FP16, and post-training INT8 precision using TensorRT~\cite{tensorrt} inference engines, and measure throughput and energy on a Jetson Orin~NX.

Our evaluation reveals several deployment-relevant findings. Capture-chain corruptions cause the greatest degradation across architectures, while dense scenes exhibit larger absolute errors and sparse scenes greater relative degradation. Clean accuracy, corrupted error, and relative degradation produce different model rankings, and counting accuracy alone does not capture localization fidelity. FP16 largely preserves FP32 corruption response, whereas INT8 introduces larger, model-dependent changes without a consistent trend. INT8 also provides no consistent latency or energy gains, with efficiency depending on the compiled model--engine structure. These findings highlight the need to jointly evaluate corruption robustness, numerical precision, deployment cost, and required output representation.

\noindent Our main contributions are:
\begin{itemize}
    \item \textbf{A deployment-oriented corruption benchmark} with 23 corruptions across six families and annotation-consistent updates for geometry- and visibility-altering corruptions.
    \item \textbf{A comprehensive compiled-inference evaluation} of 26 models across four datasets using FP32, FP16, and post-training INT8 TensorRT engines, paired with measured edge throughput and energy.
    \item \textbf{A joint analysis of robustness and efficiency} across corruption severity, crowd size, localization, numerical precision, and engine-level execution behavior.
\end{itemize}
\section{Related Work}
\label{sec:related-work}

\noindent\textbf{Crowd counting and deployment.}
Crowd-counting methods span density regression~\cite{zhang2016single,li2018csrnet,wang2020distribution}, point localization~\cite{song2021rethinking,liu2023point,chen2024improving}, direct regression~\cite{liang2022transcrowd}, and transformer or vision--language architectures~\cite{lin2024gramformer,ma2025clip}. Efficient models are typically compared in terms of parameter count, FLOPs, throughput, latency, and energy~\cite{wang2020mobilecount,ma2025zip,gao2025effectiveness,khan2023lcdnet,khan2025edgekd, hasan2026unified}. However, these measures do not capture how robustness and energy consumption change after compilation into reduced-precision engines. We therefore evaluate corrupted accuracy, latency, and energy across different output representations and numerical precisions.

\noindent\textbf{Corruption robustness.}
Common-corruption benchmarks enable systematic evaluation of classification and dense vision models~\cite{hendrycks2019imagenetc,kamann2021segrobustness,kong2023robodepth,kar2022threedcc,dong2023threeddet}. Crowd-counting studies have investigated adverse weather, adversarial perturbations, domain generalization, and compression~\cite{sindagi2022jhucrowdpp,huang2023awccnet,wu2021advcrowd,peng2024single,bakhtiarnia2022compression}, while a concurrent benchmark examines six corruptions for open-vocabulary object counting~\cite{ma2026tooc}. Our work focuses on fixed-category crowd counting and corruptions spanning the image-formation and transmission pipeline. Because geometric and visibility changes can alter the valid counting target, we update the corresponding head annotations and assess robustness using both absolute corrupted error and degradation relative to the clean baseline.

\noindent\textbf{Quantization under distribution shift.}
Post-training quantization converts trained models to reduced precision through calibration and rounding~\cite{wu2020integer,nagel2020adaround}, although accuracy and efficiency depend on architecture, particularly for depthwise-separable designs~\cite{sheng2018quantfriendly,yun2021mobilenets}. Prior studies report mixed effects on robustness under distribution shift~\cite{xiao2023robustmq,xiao2023benchmarkingquantized,diffenderfer2021winninghand,varghese2022joint,yuan2023ptqreliability}, while real-engine studies largely examine quantized accuracy and deployment cost separately~\cite{karimov2025quantization,arani2022comprehensive,araghi2026rectiq}. We instead evaluate FP32, FP16, and post-training INT8 TensorRT engines within a unified framework, jointly measuring corruption response, throughput, and energy.

\section{The \suite Benchmark}
\label{sec:benchmark}

\subsection{Task and corruption suite}
\label{sec:task}

\suite evaluates density-estimation, point-localization, and direct-regression
models through their final predicted counts. It applies 23 corruptions at three
severity levels to the full labeled test splits of ShanghaiTech~A/B, UCF-QNRF,
and JHU-Crowd++~\cite{zhang2016single,idrees2018qnrf,sindagi2022jhucrowdpp}.
The 2{,}432 test images span fixed-camera views, heterogeneous web imagery,
dense scenes, and adverse capture conditions. Each model--dataset--precision
configuration is evaluated on the corresponding full test split.

\begin{figure}[t]
  \centering
  \begin{tikzpicture}[
      x=1cm,
      y=1cm,
      flow/.style={-{Latex[length=2.0mm]}, line width=0.85pt,
                   draw=grpD!90!black},
      stage/.style={draw=black!55, rounded corners=2pt,
                    minimum width=2.05cm, minimum height=0.48cm,
                    inner xsep=3pt, align=center, font=\footnotesize},
      content/.style={draw=black!55, rounded corners=2pt,
                      minimum width=1.70cm, minimum height=0.56cm,
                      align=center, font=\footnotesize},
      scene image/.style={draw=dsA!70!black, line width=0.65pt,
                          inner sep=0pt, outer sep=0pt},
      camera box/.style={draw=black!55, dashed, rounded corners=3pt,
                         inner xsep=4pt, inner ysep=5pt}
    ]
    \node[content, fill=dsB!15, draw=dsB!75!black] (clutter) at (-2.80,0)
      {Scene clutter\\[-1pt]\scriptsize e.g., object distractor};
    \node[content, fill=dsD!13, draw=dsD!70!black] (visibility) at (2.55,0)
      {Visibility loss\\[-1pt]\scriptsize e.g., occlusion};
    \node[scene image] (scene) at (0,0)
      {\includegraphics[width=2.55cm]
       {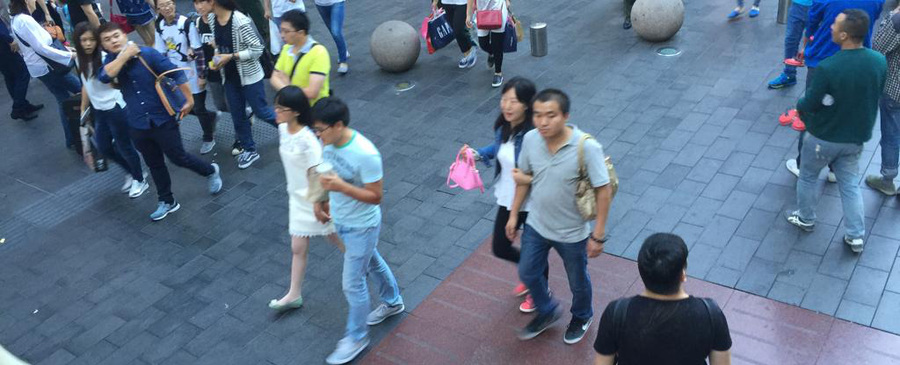}};
    \node[font=\footnotesize\bfseries, fill=white, fill opacity=0.84,
          text opacity=1, inner sep=1.2pt] at ([yshift=-1.8mm]scene.north)
      {Observable scene};
    \node[stage, fill=dsC!12, draw=dsC!70!black] (environment) at (-2.55,-1.55)
      {Environment\\[-1pt]\scriptsize e.g., rain};
    \node[stage, fill=grpD!11, draw=grpD!70!black] (optics) at (0,-1.55)
      {Optics \& motion\\[-1pt]\scriptsize e.g., motion blur};
    \node[stage, fill=int8col!10, draw=int8col!70!black] (sensor) at (2.80,-1.55)
      {Sensor \& exposure\\[-1pt]\scriptsize e.g., low light};
    \node[stage, fill=grpE!10, draw=grpE!70!black] (encoding) at (2.80,-2.70)
      {Encoding \& transport\\[-1pt]\scriptsize e.g., H.264};
    \node[stage, fill=fp16col!10, draw=fp16col!70!black] (model) at (-0.25,-2.70)
      {Counting model};

    \draw[flow] (clutter.east) -- (scene.west);
    \draw[flow] (visibility.west) -- (scene.east);
    \draw[flow, rounded corners=2pt]
      (scene.south) -- (0,-0.82) -| (environment.north);
    \draw[flow] (environment.east) -- (optics.west);
    \draw[flow] (optics.east) -- (sensor.west);
    \draw[flow] (sensor.south) -- (encoding.north);
    \draw[flow] (encoding.west) -- (model.east);

    \node[camera box, fit=(optics)(sensor),
          label={[font=\footnotesize\bfseries, text=black!65]above:Camera}] {};
  \end{tikzpicture}
  \caption{\textbf{Corruption sources along the capture pipeline.} Four
  families act during image formation and delivery; scene clutter and
  visibility loss act beforehand.}
  \label{fig:capture-process}
\vspace{-1em}
\end{figure}
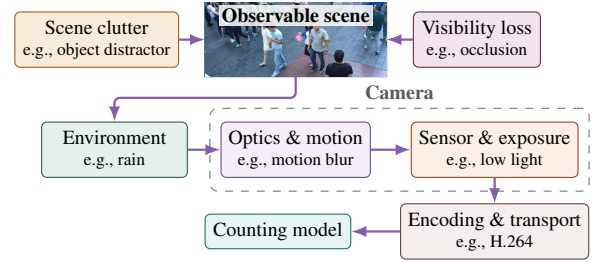

We organize the corruption set $\mathcal C$ into six sources of degradation
that can occur before inference (\cref{fig:capture-process,tab:overview}).
Four arise during image formation and delivery: environmental conditions,
optics and motion, sensor and exposure, and encoding and transport. The
remaining families address scene content: clutter introduces plausible
distractors, while visibility loss removes observable people. Of the 23
corruptions, 16 preserve all annotations; for the remaining seven, we
transform annotations to exclude heads removed by cropping, warping, or
occlusion. To maintain a comparable target scale, we normalize the resulting
MAE by the visible-head fraction:
$\mathrm{MAE}^{\mathrm{kept}}=\mathrm{MAE}/(1-f_{\mathrm{drop}})$,
which equals MAE for the 16 annotation-preserving corruptions
($f_{\mathrm{drop}}=0$). The removed fraction has a median of 9.4\%
and a mean of 13.0\%.

\begin{table}[t]
\centering
\caption{\textbf{\suite at a glance.} \textbf{(a)} Full labeled test splits,
2{,}432 images. \textbf{(b)} 23 corruptions in six families;
\emph{alt.} counts the 7 geometry-altering members.}
\label{tab:overview}
\scriptsize
\renewcommand{\arraystretch}{0.92}
\setlength{\tabcolsep}{2.4pt}
\begin{tabular}{@{}l l r c c r@{}}
\multicolumn{6}{@{}l}{\textbf{(a) base datasets and the instances they generate}} \\
\toprule
Dataset & Inst. & \# test images & Resolution\ (MP) & count min\,/\,max & mean \\
\midrule
ShanghaiTech A & \textbf{SHA-C} & 182 & $0.06$--$1.02$ & 66 / 2{{,}}256 & 433.3 \\
ShanghaiTech B & \textbf{SHB-C} & 316 & $0.79$ & 9 / 539 & 123.8 \\
UCF-QNRF & \textbf{QNRF-C} & 334 & $0.08$--$36.15$ & 65 / 4{{,}}535 & 718.9 \\
JHU-Crowd++ & \textbf{JHU-C} & 1{,}600 & $0.05$--$36.15$ & 0 / 9{{,}}059 & 326.2 \\
\bottomrule
\end{tabular}

\vspace{2pt}

\setlength{\tabcolsep}{2.2pt}
\begin{tabular}{@{}l l c c p{3.05cm}@{}}
\multicolumn{5}{@{}l}{\textbf{(b) corruption families, in capture-chain order}} \\
\toprule
Family & Chain & $n$ & alt. & Corruptions \\
\midrule
Environment & scene & 4 & -- & rain streaks, snow particles, depth-aware haze, cast shadow \\
Optics \& motion & optics & 4 & 1 & camera motion blur, far-field defocus, lens flare, lens contamination \\
Sensor \& exposure & sensor & 5 & -- & Gaussian noise, fixed-pattern noise, low light, highlight clipping, vignetting \\
Encoding \& transport & encoding & 4 & 1 & H.264 compression, moir\'e aliasing, head-spacing loss, packet-loss concealment \\
Scene clutter & \emph{content} & 3 & 2 & texture distractors, crowd-like distractors, object distractors \\
Visibility loss & \emph{content} & 3 & 3 & viewpoint shift, foreground occlusion, field-of-view loss \\
\bottomrule
\end{tabular}
\vspace{-1em}
\end{table}

\begin{figure*}[t]
  \centering
  \setlength{\tabcolsep}{1.2pt}
  \renewcommand{\arraystretch}{0.6}
  \begin{tabular}{@{}c cccccc@{}}
  & \scriptsize\shortstack{low\\light} & \scriptsize\shortstack{camera\\motion blur} & \scriptsize\shortstack{snow\\particles} & \scriptsize\shortstack{packet-loss\\concealment} & \scriptsize\shortstack{object\\distractors} & \scriptsize\shortstack{foreground\\occlusion} \\[1pt]
\rotatebox{90}{\scriptsize Light} & \includegraphics[width=0.152\textwidth]{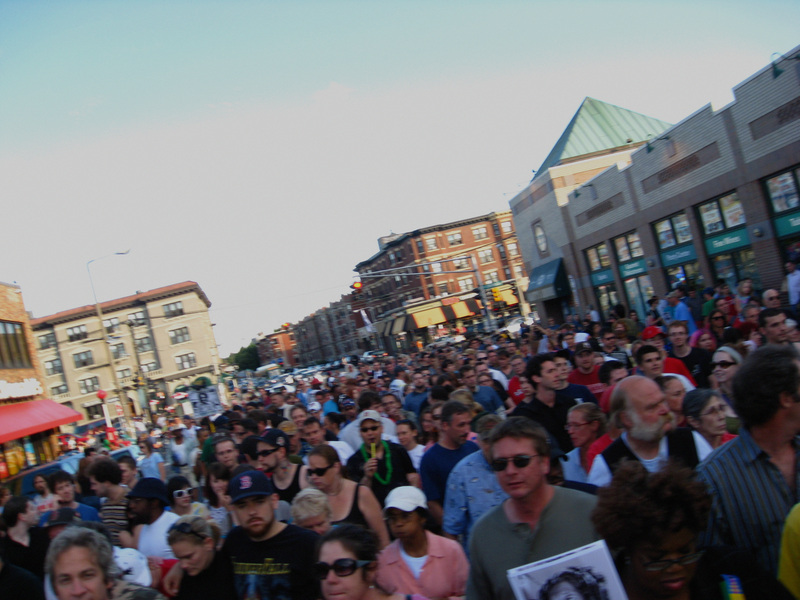} & \includegraphics[width=0.152\textwidth]{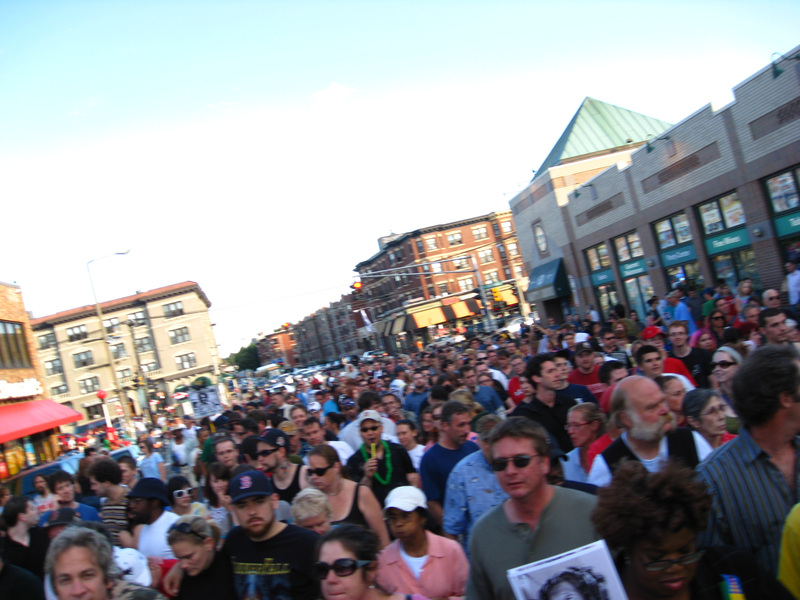} & \includegraphics[width=0.152\textwidth]{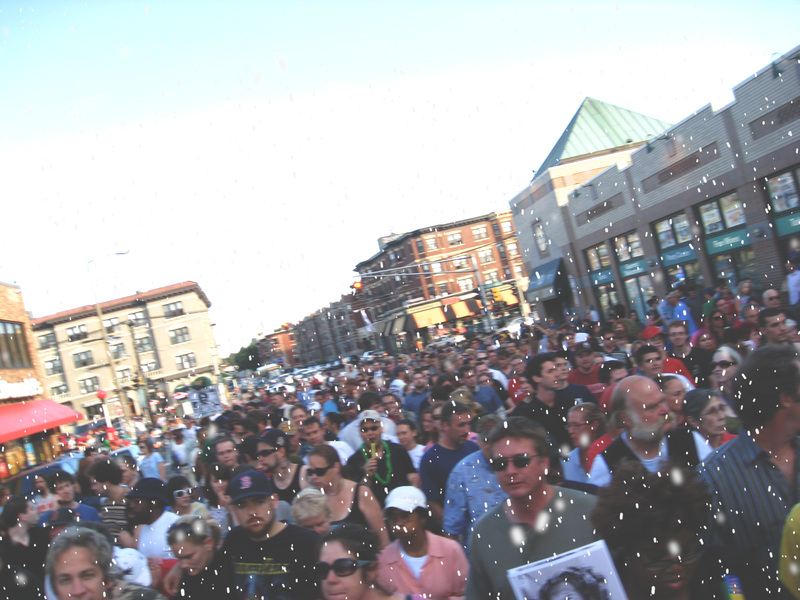} & \includegraphics[width=0.152\textwidth]{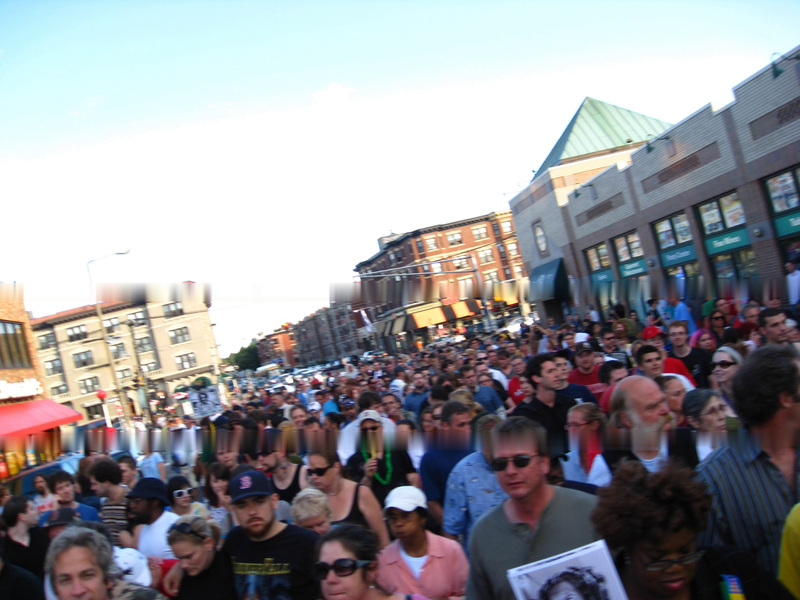} & \includegraphics[width=0.152\textwidth]{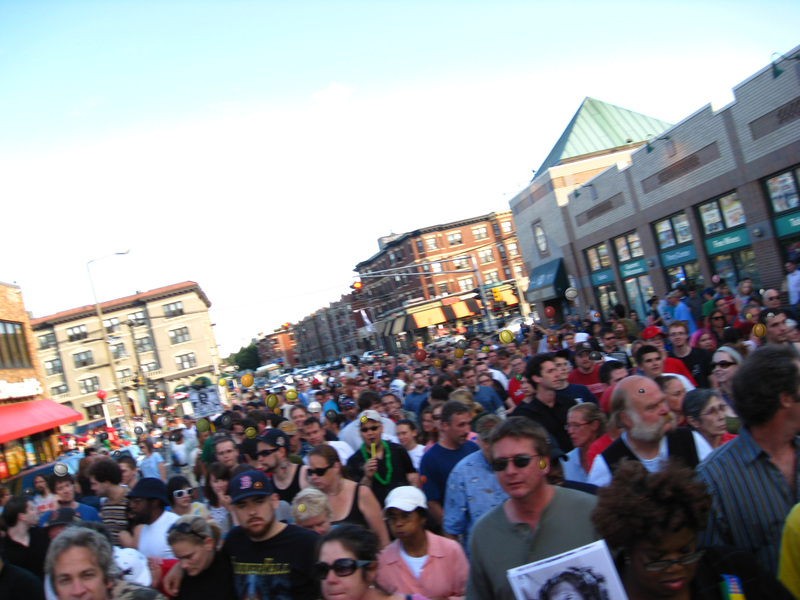} & \includegraphics[width=0.152\textwidth]{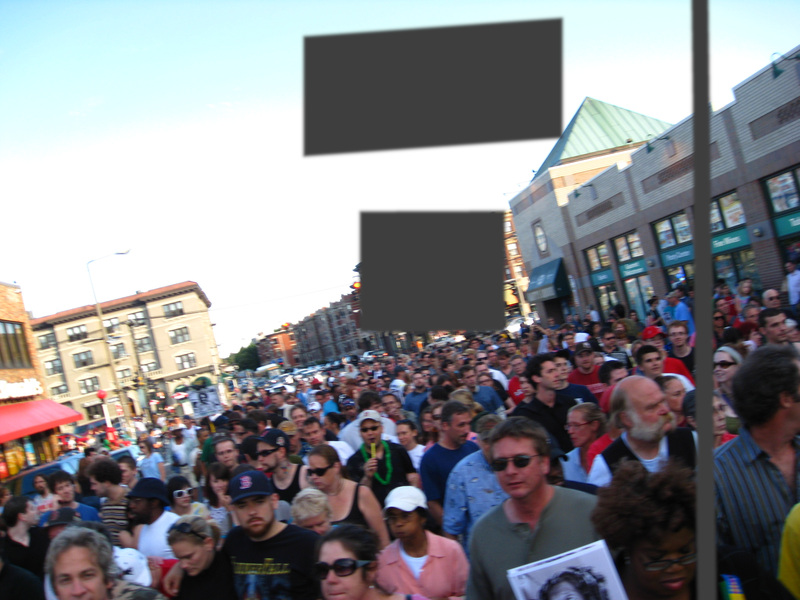} \\
\rotatebox{90}{\scriptsize Moderate} & \includegraphics[width=0.152\textwidth]{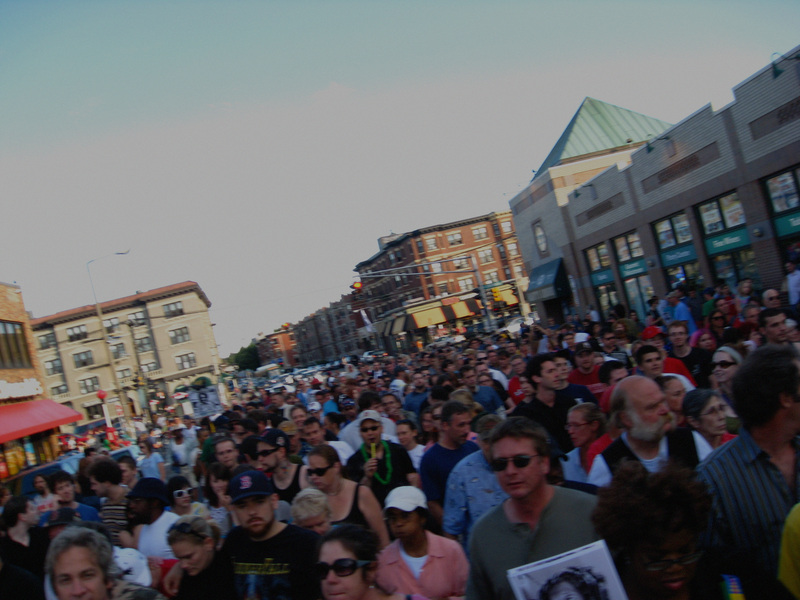} & \includegraphics[width=0.152\textwidth]{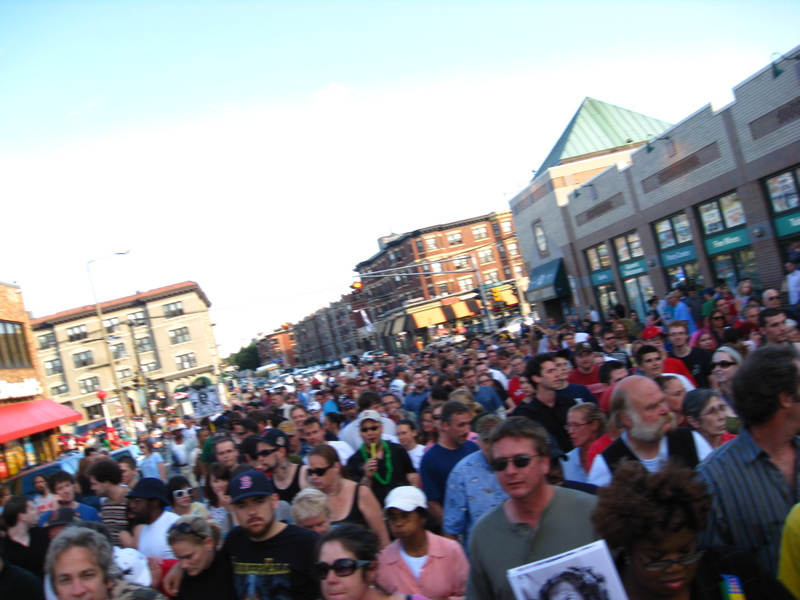} & \includegraphics[width=0.152\textwidth]{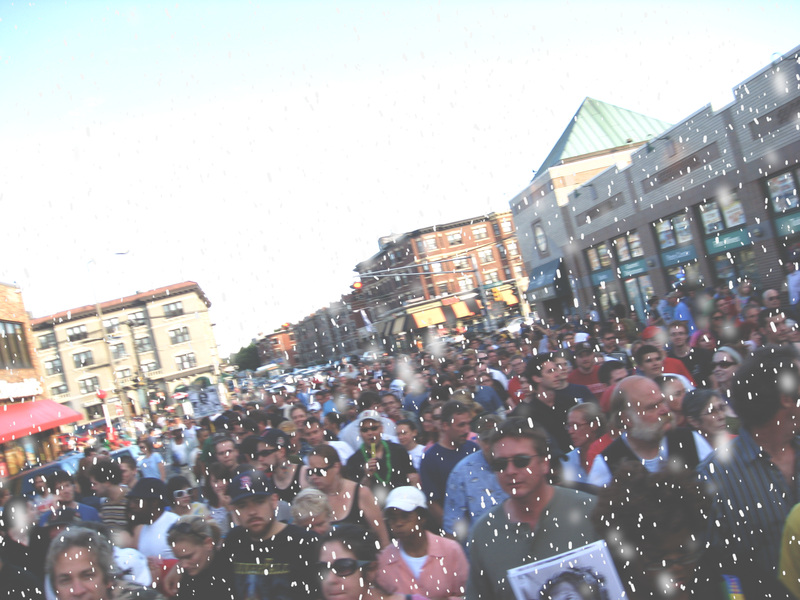} & \includegraphics[width=0.152\textwidth]{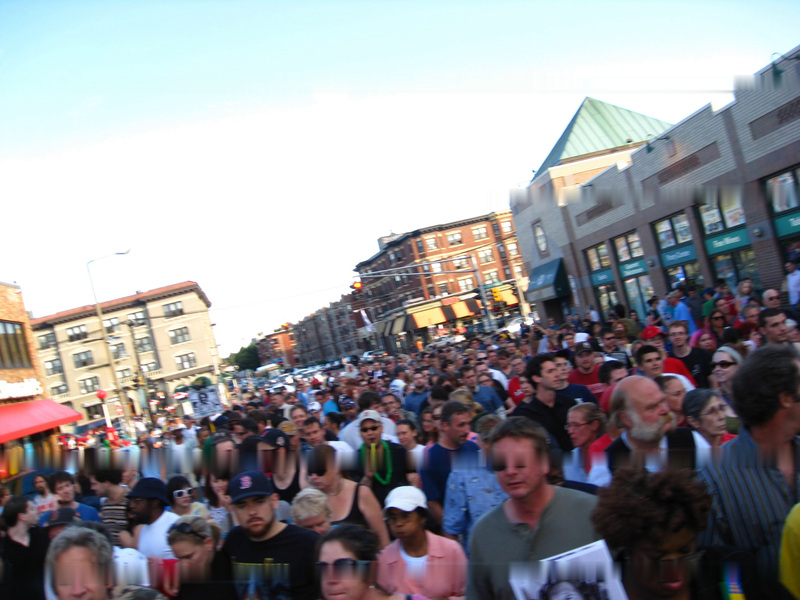} & \includegraphics[width=0.152\textwidth]{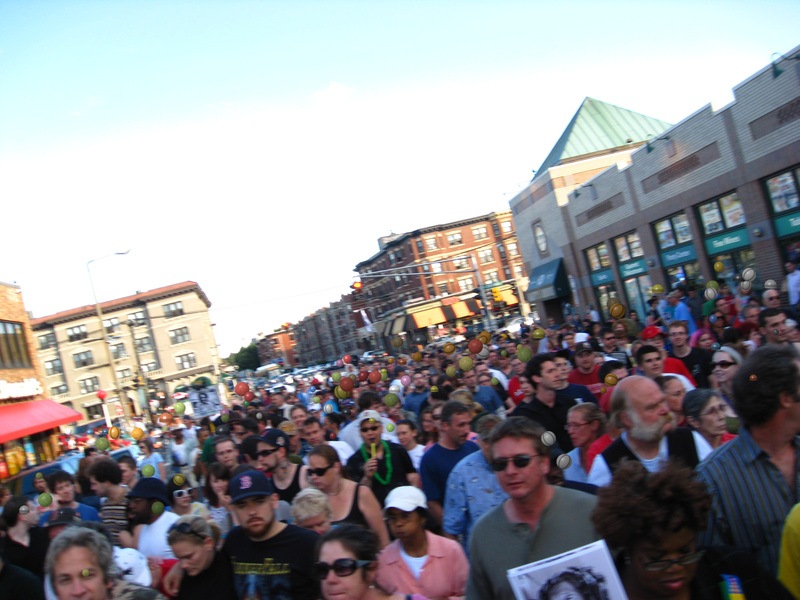} & \includegraphics[width=0.152\textwidth]{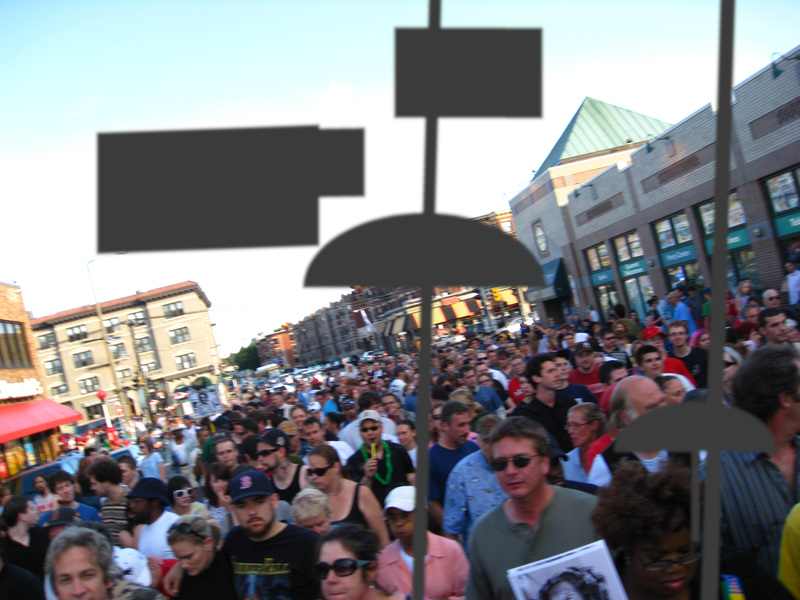} \\
\rotatebox{90}{\scriptsize High} & \includegraphics[width=0.152\textwidth]{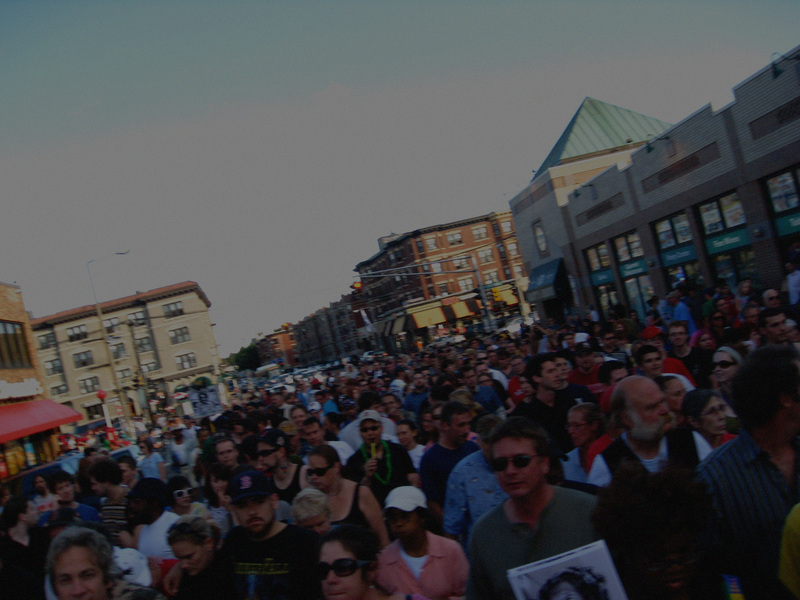} & \includegraphics[width=0.152\textwidth]{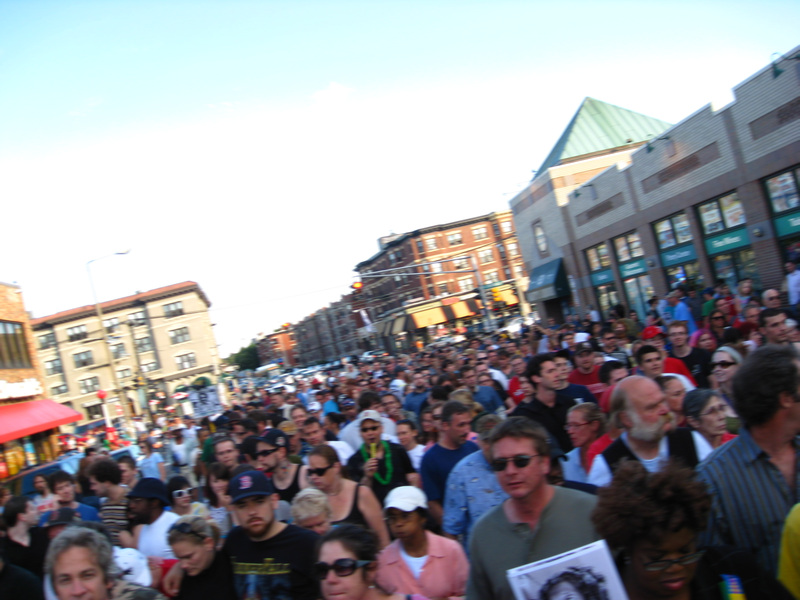} & \includegraphics[width=0.152\textwidth]{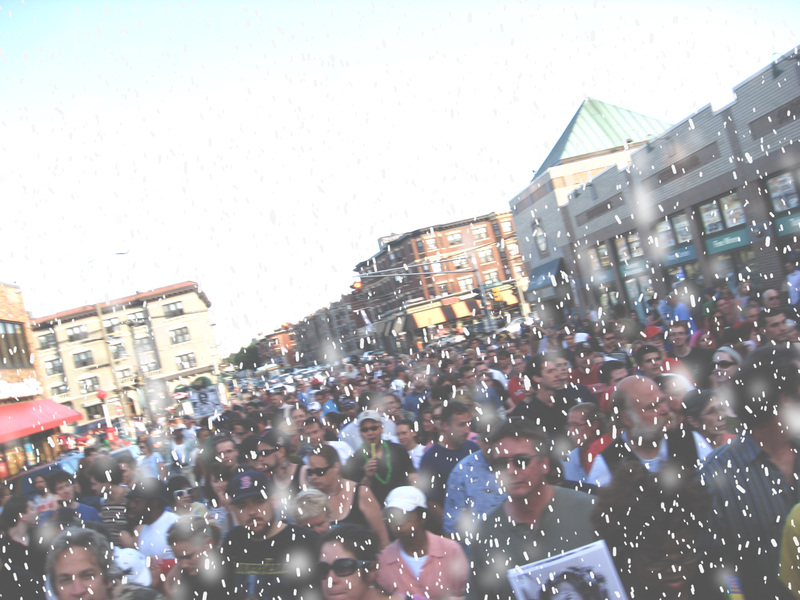} & \includegraphics[width=0.152\textwidth]{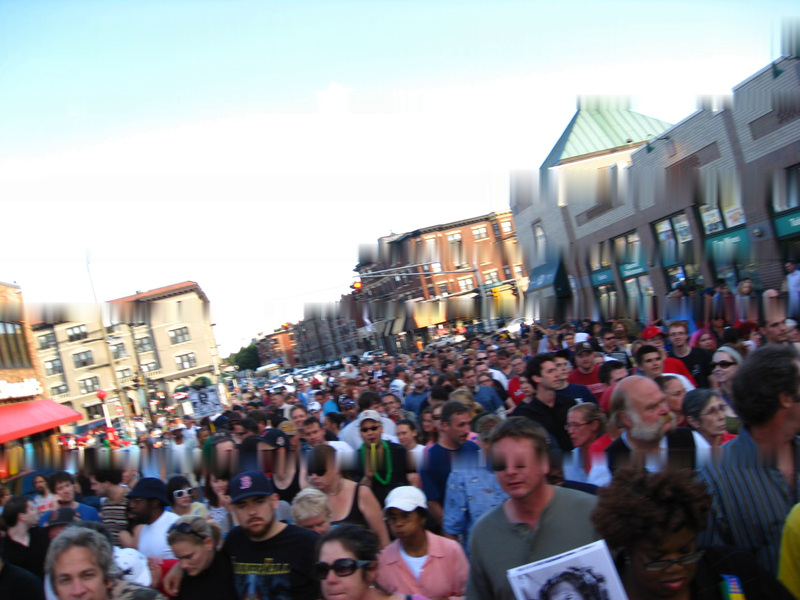} & \includegraphics[width=0.152\textwidth]{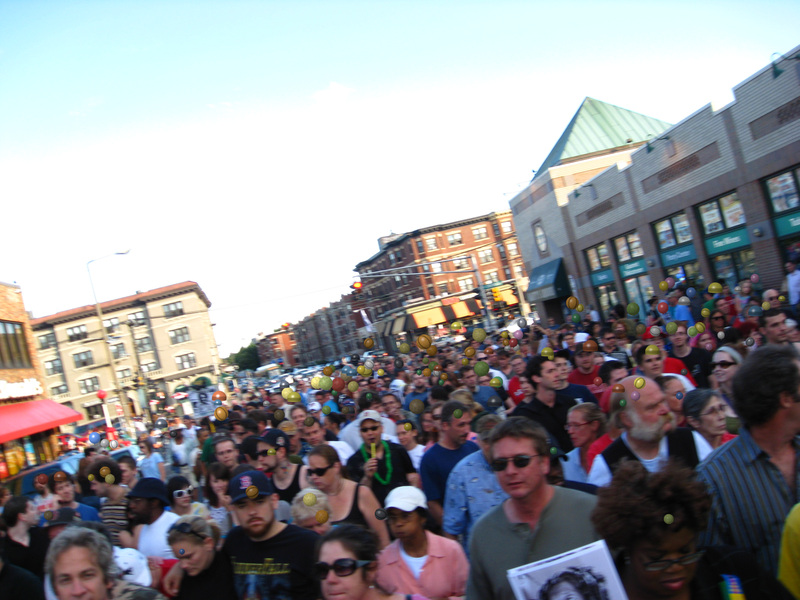} & \includegraphics[width=0.152\textwidth]{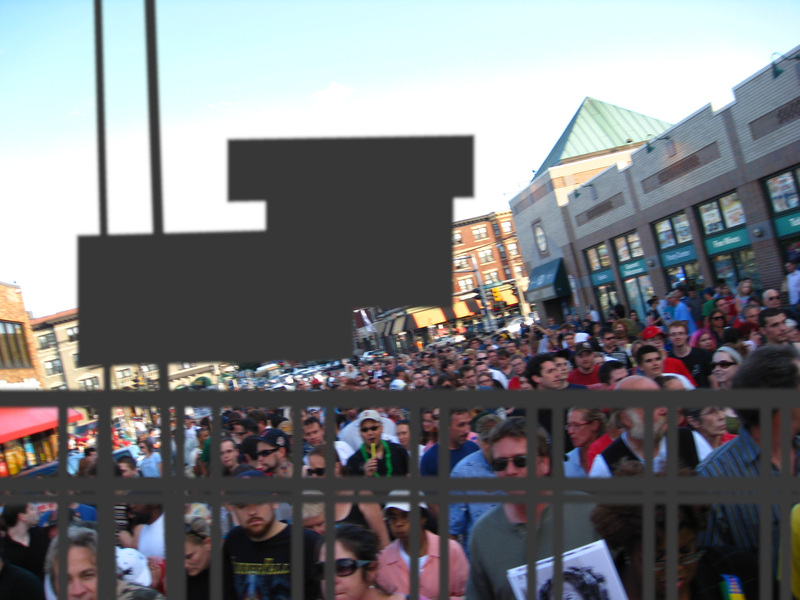} \\
  \end{tabular}

  \caption{\textbf{Corruption severity across six families.} One
  UCF-QNRF image is shown for one corruption from each family in
  \cref{tab:overview}.}
  \label{fig:corruption-gallery}
  \vspace{-0.7em}
\end{figure*}

\subsection{Corruption generation and severity calibration}
\label{sec:severity}
\label{sec:validity}

The severity set $\mathcal S=\{L,M,H\}$ denotes Light, Moderate, and High.
Levels are ordered within each corruption but do not imply equal perceptual
strength or counting difficulty across corruption types. Calibration follows
four complementary procedures. First, corruptions governed by standard
acquisition or transmission controls use interpretable generation parameters,
such as the exposure multiplier for low light and CRF for H.264 compression.
Second, effects whose apparent scale depends on crowd geometry are expressed
relative to the median spacing between neighboring heads in each image.
This converts, for example, motion-blur length and effective resolution into
image-specific pixel values. Third, rain and snow are calibrated using
appearance statistics from labeled JHU-Crowd++ training images: rain contrast
and saturation, and snow luminance and saturation, are selected to remain
within ranges observed in real weather. Fourth, a structured review of
rendered training images verifies that head annotations remain supported by
visible evidence. The supplement reports the governing parameters and
calibrated values for all 23 corruptions, together with mechanism assertions,
calibration checks, and redundancy analysis.

\subsection{Metrics}
\label{sec:metrics}

Metrics are computed separately for each model--dataset configuration. For
corruption $c\in\mathcal C$, severity $s\in\mathcal S$, and precision
$p\in\{\mathrm{FP32},\mathrm{FP16},\mathrm{INT8}\}$, let
$E_{\varnothing,p}$ denote clean MAE and $E_{c,s,p}$ denote corrupted MAE.
Here $|\mathcal C|=23$ and $|\mathcal S|=3$ are the numbers of corruptions
and severity levels, respectively. We define
\begingroup
\scriptsize
\begin{equation}
\begin{aligned}
\mcmae_p
&= \frac{1}{|\mathcal C|}\sum_{c\in\mathcal C}
\frac{1}{|\mathcal S|}\sum_{s\in\mathcal S}E_{c,s,p},\\
\mathrm{RD}_p
&= 100\,\frac{\mcmae_p-E_{\varnothing,p}}{E_{\varnothing,p}},\\
\drd(p)
&= \mathrm{RD}_p-\mathrm{RD}_{\mathrm{FP32}}.
\label{eq:robustness-metrics}
\end{aligned}
\end{equation}
\endgroup
Mean-corruption MAE ($\mcmae_p$) measures absolute corrupted error, while RD
measures degradation relative to clean error at the same precision. The
signed $\drd$ and its absolute value $|\drd|$ capture the direction and
magnitude of the precision interaction, respectively.

Adapting ImageNet-C's mCE~\cite{hendrycks2019imagenetc}, we average
$\overline{E}_{c,p}/\overline{E}^{\,\mathrm{CSRNet}}_{c,p}$ across corruptions.
Each ratio uses three-severity MAE and a CSRNet reference matched by dataset
and precision~\cite{li2018csrnet}. Lower is better, with CSRNet equal to one.
Unlike RD, mCE compares corrupted error against a fixed reference.
\section{Experimental Setup}
\label{sec:setup}

We describe the evaluated models and precision settings, INT8 calibration
procedure, and platforms used to measure accuracy and deployment cost.

\subsection{Models and precision settings}
\label{sec:roster}
\label{sec:precision}

We evaluate 26 crowd-counting models: 18 CNN-based models, five transformers,
two hybrids, and one vision--language model. Their output representations
comprise 15 density maps, five blockwise count maps, four localization-oriented
outputs, and two direct count regressors. The models span
$0.13$--$300.74$\,M parameters. Each model is trained using its
architecture-specific recipe to recover its reported clean accuracy.
\Cref{tab:model-engine} summarizes the evaluated models, exported-graph
structure, and realized INT8 engines.

\begin{table}[t]
\centering
\caption{\textbf{Model outputs and INT8 efficiency.}
D/B/P/C denote density maps, blockwise count maps, localization outputs, and direct count regression.
The \texttt{FP16 or FP32 ops.} column lists operations retained above INT8 ($^{\mathrm{f}}$ indicates FP32).
Speed and energy ratios are four-dataset medians relative to FP16.
\textbf{Bold} marks INT8 configurations with slower execution or higher energy consumption than FP16.}
\label{tab:model-engine}
\scriptsize
\renewcommand{\arraystretch}{0.88}
\setlength{\tabcolsep}{0.8pt}
\resizebox{\columnwidth}{!}{%
\begin{tabular}{@{}l r c l r r@{}}
\toprule
 & & & \multicolumn{3}{c}{Engine} \\
\cmidrule(l){4-6}
Model & $|\theta|$ (M) & \shortstack{Output\\type} & \shortstack{FP16 or\\FP32 ops.} & INT8$\times$ & $E_8/E_{16}$ \\
\midrule
\multicolumn{6}{@{}l}{\textbf{CNN-based}} \\
\quad CHF~\cite{shu2022crowd} & 21.51 & D & -- & 1.99 & 0.50 \\
\quad DM-Count~\cite{wang2020distribution} & 21.50 & D & -- & 1.98 & 0.48 \\
\quad GCHF~\cite{shu2022crowd} & 21.51 & D & -- & 1.98 & 0.49 \\
\quad CSRNet~\cite{li2018csrnet} & 16.26 & D & -- & 1.97 & 0.49 \\
\quad P2PNet~\cite{song2021rethinking} & 21.58 & P & -- & 1.94 & 0.53 \\
\quad APGCC~\cite{chen2024improving} & 18.69 & P & \texttt{\scriptsize scal}, \texttt{\scriptsize rem}$^{\mathrm{f}}$ & 1.67 & 0.58 \\
\quad STEERER-VGG19~\cite{han2023steerer} & 30.62 & D & -- & 1.65 & 0.59 \\
\quad FIDTM~\cite{liang2022focal} & 66.58 & P & \texttt{\scriptsize dec} & 1.44 & 0.68 \\
\quad ZIP-Tiny~\cite{ma2025zip} & 10.53 & B & \texttt{\scriptsize dec}, \texttt{\scriptsize head} & 1.39 & 0.84 \\
\quad STEERER-HRNet~\cite{han2023steerer} & 64.64 & D & \texttt{\scriptsize Add} & 1.35 & 0.68 \\
\quad MobileCount~\cite{wang2020mobilecount} & 3.40 & D & \texttt{\scriptsize Add}, \texttt{\scriptsize pool}, \texttt{\scriptsize head} & 1.19 & 0.92 \\
\quad ZIP-Nano~\cite{ma2025zip} & 3.37 & B & \texttt{\scriptsize dec}, \texttt{\scriptsize head} & 1.14 & 0.93 \\
\quad MPCount~\cite{peng2024single} & 33.23 & D & -- & 1.14 & 0.72 \\
\quad ZIP-Pico~\cite{ma2025zip} & 0.82 & B & \texttt{\scriptsize dec}, \texttt{\scriptsize head} & 1.08 & 0.97 \\
\quad ZIP-Small~\cite{ma2025zip} & 33.60 & B & \texttt{\scriptsize enc}, \texttt{\scriptsize L2}$^{\mathrm{f}}$, \texttt{\scriptsize keep} & \textbf{0.98} & 0.88 \\
\quad FFNet-ConvNeXt-T~\cite{gao2025effectiveness} & 34.58 & D & \texttt{\scriptsize enc}, \texttt{\scriptsize dec} & \textbf{0.97} & \textbf{1.02} \\
\quad FFNet-MobileNetV3~\cite{gao2025effectiveness} & 3.29 & D & \texttt{\scriptsize enc} & \textbf{0.87} & \textbf{1.02} \\
\quad MCNN~\cite{zhang2016single} & 0.13 & D & -- & \textbf{0.72} & \textbf{1.10} \\
\multicolumn{6}{@{}l}{\textbf{Transformer}} \\
\quad CCST~\cite{li2023ccst} & 300.74 & D & \texttt{\scriptsize Add} & 1.82 & 0.50 \\
\quad Transcrowd-GAP~\cite{liang2022transcrowd} & 90.40 & C & \texttt{\scriptsize attn}, \texttt{\scriptsize Add} & 1.21 & 0.75 \\
\quad Transcrowd-Token~\cite{liang2022transcrowd} & 86.86 & C & \texttt{\scriptsize attn}, \texttt{\scriptsize Add} & 1.20 & 0.75 \\
\quad CCTrans-Full~\cite{tian2021cctrans} & 104.61 & D & \texttt{\scriptsize enc}, \texttt{\scriptsize Add} & 1.08 & 0.84 \\
\quad CCTrans-Weak~\cite{tian2021cctrans} & 104.61 & D & \texttt{\scriptsize enc}, \texttt{\scriptsize Add}, \texttt{\scriptsize head} & 1.04 & 0.88 \\
\multicolumn{6}{@{}l}{\textbf{Hybrid}} \\
\quad Gramformer~\cite{lin2024gramformer} & 29.01 & D & -- & 1.36 & 0.67 \\
\quad PET~\cite{liu2023point} & 20.91 & P & -- & 1.34 & 0.68 \\
\multicolumn{6}{@{}l}{\textbf{Vision--language}} \\
\quad CLIP-EBC-RN50~\cite{ma2025clip} & 135.47 & B & \texttt{\scriptsize L2}$^{\mathrm{f}}$ & 2.26 & 0.44 \\
\bottomrule
\end{tabular}}
\vspace{-1em}
\end{table}

\begin{table}[t]
\centering
\caption{\textbf{Clean MAE by precision.} Entries are
FP32/FP16/INT8; a bold INT8 value is lower than at least one
higher-precision value.}
\label{tab:clean-mae}
\scriptsize
\renewcommand{\arraystretch}{0.88}
\setlength{\tabcolsep}{2.0pt}
\resizebox{\linewidth}{!}{%
\begin{tabular}{@{}l r r r r@{}}
\toprule
Model & SHA & SHB & QNRF & JHU \\
\midrule
CHF & 57.2/57.3/\textbf{55.7} & 7.3/7.3/7.9 & 84.2/84.1/85.7 & 60.4/60.4/\textbf{60.3} \\
DM-Count & 65.7/65.7/66.0 & 7.8/7.8/7.9 & 91.5/91.5/91.8 & 62.4/62.4/62.4 \\
GCHF & 59.2/59.2/\textbf{58.1} & 7.1/7.1/8.3 & 81.8/81.8/84.1 & 60.4/60.3/60.5 \\
CSRNet & 67.4/67.4/68.2 & 11.4/11.4/\textbf{11.1} & 95.8/95.9/96.7 & 68.9/68.9/\textbf{68.8} \\
P2PNet & 57.8/57.8/\textbf{57.5} & 8.0/8.0/8.1 & 101.0/101.1/102.0 & 66.9/66.9/\textbf{65.8} \\
APGCC & 57.9/57.9/\textbf{57.3} & 8.3/8.3/8.6 & 110.4/110.4/\textbf{105.5} & 72.7/72.8/73.4 \\
STEERER-VGG19 & 62.3/62.4/\textbf{59.6} & 7.9/7.9/8.6 & 86.4/86.1/91.2 & 58.8/58.8/63.1 \\
FIDTM & 66.9/66.5/67.3 & 8.5/9.0/\textbf{7.5} & 107.4/106.8/107.5 & 69.2/68.7/70.9 \\
ZIP-Tiny & 62.9/62.9/\textbf{61.3} & 7.5/7.5/8.2 & 91.7/91.8/\textbf{88.2} & 62.4/62.4/\textbf{61.5} \\
STEERER-HRNet & 54.4/54.4/59.1 & 6.3/6.3/\textbf{6.2} & 79.5/79.6/\textbf{76.7} & 54.4/54.4/\textbf{54.3} \\
MobileCount & 86.7/86.6/88.3 & 9.2/9.2/9.2 & 114.7/114.7/116.3 & 137.0/137.0/140.0 \\
ZIP-Nano & 70.5/70.6/\textbf{62.3} & 8.5/8.5/8.9 & 114.4/114.3/122.9 & 73.5/73.5/\textbf{70.0} \\
MPCount & 61.2/61.2/61.7 & 6.5/6.5/\textbf{6.4} & 110.1/110.1/\textbf{108.8} & 59.2/59.2/\textbf{57.8} \\
ZIP-Pico & 79.9/79.9/\textbf{74.9} & 9.7/9.7/9.8 & 111.6/111.5/115.4 & 76.0/75.9/91.4 \\
ZIP-Small & 71.5/71.6/\textbf{64.6} & 7.3/7.3/8.8 & 76.9/76.8/77.8 & 56.8/56.8/\textbf{55.9} \\
FFNet-ConvNeXt-T & 54.9/54.9/56.5 & 6.8/6.8/7.8 & 84.1/84.1/89.4 & 66.1/66.1/\textbf{59.5} \\
FFNet-MobileNetV3 & 70.7/70.8/\textbf{68.8} & 12.6/12.8/\textbf{11.4} & 107.9/108.1/109.3 & 65.1/65.1/65.9 \\
MCNN & 99.4/99.4/\textbf{98.4} & 18.3/18.3/\textbf{18.2} & 177.8/177.8/\textbf{174.7} & 118.1/118.1/\textbf{117.9} \\
CCST & 63.6/63.6/\textbf{63.2} & 8.1/8.1/8.3 & 96.3/96.2/96.7 & 61.4/61.4/\textbf{61.2} \\
Transcrowd-GAP & 67.0/67.0/68.4 & 10.2/10.2/10.4 & 91.4/91.3/96.1 & 68.0/68.0/\textbf{67.3} \\
Transcrowd-Token & 70.8/70.7/73.1 & 9.1/9.1/9.2 & 91.4/91.4/95.4 & 62.9/62.9/64.1 \\
CCTrans-Full & 57.3/57.4/59.7 & 7.5/7.5/8.1 & 92.3/92.3/\textbf{91.2} & 56.8/56.8/59.1 \\
CCTrans-Weak & 75.6/75.5/75.9 & 8.3/8.3/8.7 & 109.7/109.8/\textbf{105.3} & 61.8/61.8/62.4 \\
Gramformer & 60.3/60.3/60.4 & 7.2/7.2/8.0 & 80.6/80.7/85.8 & 55.9/55.9/59.3 \\
PET & 52.4/52.3/54.6 & 6.8/6.7/6.9 & 100.5/100.6/\textbf{98.7} & 61.9/61.9/\textbf{61.1} \\
CLIP-EBC-RN50 & 56.4/56.4/57.2 & 6.7/6.7/6.9 & 91.2/91.2/91.7 & 60.1/60.1/60.3 \\
\bottomrule
\end{tabular}%
}
\vspace{-2em}
\end{table}

We evaluate every model--dataset configuration using TensorRT~\cite{tensorrt}
engines executed at FP32, FP16, and post-training INT8 precision. FP32 serves
as the deployment baseline, while FP16 provides a compiled reduced-precision
reference. The median absolute difference in clean MAE between FP16 and FP32
is only $0.048\%$, providing a near-equivalent baseline for assessing the
additional effects of INT8.

\subsection{INT8 calibration}
\label{sec:int8-calibration}
\label{sec:int8-selection}

We apply graph-level post-training quantization to each exported ONNX graph
using NVIDIA ModelOpt~\cite{nvidia-modelopt}. Eligible
weights and activations are quantized to INT8, while unsupported operations
and accuracy-sensitive blocks remain in FP16 or FP32, as reported in
\cref{tab:model-engine}. For each model--dataset configuration, we calibrate
using a fixed, count-stratified subset of clean training images, including
large images and extreme aspect ratios representative of deployed input
shapes. The subsets contain 100/120/370/690 images for SHA/SHB/QNRF/JHU,
respectively. We estimate activation ranges from these images and select a
clipping threshold for each configuration. This threshold limits extreme
activation values before mapping them to the INT8 numerical range.

\subsection{Evaluation platforms}
\label{sec:edge-protocol}

We train the models, calibrate INT8 configurations, export ONNX graphs, and
build A100-specific TensorRT engines on an NVIDIA A100 80\,GB PCIe server
with 32 Intel Xeon Platinum 8358 CPU cores and 256\,GB host memory.
Using these A100 engines, we compute clean and corrupted MAE, RMSE, and
localization metrics on the complete labeled test split of each dataset.

We measure deployment cost on a Jetson Orin NX 16\,GB operating in 40\,W
mode. TensorRT builds Jetson-specific engines from the same exported ONNX
graphs. We neither recalibrate the INT8 configurations nor re-export the
graphs for the Jetson. For each model--dataset--precision configuration,
we measure throughput and energy on the same stratified 100-image subset.
\section{Results}
\label{sec:results}

\subsection{Clean accuracy and metric interpretation}
\label{sec:clean}

\Cref{tab:clean-mae} reports clean MAE at each precision, providing a baseline for interpreting corruption robustness. Reduced precision does not uniformly degrade clean accuracy: INT8 and FP16 outperform FP32 in 41 and 48 of the 104 model–dataset comparisons, respectively, with the largest improvement ($8.16$) observed for ZIP-Nano on ShanghaiTech~A.

Relative degradation (RD) and mean corruption error (mCE), both defined in \cref{sec:metrics}, capture distinct aspects of robustness: a model with high clean error may exhibit low RD despite high corrupted error. RD is also sensitive to the clean-error denominator, particularly on ShanghaiTech~B, where median clean MAE is $7.94$, compared with $63.23$ on ShanghaiTech~A. We therefore report clean error, corrupted error, and RD jointly.

\subsection{Research questions and findings}
\label{sec:research-questions}

Using these baselines, we address five questions linking corruption robustness to deployment decisions.

\noindent\textit{\textbf{RQ1:} Which degradations pose the greatest counting risk?} We rank FP32 corruption effects using MAE and RMSE.

\noindent\textit{\textbf{RQ2:} Do clean accuracy and output representation predict robustness?} We compare model rankings across corruptions, precisions, and output representations.

\noindent\textit{\textbf{RQ3:} How do crowd size and output requirements affect failures?} We examine absolute and proportional damage, alongside counting and localization errors.

\noindent\textit{\textbf{RQ4:} Does reduced precision improve robustness and efficiency?} We compare FP16 and INT8 corruption responses, Jetson costs, and engine diagnostics.

\noindent\textit{\textbf{RQ5:} Does joint robustness--cost evaluation change model selection?} We pair A100 corrupted error with Jetson energy to compare frontiers across datasets and precisions.

\smallskip

\noindent\textbf{RQ1: Corruption difficulty and metric agreement.}

\begin{figure}[t]
\centering
\includegraphics[width=\linewidth]{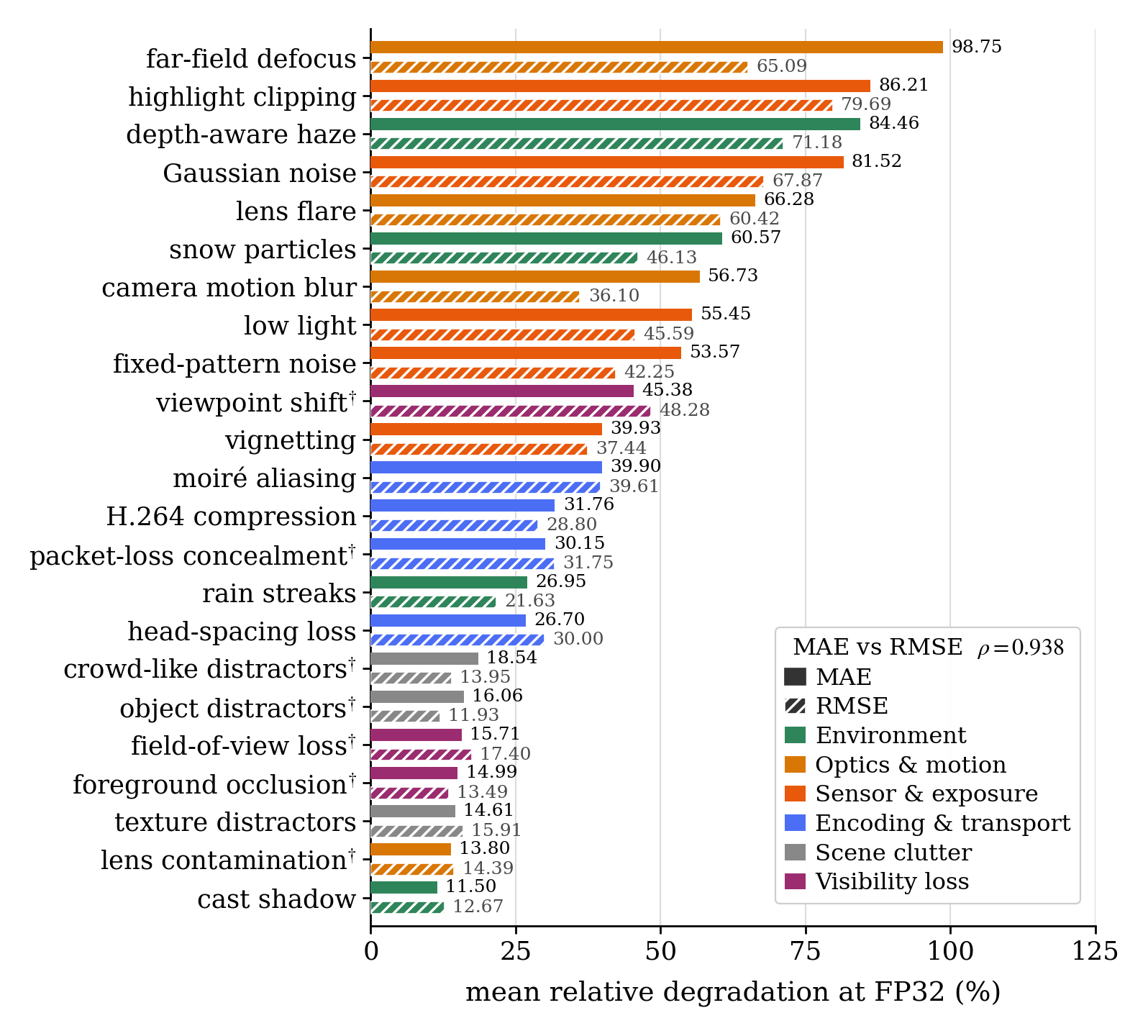}
\caption{\textbf{FP32 corruption sensitivity.} Mean relative degradation across 26 models, four datasets, and three severities, ranked by MAE. $^{\dagger}$ Geometry-altering corruptions use transformed GT.}
\label{fig:corruption-ranking}
\vspace{-1em}
\end{figure}

\noindent\Cref{fig:corruption-ranking} ranks the 23 corruptions by mean relative degradation at FP32,
$\overline{\mathrm{RD}}(c)=\tfrac{1}{|\mathcal D|}\sum_d
\tfrac{1}{|\mathcal M||\mathcal S|}\sum_{m,s}\mathrm{RD}(c,d,m,s)$,
across $|\mathcal M|=26$ models, $|\mathcal S|=3$ severities, and
$|\mathcal D|=4$ datasets, ensuring that no dataset dominates through its image count.
Under the evaluated benchmark settings, far-field defocus induces the largest degradation ($98.75\%$), followed by highlight clipping ($86.21\%$), depth-aware haze ($84.46\%$), Gaussian noise ($81.52\%$), and lens flare ($66.28\%$). The nine most detrimental corruptions arise during image formation and transmission, whereas scene-clutter corruptions rank lower. Cast shadow exhibits the smallest mean degradation ($11.50\%$). This ordering reflects the underlying mechanisms: defocus suppresses head-scale structure, sensor noise obscures local evidence, and haze and highlight clipping reduce head--background contrast.

The ranking is robust to the choice of counting metric. \Cref{fig:corruption-ranking} evaluates each corruption using both MAE and RMSE, computed relative to the corresponding clean baseline. The two metrics agree strongly (Spearman $\rho=0.938$ pooled; $0.841$--$0.973$ across individual datasets) and identify the same five most challenging corruptions. Camera motion blur drops five positions under RMSE, whereas viewpoint shift rises four, suggesting different concentrations of error across scenes.

\smallskip
\noindent\textbf{RQ2: Corrupted error, degradation, and output representation.}

\noindent\Cref{tab:model-robustness} reports $\mcmae$ for each model and dataset, together with median RD and mean mCE at FP32. Clean accuracy does not preserve the corrupted-error ranking: the model with the lowest clean MAE differs from that with the lowest $\mcmae$ on three of four datasets, with ZIP-Small leading both only on UCF-QNRF. Model preference also varies with precision. STEERER-HRNet and CCTrans-Full minimize FP32 $\mcmae$ on ShanghaiTech~A and JHU-Crowd++, respectively, whereas PET and STEERER-HRNet minimize the corresponding INT8 values. The FP32 summaries further distinguish robustness characteristics: median RD ranges from $12.2\%$ for CCST to $30.4\%$ for MobileCount, yet neither achieves the lowest corrupted error on any dataset. Thus, clean accuracy alone does not reliably predict corrupted performance, relative degradation, or reference-normalized robustness.

\begin{table*}[t]
\centering
\caption{\textbf{Model robustness across datasets and corruption families.}
\emph{Left:} Per-dataset mcMAE at FP32/INT8, followed by median RD and mean mCE for each evaluated precision.
\emph{Right:} FP32 RD by family: environment (Env), optics/motion (Opt), sensor/exposure (Sen), encoding/transport (Enc), clutter (Clu), and visibility loss (Vis).
Lower is better. \textbf{Bold} indicates the best model in each left-hand column and each model's worst-performing family on the right.}
\label{tab:model-robustness}
\fontsize{6}{7.2}\selectfont
\renewcommand{\arraystretch}{0.88}
\setlength{\tabcolsep}{2.6pt}
\begin{tabular}{@{}l | r r r r | r r | r r | r r r r r r@{}}
\toprule
& \multicolumn{4}{c|}{\mcmae\ (FP32/INT8)} & \multicolumn{2}{c|}{FP32}
& \multicolumn{2}{c|}{INT8}
& \multicolumn{6}{c}{FP32 RD\,\% by corruption family} \\
\cmidrule(lr){2-5}\cmidrule(lr){6-7}\cmidrule(lr){8-9}\cmidrule(l){10-15}
Model & SHA & SHB & QNRF & JHU & RD\,\% & mCE & RD\,\% & mCE & Env & Opt & Sen & Enc & Clu & Vis \\
\midrule
CHF & 74.8/73.6 & 17.4/16.8 & 107.0/107.0 & 72.2/72.8 & 28.9 & 0.821 & 28.5 & 0.824 & 62.6 & 77.5 & \textbf{78.6} & 38.2 & 11.5 & 35.3 \\
DM-Count & 81.0/81.1 & 16.3/15.9 & 110.0/109.5 & 73.5/73.0 & 21.7 & 0.840 & 21.1 & 0.838 & 38.2 & 60.4 & \textbf{68.7} & 32.7 & 12.8 & 22.4 \\
GCHF & 73.4/72.5 & 14.2/14.6 & 104.9/105.9 & 74.7/75.1 & 26.1 & 0.797 & 25.3 & 0.812 & 39.2 & 60.4 & \textbf{65.6} & 33.0 & 18.3 & 33.1 \\
CSRNet & 87.2/87.9 & 26.5/26.1 & 120.8/120.1 & 86.2/85.8 & 27.8 & 1.000 & 26.8 & 1.000 & 66.8 & 71.6 & \textbf{87.9} & 38.0 & 7.8 & 20.5 \\
P2PNet & 80.7/82.1 & 18.7/18.4 & 117.7/118.4 & 77.5/77.2 & 28.1 & 0.887 & 30.1 & 0.892 & 49.9 & 62.7 & \textbf{96.8} & 35.1 & 15.3 & 21.8 \\
APGCC & 75.4/76.1 & 18.5/18.8 & 136.1/131.0 & 80.9/81.2 & 26.8 & 0.914 & 28.4 & 0.912 & 52.0 & 67.2 & \textbf{76.8} & 32.8 & 9.5 & 19.0 \\
STEERER-VGG19 & 79.6/75.9 & 15.0/15.5 & 104.0/106.8 & 72.8/75.6 & 25.9 & 0.810 & 23.5 & 0.821 & 26.5 & 59.5 & \textbf{73.4} & 40.3 & 9.0 & 11.6 \\
FIDTM & 80.4/80.6 & 18.8/17.2 & 129.1/127.9 & 82.8/85.4 & 20.2 & 0.927 & 20.1 & 0.921 & 46.3 & 65.4 & \textbf{71.9} & 31.1 & 6.4 & 31.9 \\
ZIP-Tiny & 78.7/75.6 & 14.4/15.8 & 109.2/106.1 & 79.8/73.1 & 26.5 & 0.841 & 21.8 & 0.817 & 32.2 & \textbf{66.7} & 53.6 & 28.2 & 14.7 & 39.6 \\
STEERER-HRNet & \textbf{68.9}/71.3 & 14.0/13.4 & 93.9/91.7 & 64.0/\textbf{63.9} & 22.3 & \textbf{0.722} & 20.1 & \textbf{0.724} & 31.9 & 61.7 & \textbf{84.0} & 39.9 & 20.7 & 15.4 \\
MobileCount & 108.3/108.0 & 26.7/25.9 & 134.8/135.8 & 186.2/192.3 & 30.4 & 1.367 & 29.8 & 1.385 & 97.9 & 76.1 & \textbf{120.8} & 32.7 & 9.2 & 26.2 \\
ZIP-Nano & 86.1/82.9 & 17.6/19.1 & 135.9/142.7 & 79.2/79.8 & 20.4 & 0.955 & 24.6 & 0.973 & 44.2 & \textbf{56.9} & 47.5 & 26.7 & 16.7 & 31.6 \\
MPCount & 70.9/71.4 & \textbf{12.1}/\textbf{11.9} & 141.2/144.1 & 66.9/67.7 & 22.1 & 0.828 & 24.7 & 0.837 & 28.7 & \textbf{55.4} & 43.5 & 24.5 & 34.8 & 22.2 \\
ZIP-Pico & 100.6/92.4 & 19.7/23.4 & 133.9/138.5 & 85.0/95.8 & 22.9 & 1.024 & 21.7 & 1.079 & \textbf{54.6} & 53.5 & 54.2 & 21.5 & 21.2 & 23.2 \\
ZIP-Small & 77.8/72.2 & 12.3/14.2 & \textbf{88.4}/\textbf{91.4} & 67.2/65.9 & 16.7 & 0.744 & 17.6 & 0.761 & 22.5 & \textbf{37.9} & 37.4 & 18.5 & 24.8 & 21.5 \\
FFNet-ConvNeXt-T & 70.2/71.6 & 12.9/13.8 & 99.4/105.8 & 75.4/70.8 & 23.0 & 0.772 & 22.9 & 0.789 & 31.1 & \textbf{54.5} & 40.4 & 39.2 & 18.7 & 34.7 \\
FFNet-MobileNetV3 & 96.0/94.7 & 33.3/29.0 & 133.1/130.7 & 75.4/75.7 & 29.5 & 1.105 & 28.6 & 1.062 & 62.6 & 66.0 & \textbf{97.7} & 48.5 & 11.7 & 45.6 \\
MCNN & 124.5/128.7 & 41.9/52.4 & 210.6/204.9 & 144.0/142.8 & 23.6 & 1.623 & 25.9 & 1.842 & \textbf{86.1} & 67.4 & 57.1 & 35.9 & 12.1 & 14.3 \\
CCST & 72.3/72.0 & 13.8/13.7 & 106.5/107.5 & 67.5/67.9 & \textbf{12.2} & 0.798 & \textbf{12.6} & 0.802 & 23.3 & 21.1 & 19.5 & \textbf{43.3} & 25.2 & 24.3 \\
Transcrowd-GAP & 77.1/78.9 & 15.3/15.3 & 102.0/106.3 & 78.4/77.4 & 15.2 & 0.844 & 15.1 & 0.860 & 25.1 & \textbf{32.8} & 20.7 & 19.4 & 17.1 & 22.2 \\
Transcrowd-Token & 80.4/83.8 & 15.0/14.9 & 102.3/106.1 & 69.2/70.9 & 12.8 & 0.818 & 13.0 & 0.845 & 30.2 & \textbf{41.9} & 20.8 & 17.1 & 17.7 & 22.9 \\
CCTrans-Full & 70.5/73.2 & 12.5/13.7 & 99.5/101.2 & \textbf{63.8}/67.8 & 17.6 & 0.742 & 18.7 & 0.773 & 23.0 & \textbf{36.2} & 30.8 & 26.6 & 21.6 & 21.7 \\
CCTrans-Weak & 85.9/86.6 & 14.5/14.9 & 119.1/115.9 & 68.8/72.3 & 12.5 & 0.866 & 14.9 & 0.876 & 20.3 & \textbf{41.6} & 26.3 & 28.5 & 14.9 & 28.9 \\
Gramformer & 74.7/74.8 & 17.4/17.9 & 106.3/111.0 & 71.8/73.5 & 30.2 & 0.811 & 26.7 & 0.844 & 60.3 & 76.4 & \textbf{94.4} & 38.2 & 20.5 & 22.2 \\
PET & 70.8/\textbf{69.9} & 17.9/18.4 & 114.1/110.9 & 70.9/68.4 & 24.9 & 0.819 & 20.2 & 0.813 & 71.6 & 72.4 & \textbf{100.1} & 30.2 & 14.3 & 21.9 \\
CLIP-EBC-RN50 & 69.4/70.1 & 17.2/17.9 & 116.5/117.0 & 69.6/68.9 & 25.5 & 0.811 & 25.1 & 0.822 & 65.5 & \textbf{88.2} & 78.2 & 35.1 & 19.9 & 25.5 \\
\midrule
\multicolumn{9}{@{}r|}{\textit{Mean over models}} & \textit{45.9} & \textit{58.9} & \textit{63.3} & \textit{32.1} & \textit{16.4} & \textit{25.4} \\
\bottomrule
\end{tabular}
\vspace{-1em}
\end{table*}

The family breakdown in the same table shows that all 26 models exhibit their largest degradation within a capture-chain family. The sensor/exposure and optics/motion families dominate, while scene clutter is least detrimental for most models. The hardest family nevertheless varies across models, motivating evaluation against the corruptions most relevant to each deployment.

\Cref{tab:output-representation} summarizes relative degradation, reference-normalized corrupted error, and the INT8 interaction by output representation. The two direct-count Transcrowd variants achieve the lowest median RD ($14.0\%$) and smallest INT8 interaction ($|\drd|=0.75$ percentage points), while density-map models attain the lowest median mCE ($0.857$). No representation is uniformly superior: direct-count models perform best across the four capture-chain families, whereas localization-oriented models perform best under scene clutter and visibility loss, as detailed by family in the supplement.

\begin{table}[t]
\centering
\caption{\textbf{Robustness by output representation.}
Group medians across models. All metrics are FP32 except $|\drd|$,
which measures the absolute RD change from FP32 to INT8.
Lower is better.}
\label{tab:output-representation}
\scriptsize
\renewcommand{\arraystretch}{0.95}
\setlength{\tabcolsep}{5.0pt}
\begin{tabular}{@{}l r r r r@{}}
\toprule
Output representation & $n$ & RD (\%) & mCE & $|\drd|$ \\
\midrule
\textbf{D}~Density map & 15 & 23.6 & 0.857 & 2.55 \\
\textbf{B}~Blockwise count map & 5 & 22.9 & 0.916 & 2.74 \\
\textbf{P}~Localization-oriented & 4 & 25.9 & 0.913 & 1.97 \\
\textbf{C}~Direct count & 2 & 14.0 & 0.859 & 0.75 \\
\bottomrule
\end{tabular}
\vspace{-1em}
\end{table}

\smallskip
\noindent\textbf{RQ3: Crowd size and localization fidelity.}

\begin{figure}[t]
\centering
\includegraphics[width=\linewidth]{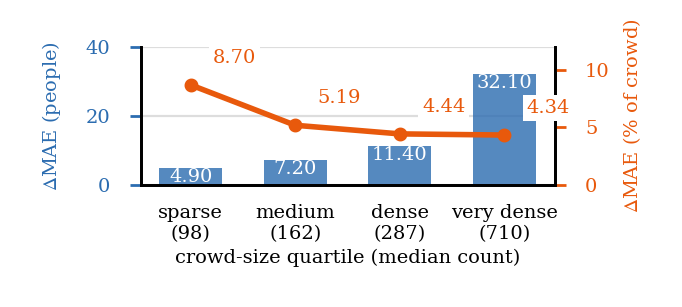}
\caption{\textbf{Corruption error by crowd size.}
Images are grouped by dataset-specific clean-count quartiles.
Parentheses report the median of the four dataset medians.}
\label{fig:crowd-size}
\vspace{-1em}
\end{figure}

\noindent We stratify each dataset into quartiles of clean ground-truth count, so ``sparse'' and ``dense'' remain dataset-specific. Quartile thresholds are 211, 308, and 501 people for ShanghaiTech~A; 55, 93, and 167 for ShanghaiTech~B; 242, 434, and 880 for UCF-QNRF; and 46, 118, and 299 for JHU-Crowd++. In \cref{fig:crowd-size}, the count beneath each quartile denotes the median of the four dataset-specific medians: 98, 162, 287, and 710 people. Across these quartiles, the corruption-induced increase in MAE rises from $4.90$ people in the sparsest quartile to $32.10$ in the densest, a $6.55\times$ difference. Normalizing by crowd size reverses this trend, with increases of $8.70$ and $4.34$ percentage points in the sparsest and densest quartiles, respectively. Thus, dense scenes incur larger absolute errors, whereas sparse scenes exhibit greater proportional degradation.

\begin{figure}[t]
\centering
\includegraphics[width=\linewidth]{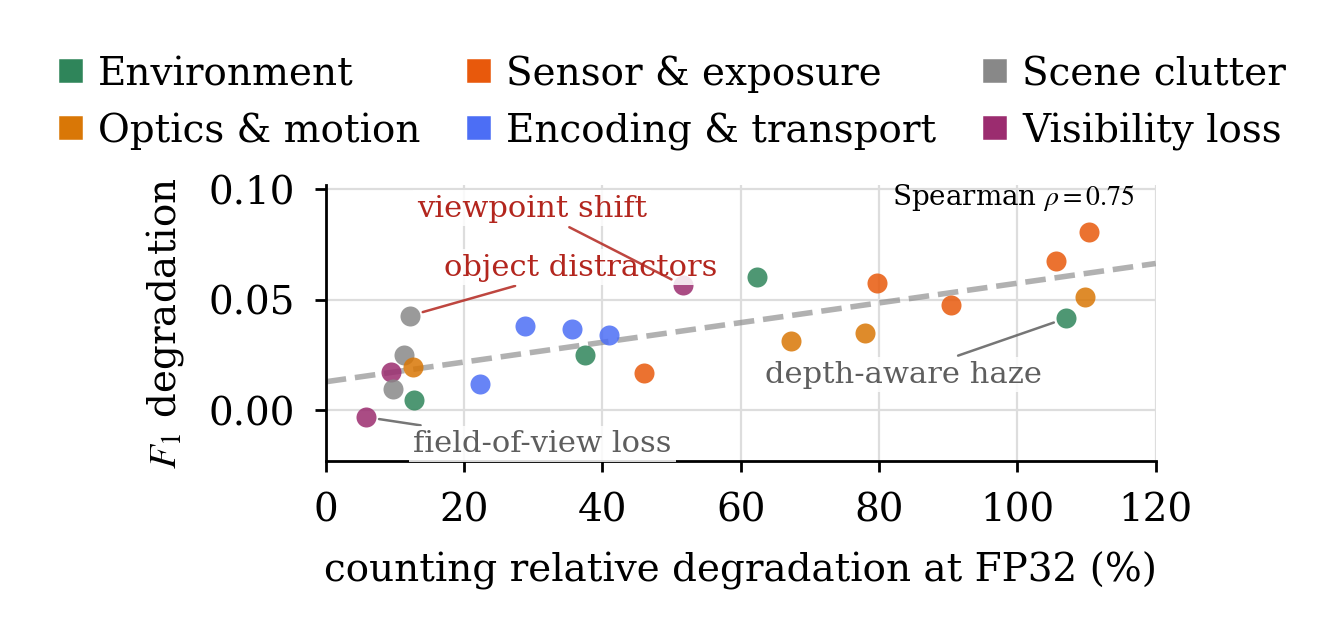}
\caption{\textbf{Counting vs. localization degradation.}
Each point represents an FP32 corruption across four native-point models.
The dashed line shows the least-squares fit. Labels identify the largest deviations above and below it.}

\label{fig:count-vs-loc}
\vspace{-1em}
\end{figure}

Count-based error does not capture localization fidelity when the predicted total is preserved. The four models with native point outputs allow these effects to be examined separately. Across corruptions, counting and localization degradation are correlated but distinct (\cref{fig:count-vs-loc}): object distractors and viewpoint shift affect localization more strongly, whereas field-of-view loss and depth-aware haze show the opposite trend. At FP32, recall decreases by $0.052$ on average, compared with $0.013$ for precision, indicating that corruption mostly suppresses valid detections rather than introducing spurious localizations.

\smallskip
\noindent\textbf{RQ4: Quantization as a model--engine interaction.}

\noindent All three precisions use TensorRT engines; FP16 serves as a reduced-precision reference for assessing the additional changes associated with INT8. Across 104 paired model--dataset comparisons, the median absolute change in relative degradation ($|\drd|$) is $0.04$ percentage points for FP16 and $1.91$ for INT8, with INT8 deviating more in 102 cases. FP16 changes cluster near zero (95th percentile: $0.39$ percentage points), a threshold exceeded by 91 INT8 comparisons. The four FP16 changes above one percentage point all occur on ShanghaiTech~B, where the small clean-error denominator amplifies relative measures. Thus, FP16 largely preserves the FP32 corruption response, whereas INT8 produces larger shifts under the evaluated engine configurations.

\begin{figure}[t]
\centering
\begin{subfigure}{\linewidth}
\centering
\includegraphics[width=0.85\linewidth]{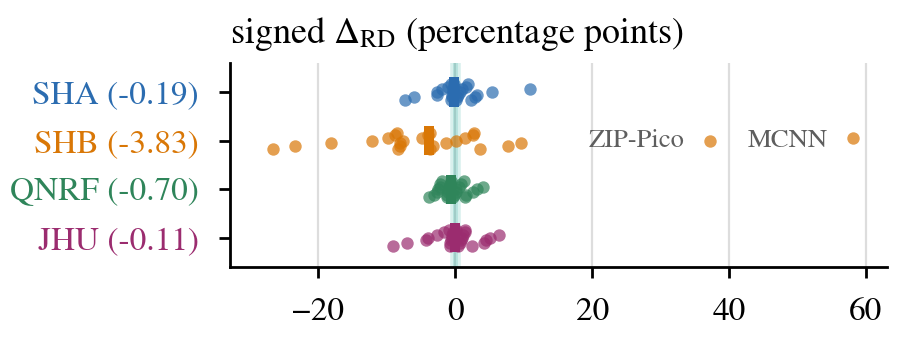}
\caption{Direction, by dataset.}
\label{fig:precision-engine-a}
\end{subfigure}\\[2pt]
\begin{subfigure}{\linewidth}
\centering
\includegraphics[width=0.85\linewidth]{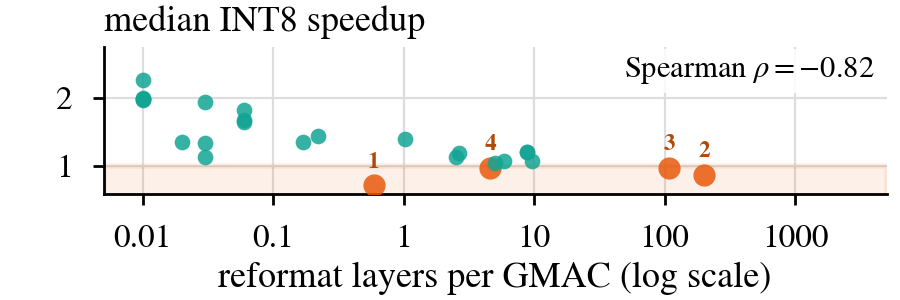}
\caption{Device realization.}
\label{fig:precision-engine-b}
\end{subfigure}
\caption{\textbf{Precision effects and INT8 performance.}
\textbf{(a)} Shading marks the FP16 95th-percentile range ($\pm 0.39$ percentage points).
\textbf{(b)} Values below 1 indicate slowdown.
Labels: 1~MCNN, 2~FFNet-MobileNetV3, 3~FFNet-ConvNeXt-T, and 4~ZIP-Small.}

\label{fig:precision-engine}
\vspace{-1em}
\end{figure}

INT8 has no systematic direction of effect. \Cref{fig:precision-engine}a shows that INT8 decreases RD in 60 comparisons and increases it in 44. An exact two-sided sign test yields $p=0.141$, indicating that this imbalance is not statistically distinguishable from an even split. Aggregating to one median per model yields the same conclusion (16 versus 10; $p=0.327$). The largest shifts occur on ShanghaiTech~B, where the small clean-error denominator amplifies relative changes. Across corruptions, the largest median absolute interactions occur for far-field defocus ($6.01$ percentage points), Gaussian noise ($5.91$), depth-aware haze ($3.86$), and H.264 compression ($3.82$).

Efficiency is also model-dependent. INT8 is slower than FP16 in 17 of 104 device measurements, 13 of which also consume more energy. Four models remain slower after aggregation across datasets. This variation is partly associated with the compiled engine. TensorRT inserts \emph{reformat} layers when adjacent kernels require different tensor layouts or precisions, introducing conversion overhead without useful computation. After normalizing the number of \emph{reformat} layers by model complexity, measured in giga multiply--accumulate operations (GMAC), higher reformat overhead is strongly associated with lower median GPU-time speedup across the 26 models (Spearman $\rho=-0.82$, \cref{fig:precision-engine}b). INT8 slowdowns occur both in models with limited INT8 coverage and in those requiring frequent layout or precision conversions.

\begin{figure}[t]
\centering
\includegraphics[width=0.85\linewidth]{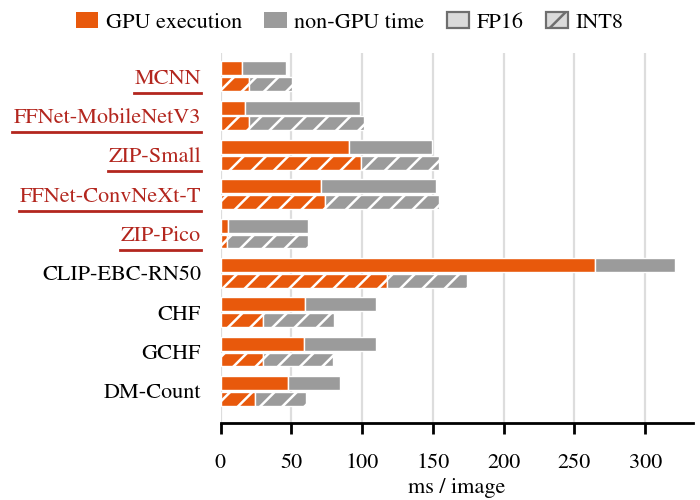}
\caption{\textbf{Runtime decomposition.}
Median end-to-end time per image across four datasets for five models with INT8 slowdown (red) and four faster controls.
Hatched bars are INT8; non-GPU time includes preprocessing, transfers, dispatch, and residual.}

\label{fig:runtime-stages}
\vspace{-1em}
\end{figure}

\Cref{tab:model-engine} reports operations retained above INT8, while \cref{fig:runtime-stages} decomposes end-to-end latency for the five models exhibiting at least one INT8 slowdown and four faster controls. Host preprocessing dominates total latency, limiting the end-to-end benefit of GPU acceleration. When preprocessing is excluded, however, the slowdowns persist in GPU execution rather than the remaining non-GPU time, making image decoding, host--device transfer, and dispatch unlikely explanations. This localizes the regressions to GPU execution but does not establish reformats or limited INT8 coverage as their cause. The associations apply to the evaluated TensorRT configurations and Jetson operating mode.

\smallskip
\noindent\textbf{RQ5: Joint robustness--cost selection.}

\noindent Deployment cost varies by nearly three orders of magnitude on the same device. Across 312 executed configurations (26 models $\times$ 4 datasets $\times$ 3 precisions), energy consumption ranges from $0.20$ to $184.48$\,J/image. The frontiers pair full-test-split robustness from A100 engines with Jetson cost measurements for corresponding graphs and precisions; they do not measure corrupted accuracy on Jetson.

\begin{figure}[t]
\centering
\includegraphics[width=\linewidth]{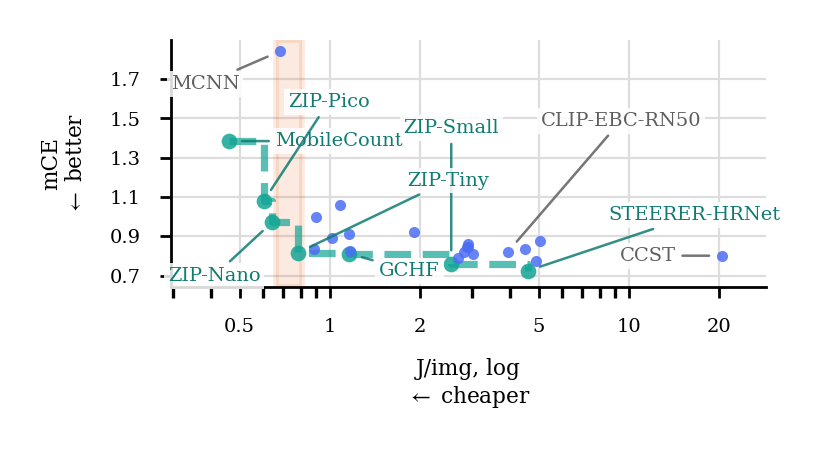}
\caption{\textbf{Robustness--energy Pareto frontier.}
Median INT8 energy on Jetson versus mCE across four datasets.
Shading indicates an equal-cost band; green labels mark Pareto-optimal models.}

\label{fig:frontier}
\vspace{-1em}
\end{figure}

A model is Pareto-undominated if no evaluated alternative achieves lower corrupted error at equal or lower energy, or lower energy at equal or lower corrupted error. \Cref{fig:frontier} identifies seven INT8 models on this frontier: MobileCount, ZIP-Pico, ZIP-Nano, ZIP-Tiny, GCHF, ZIP-Small, and STEERER-HRNet. These models span the cost range, but energy alone does not predict corrupted accuracy: within the shaded equal-cost band, mCE varies by a factor of $2.25$. Similarly, both the smallest model and the model with the largest INT8 speedup are dominated in the energy--mCE space. These results suggest that parameter count, clean error, and nominal precision provide only partial indications of deployment performance.

\begin{figure}[t]
\centering
\includegraphics[width=\linewidth]{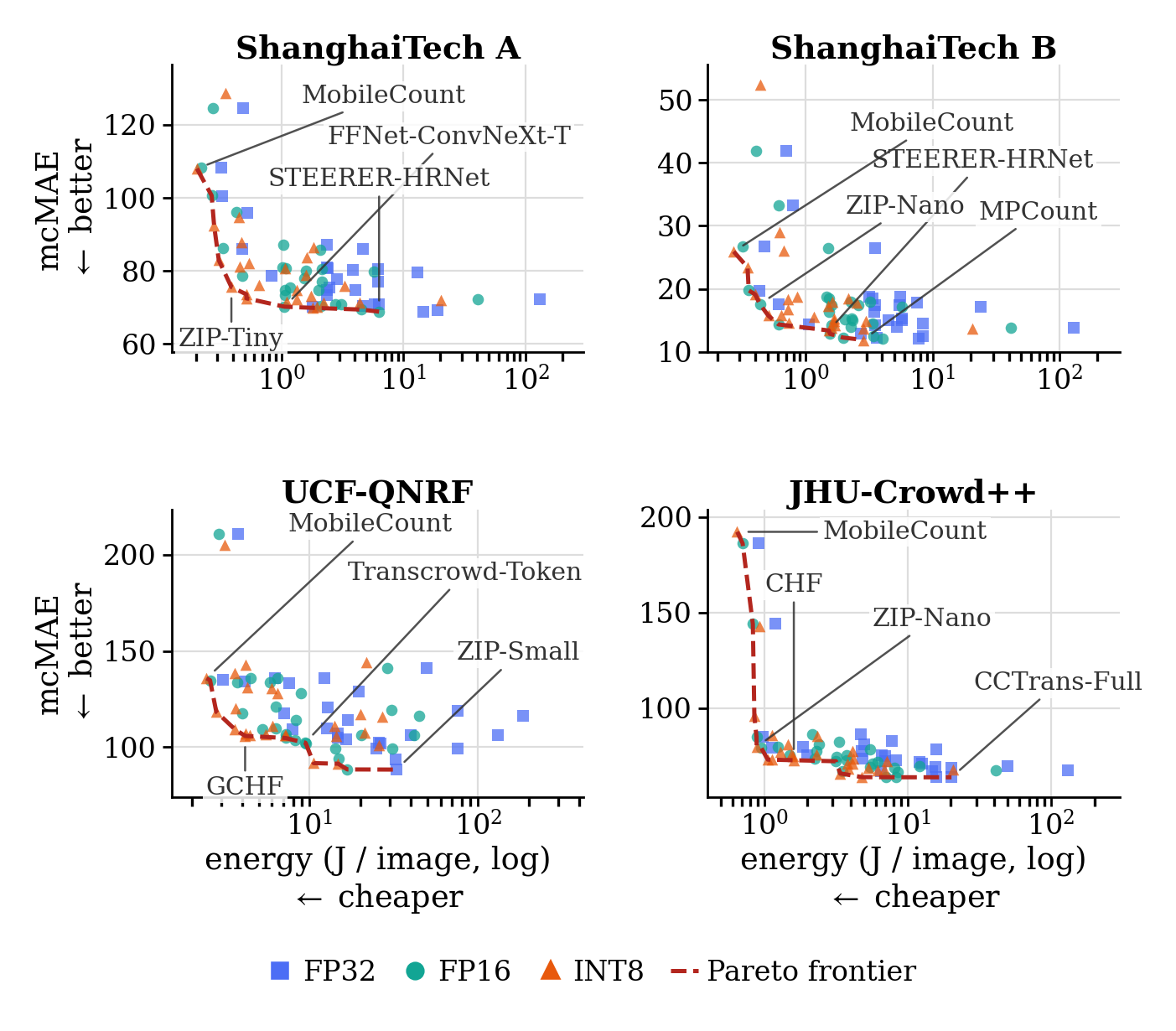}
\caption{\textbf{Accuracy, robustness, and energy trade-offs.}
Each point represents a model--precision configuration, plotted by Jetson energy per image and mean corruption MAE.
The dashed line shows the Pareto frontier.}

\label{fig:frontier-per-dataset}
\vspace{-1em}
\end{figure}

Because deployment occurs under a specific data distribution, \Cref{fig:frontier-per-dataset} evaluates dataset-specific $\mcmae$ across all three precisions. Of the 78 configurations per dataset, 11--15 are Pareto-undominated, representing 8--10 distinct models. Frontier membership varies with both dataset and precision, showing that the aggregate frontier summarizes trade-offs rather than a transferable ranking. MobileCount at INT8 is the lowest-energy configuration on all four datasets, whereas the lowest-error endpoint differs by dataset: STEERER-HRNet on ShanghaiTech~A, MPCount on ShanghaiTech~B, ZIP-Small on UCF-QNRF, and CCTrans-Full on JHU-Crowd++. Intermediate points span efficient, frequency-domain, and transformer architectures, with no single family dominating. Precision affects both energy and $\mcmae$, and the same model may remain Pareto-undominated at multiple precisions. For example, ZIP-Small lies on the UCF-QNRF frontier at INT8, FP16, and FP32. Deployment decisions should therefore account for the expected operating conditions and execution precision.

\section{Discussion}
\label{sec:discussion}
\label{sec:implications}

Our findings show that clean accuracy, corrupted error, numerical precision,
and measured deployment cost capture distinct aspects of model selection.
Precision-dependent Pareto frontiers further highlight the need to consider
these factors jointly.

\noindent\textbf{Synthetic-to-real transfer.}
\suite is a controlled stress test rather than a validation on naturally
degraded imagery. Although rain and snow are calibrated using real-weather
appearance statistics, model rankings and precision effects may not transfer
to real captures. Synthetic generation may introduce model-dependent biases,
and combined degradations such as low light, motion blur, and compression
remain untested.

\noindent\textbf{Deployment scope and attribution.}
Our findings are limited to the evaluated models, datasets, TensorRT
configurations, and Jetson operating mode. A100 robustness and Jetson
deployment cost are paired across corresponding model--precision
configurations, with platform-specific engines built from the same exported
ONNX graphs without additional INT8 recalibration. Associations with output
representation and engine structure remain descriptive rather than causal,
as models differ in architecture, capacity, and training, and the effects
of individual graph operations were not isolated. Small latency and energy differences should be interpreted cautiously, as system-level performance variability can affect comparisons and motivate repeated measurements with uncertainty estimates~\cite{li2015precise,irving2020computer}.
\section{Conclusion}
\label{sec:conclusion}

We introduced \suite, a controlled benchmark for jointly evaluating corruption robustness, numerical precision, and deployment cost in crowd counting. Across 26 models and four datasets, clean accuracy does not reliably predict corrupted performance, while INT8 offers no consistent gains in robustness or efficiency. Precision also changes model rankings and robustness--cost frontiers, highlighting the need for joint model--precision selection. Validation on naturally degraded imagery and other hardware platforms remains future work.

\ifwacvreview\else\ifincludeacknowledgments
  \section*{Declaration of AI-Assisted Technologies}
Generative AI tools, Claude and ChatGPT, were used to assist with coding, LaTeX formatting, and language refinement for manuscript preparation. AI-assisted materials were reviewed and edited by the authors as appropriate for accuracy and factuality.

\section*{Acknowledgments}

This work is supported in part by an NSF Grant 2618897. We appreciate the computational resources provided by the Louisiana Optical Network Infrastructure (LONI) and valuable suggestions from anonymous reviewers.

\fi\fi

    \bibliographystyle{venues/wacv-2027/template/ieeenat_fullname}
    \bibliography{bibliography/references}

\ifincludeappendix
  \maketitlesupplementary
  \appendix

\numberwithin{table}{section}
\numberwithin{figure}{section}

\noindent
This supplement provides the methodological details, complete experimental results, and deployment analyses supporting the main paper. \Cref{sec:supp-benchmark} specifies the 23 corruptions, severity calibration, and annotation handling. \Cref{sec:supp-validation} presents benchmark validity analyses, including severity response, corruption redundancy, and synthetic-weather calibration. \Cref{sec:supp-roster} compares the reproduced clean FP32 accuracy of the evaluated models with published results. \Cref{sec:supp-robustness} provides the complete robustness measurements and additional analyses by output representation, crowd size, and localization performance. \Cref{sec:supp-precision} examines precision-dependent robustness in greater detail while \cref{sec:supp-int8-calibration} documents the INT8 calibration procedure. Finally, \cref{sec:supp-deployment} reports the full edge-deployment measurements and analyzes the runtime and compiled-engine characteristics associated with INT8 performance.

\section{Benchmark Specification}
\label{sec:supp-benchmark}

\subsection{Corruption and severity specification}

\Cref{tab:corruption-taxonomy-full} specifies each corruption, its severity-dependent parameters, underlying mechanism, and annotation policy. Corruptions marked $\dagger$ alter the valid counting target and are therefore evaluated using transformed annotations and kept-annotation error. Several corruptions follow established models of camera or transmission effects, including sensor noise, atmospheric scattering, and block-based video compression. Others---such as distractor content, structured foreground occlusion, camera re-aiming, and depth-dependent defocus---lack standard parametric formulations for crowd imagery and are defined from their underlying physical effect, then calibrated using the calibration procedure described in \cref{sec:supp-calibration}.

\begin{table*}[tp]
\centering
\caption{\textbf{Corruption specifications and severity calibration.}
Generation mechanisms and parameter settings for all 23 corruptions at Light, Moderate, and High severity.
$\downarrow$ indicates parameters that decrease with increasing severity; $^\dagger$ marks corruptions requiring transformed head annotations.}
\label{tab:corruption-taxonomy-full}
\scriptsize
\setlength{\tabcolsep}{2.7pt}
\renewcommand{\arraystretch}{0.9}
\resizebox{\textwidth}{!}{%
\begin{tabular}{@{}l l l ccc c@{}}
\toprule
Corruption & Mechanism & Parameter & Light & Mod. & High & \\
\midrule
\multicolumn{7}{@{}l}{\textbf{Sensor \& exposure}} \\
\quad\texttt{\scriptsize fixed\_pattern\_sensor\_noise} & stuck/hot pixels, column FPN~\cite{healey1994ccd} & column $\sigma$ & 0.002 & 0.008 & 0.012 &  \\
 & & \texttt{\scriptsize stuck\_fraction} & 0.00005 & 0.0002 & 0.0003 &  \\
 & & \texttt{\scriptsize row\_sigma} & 0.002 & 0.008 & 0.012 &  \\
 & & \texttt{\scriptsize temporal\_sigma} & 0.002 & 0.008 & 0.012 &  \\
\quad\texttt{\scriptsize gaussian\_noise} & sensor read noise~\cite{healey1994ccd} & $\sigma$ & 0.010 & 0.025 & 0.050 &  \\
\quad\texttt{\scriptsize highlight\_clipping\_bloom} & overexposure, clipping, veiling glare~\cite{talvala2007glare} & exposure & 2.4 & 3.2 & 3.8 &  \\
 & & \texttt{\scriptsize bloom\_sigma\_frac} & 0.010 & 0.018 & 0.026 &  \\
 & & \texttt{\scriptsize bloom\_gain} & 0.35 & 0.55 & 0.72 &  \\
\quad\texttt{\scriptsize low\_light} & underexposure + photon shot noise~\cite{wei2020noise} & exposure\,$\downarrow$ & 0.6 & 0.4 & 0.2 &  \\
 & & \texttt{\scriptsize photons}\,$\downarrow$ & 1200.0 & 800.0 & 400.0 &  \\
 & & \texttt{\scriptsize read\_sigma} & 0.0015 & 0.0030 & 0.0060 &  \\
\quad\texttt{\scriptsize vignetting\_lens\_shading} & radial illumination falloff~\cite{zheng2009vignetting} & $k$ & 1.25 & 1.80 & 2.70 &  \\
\multicolumn{7}{@{}l}{\textbf{Optics \& motion}} \\
\quad\texttt{\scriptsize camera\_motion\_blur} & camera shake during exposure~\cite{nah2017deep} & length/spacing & 0.05 & 0.15 & 0.25 &  \\
\quad\texttt{\scriptsize far\_field\_defocus} & depth-dependent focus loss & blur/spacing & 0.05 & 0.10 & 0.15 &  \\
 & & \texttt{\scriptsize layers} & 2 & 3 & 4 &  \\
\quad\texttt{\scriptsize lens\_contamination} & droplets/dirt on housing~\cite{you2016raindrop} & drop radius & 0.01 & 0.02 & 0.03 & $\dagger$ \\
 & & \texttt{\scriptsize density}\,$\downarrow$ & 150.0 & 65.0 & 38.0 &  \\
 & & \texttt{\scriptsize radius\_lognormal\_sigma} & 0.3 & 0.4 & 0.5 &  \\
 & & \texttt{\scriptsize refraction\_strength} & 0.3 & 0.4 & 0.5 &  \\
 & & \texttt{\scriptsize blur\_radius\_frac} & 0.15 & 0.20 & 0.25 &  \\
 & & \texttt{\scriptsize brighten} & 0.03 & 0.05 & 0.07 &  \\
 & & \texttt{\scriptsize edge\_feather\_frac}\,$\downarrow$ & 0.20 & 0.15 & 0.10 &  \\
\quad\texttt{\scriptsize lens\_flare\_veiling} & scatter + internal reflection~\cite{wu2021flare} & veil strength & 0.07 & 0.13 & 0.24 &  \\
 & & \texttt{\scriptsize core\_gain} & 0.45 & 0.70 & 1.20 &  \\
 & & \texttt{\scriptsize glow\_radius\_frac} & 0.12 & 0.17 & 0.24 &  \\
 & & \texttt{\scriptsize veil\_radius\_frac} & 0.55 & 0.62 & 0.72 &  \\
 & & \texttt{\scriptsize ghost\_count} & 2 & 3 & 3 &  \\
 & & \texttt{\scriptsize ghost\_strength} & 0.010 & 0.018 & 0.028 &  \\
 & & \texttt{\scriptsize ghost\_radius\_frac} & 0.07 & 0.08 & 0.09 &  \\
\multicolumn{7}{@{}l}{\textbf{Environment}} \\
\quad\texttt{\scriptsize cast\_shadow\_illumination} & hard shadow across scene & coverage & 0.30 & 0.45 & 0.60 &  \\
 & & \texttt{\scriptsize umbra\_transmission}\,$\downarrow$ & 0.62 & 0.38 & 0.20 &  \\
 & & \texttt{\scriptsize penumbra\_frac}\,$\downarrow$ & 0.10 & 0.07 & 0.05 &  \\
 & & \texttt{\scriptsize boundary\_wave\_amp} & 0.10 & 0.13 & 0.16 &  \\
\quad\texttt{\scriptsize depth\_aware\_haze} & Koschmieder scattering by depth~\cite{he2011dehaze} & $\beta$ & 0.18 & 0.32 & 0.55 &  \\
 & & \texttt{\scriptsize airlight} & 0.80 & 0.85 & 0.90 &  \\
\quad\texttt{\scriptsize rain\_streaks} & motion-blurred falling drops~\cite{garg2007rain} & density & 300.0 & 600.0 & 900.0 &  \\
 & & \texttt{\scriptsize length\_frac} & 0.010 & 0.015 & 0.020 &  \\
 & & \texttt{\scriptsize angle\_jitter\_deg} & 5.0 & 6.0 & 7.0 &  \\
 & & \texttt{\scriptsize opacity} & 0.25 & 0.30 & 0.35 &  \\
 & & \texttt{\scriptsize veil} & 0.03 & 0.06 & 0.09 &  \\
\quad\texttt{\scriptsize snow\_particles} & occluding snow, depth-layered & density & 600.0 & 800.0 & 950.0 &  \\
 & & \texttt{\scriptsize size\_frac} & 0.00175 & 0.00220 & 0.00260 &  \\
 & & \texttt{\scriptsize streak\_ratio} & 1.5 & 1.6 & 1.7 &  \\
 & & \texttt{\scriptsize angle\_jitter\_deg} & 14.0 & 15.0 & 16.0 &  \\
 & & \texttt{\scriptsize near\_fraction} & 0.030 & 0.045 & 0.055 &  \\
 & & \texttt{\scriptsize near\_size\_ratio} & 3.5 & 4.0 & 4.5 &  \\
 & & \texttt{\scriptsize near\_blur\_frac} & 0.004 & 0.007 & 0.010 &  \\
 & & \texttt{\scriptsize veil} & 0.08 & 0.13 & 0.17 &  \\
\multicolumn{7}{@{}l}{\textbf{Encoding \& transport}} \\
\quad\texttt{\scriptsize h264\_crf\_compression} & real H.264 encode--decode~\cite{wiegand2003h264} & CRF & 30 & 35 & 40 &  \\
\quad\texttt{\scriptsize moire\_aliasing} & resampling frequency aliasing & sample factor\,$\downarrow$ & 0.70 & 0.60 & 0.50 &  \\
\quad\texttt{\scriptsize packet\_loss\_concealment} & dropped slices, decoder concealment~\cite{wang1998error} & loss rate & 0.0625 & 0.125 & 0.25 & $\dagger$ \\
\quad\texttt{\scriptsize spacing\_resolution\_loss} & effective resolution, in person-spacings~\cite{dong2016srcnn} & spacing frac.\,$\downarrow$ & 0.8 & 0.7 & 0.6 &  \\
\multicolumn{7}{@{}l}{\textbf{Scene clutter}} \\
\quad\texttt{\scriptsize crowd\_like\_distractors} & person-like non-people added & count & 5 & 15 & 30 & $\dagger$ \\
 & & \texttt{\scriptsize height\_spacing\_frac}\,$\downarrow$ & 3.5 & 3.0 & 2.5 &  \\
\quad\texttt{\scriptsize object\_distractors} & non-person clutter added & count/GT & 0.2 & 0.4 & 0.6 & $\dagger$ \\
\quad\texttt{\scriptsize texture\_distractors} & crowd-textured regions added & coverage & 0.15 & 0.28 & 0.45 &  \\
\multicolumn{7}{@{}l}{\textbf{Visibility loss}} \\
\quad\texttt{\scriptsize fov\_loss\_camera\_reaim} & field of view cropped away & crop frac. & 0.15 & 0.30 & 0.45 & $\dagger$ \\
 & & \texttt{\scriptsize direction} & right & corner & corner &  \\
\quad\texttt{\scriptsize structured\_foreground\_occlusion} & poles, railings, banners & coverage & 0.06 & 0.15 & 0.28 & $\dagger$ \\
 & & \texttt{\scriptsize occluder\_darkness}\,$\downarrow$ & 0.18 & 0.16 & 0.14 &  \\
 & & \texttt{\scriptsize occluder\_mix} & pole, railing, banner & + umbrella & + umbrella, spectator &  \\
\quad\texttt{\scriptsize viewpoint\_perspective\_shift} & camera re-aim, perspective warp & top compr. & 0.15 & 0.30 & 0.45 & $\dagger$ \\
\bottomrule
\end{tabular}}

\end{table*}

\subsection{Severity calibration}
\label{sec:supp-calibration}

Severity levels are calibrated through four complementary procedures. First, corruptions governed by standard acquisition or transmission controls use interpretable generation parameters, such as the exposure multiplier for low light and the CRF for H.264 compression. Second, effects whose apparent scale depends on crowd geometry are expressed relative to each image's median head spacing (\cref{sec:supp-spacing}), converting quantities such as motion-blur length and effective resolution into image-specific pixel values. Third, rain and snow are calibrated against appearance statistics of labeled real-weather images from the JHU-Crowd++ training split (\cref{sec:supp-weather}). Fourth, a structured review of rendered training images verifies that head annotations remain supported by visible evidence. Light, Moderate, and High order intensity within each corruption; they do not equate perceptual strength or counting difficulty across corruption types.

\subsection{Spacing-relative severities}
\label{sec:supp-spacing}

The effects of blur, resolution loss, and distractor scale
depend on crowd geometry. For six corruptions, severity
parameters are expressed relative to each image's median
head spacing, defined as the median, over annotated heads,
of their mean distance to the three nearest annotated
neighbors. This adapts corruption magnitudes to each
image's crowd geometry. For resolution loss, the target
spacing is additionally capped at the dataset's training-split
median. \Cref{tab:spacing-pixels} reports representative
pixel magnitudes for four of these corruptions, evaluated
at each dataset's median head spacing; actual magnitudes
vary across images.

\begin{table}[t]
\centering
\caption{\textbf{Spacing-relative severity calibration.}
Parameters are expressed relative to each image's median head spacing. The table reports representative pixel magnitudes evaluated at each dataset's median head spacing.}
\label{tab:spacing-pixels}
\scriptsize
\renewcommand{\arraystretch}{0.92}
\setlength{\tabcolsep}{3.4pt}
\begin{tabular}{@{}l c r rrrr@{}}
\toprule
 & & & \multicolumn{4}{c}{magnitude per dataset (px)} \\
\cmidrule(l){4-7}
Corruption & Level & spacing scale & SHA & SHB & QNRF & JHU \\
\midrule
\multicolumn{3}{@{}l}{\textit{median head spacing} (px)} & 17.23 & 33.11 & 34.78 & 32.99 \\
\midrule
\quad camera motion blur & L & 0.050 & 0.9 & 1.7 & 1.7 & 1.6 \\
\quad & M & 0.150 & 2.6 & 5.0 & 5.2 & 4.9 \\
\quad & H & 0.250 & 4.3 & 8.3 & 8.7 & 8.2 \\
\quad far-field defocus & L & 0.050 & 0.9 & 1.7 & 1.7 & 1.6 \\
\quad & M & 0.100 & 1.7 & 3.3 & 3.5 & 3.3 \\
\quad & H & 0.150 & 2.6 & 5.0 & 5.2 & 4.9 \\
\quad head-spacing loss & L & 0.800 & 13.8 & 26.5 & 27.8 & 26.4 \\
\quad & M & 0.700 & 12.1 & 23.2 & 24.3 & 23.1 \\
\quad & H & 0.600 & 10.3 & 19.9 & 20.9 & 19.8 \\
\quad crowd-like distractors & L & 3.500 & 60.3 & 115.9 & 121.7 & 115.5 \\
\quad & M & 3.000 & 51.7 & 99.3 & 104.3 & 99.0 \\
\quad & H & 2.500 & 43.1 & 82.8 & 87.0 & 82.5 \\
\bottomrule
\end{tabular}
\vspace{-0.7em}
\end{table}

\subsection{Annotation integrity}

Seven corruptions alter the valid counting target by cropping,
warping, or occluding regions containing annotated heads.
\Cref{tab:gt-integrity} summarizes the minimum, median, and
maximum fractions of removed head annotations for each
corruption across the evaluated settings.

\begin{table}[htbp]
\centering
\caption{\textbf{Annotation loss under target-altering corruptions.}
Minimum, median, and maximum fractions of head annotations removed by each transformation across all evaluated cells.}
\label{tab:gt-integrity}
\scriptsize
\setlength{\tabcolsep}{3.2pt}
\renewcommand{\arraystretch}{0.92}
\begin{tabular}{@{}l l rrr@{}}
\toprule
& point & \multicolumn{3}{c}{heads removed (\%)} \\
\cmidrule(lr){3-5}
Corruption & rule & min & med & max \\
\midrule
field-of-view loss & crop & 12.3 & 25.1 & 80.3 \\
foreground occlusion & occlusion & 5.5 & 16.2 & 30.7 \\
object distractors & occlusion & 6.3 & 11.9 & 21.5 \\
packet-loss concealment & occlusion & 5.8 & 11.9 & 26.3 \\
crowd-like distractors & occlusion & 3.6 & 10.4 & 29.8 \\
lens contamination & occlusion & 1.3 & 2.8 & 5.0 \\
viewpoint shift & warp & 0.2 & 1.1 & 1.5 \\
\midrule
\emph{all altering} &  & 0.2 & 9.4 & 80.3 \\
\bottomrule
\end{tabular}

\end{table}

\section{Benchmark Validation}
\label{sec:supp-validation}

\subsection{Corruption visualizations}

\Cref{fig:corruption-gallery-full} shows all 23 corruptions at three calibrated severities on the same UCF-QNRF image, enabling direct visual comparison across the suite. The figure illustrates appearance only.

\begin{figure*}[p]
  \centering
  \setlength{\tabcolsep}{1.1pt}
  \renewcommand{\arraystretch}{1.15}
  \begin{tabular}{@{}r ccc @{\hspace{7pt}} r ccc@{}}
  & \scriptsize Light & \scriptsize Moderate & \scriptsize High& & \scriptsize Light & \scriptsize Moderate & \scriptsize High \\[1pt]
\scriptsize Gaussian noise\hspace{2pt} & \raisebox{-0.5\height}{\includegraphics[width=0.116\textwidth]{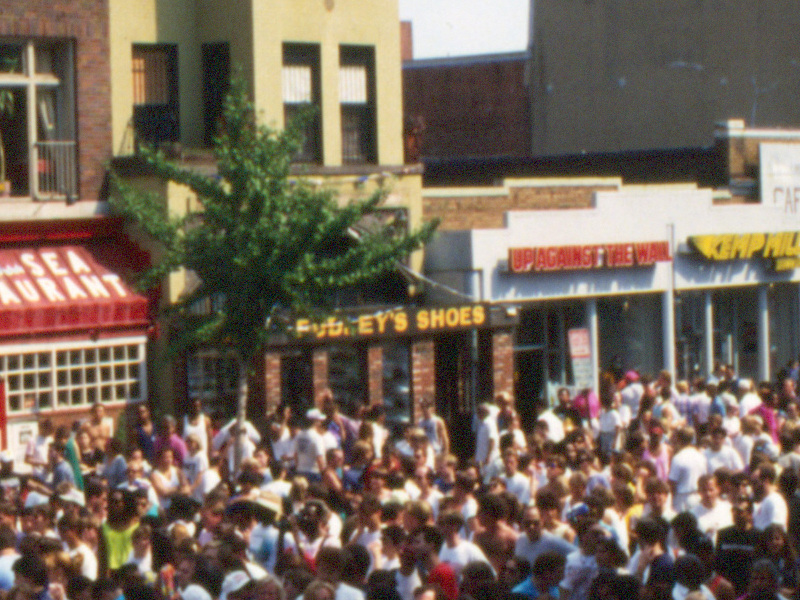}} & \raisebox{-0.5\height}{\includegraphics[width=0.116\textwidth]{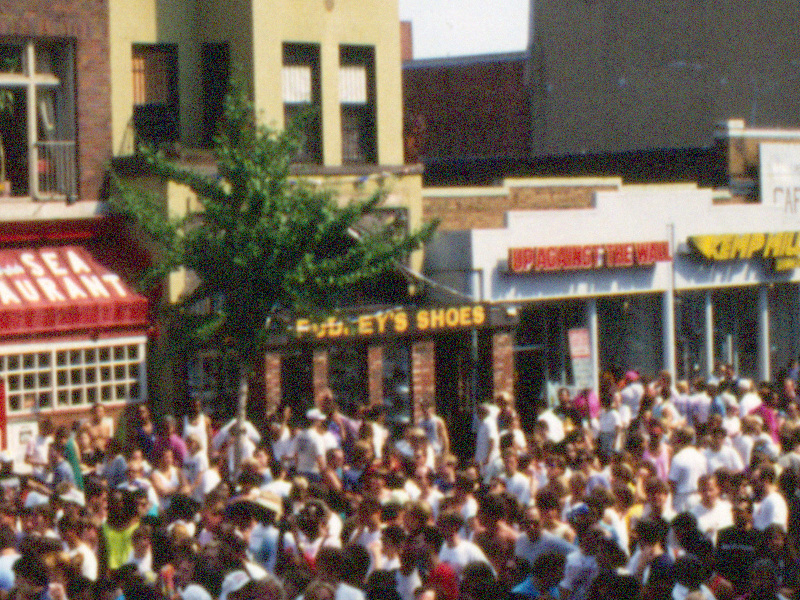}} & \raisebox{-0.5\height}{\includegraphics[width=0.116\textwidth]{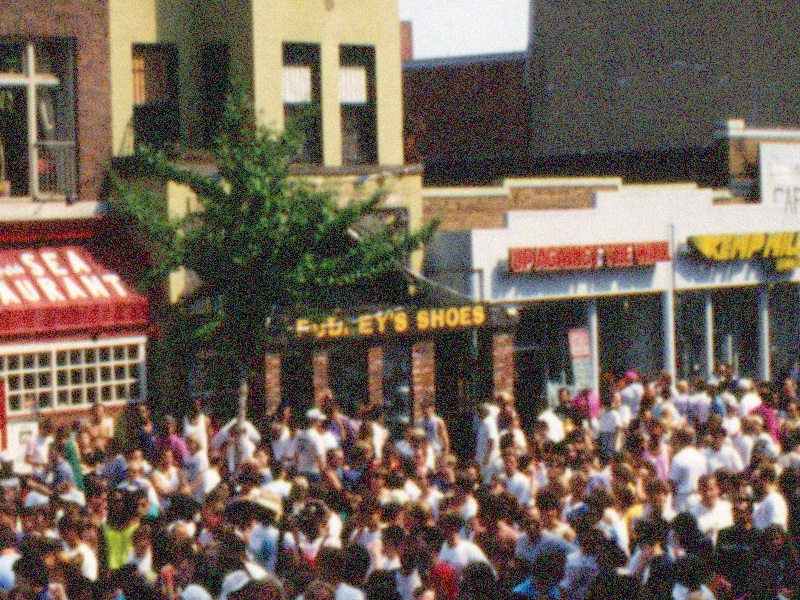}} & \scriptsize cast shadow\hspace{2pt} & \raisebox{-0.5\height}{\includegraphics[width=0.116\textwidth]{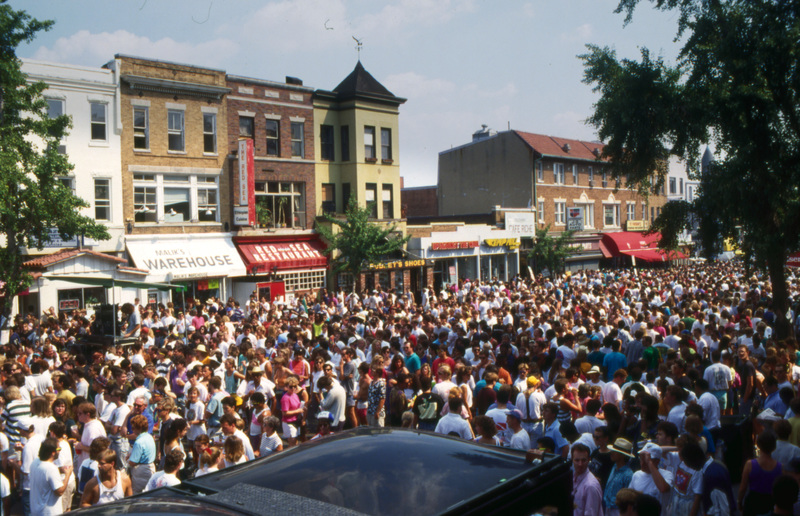}} & \raisebox{-0.5\height}{\includegraphics[width=0.116\textwidth]{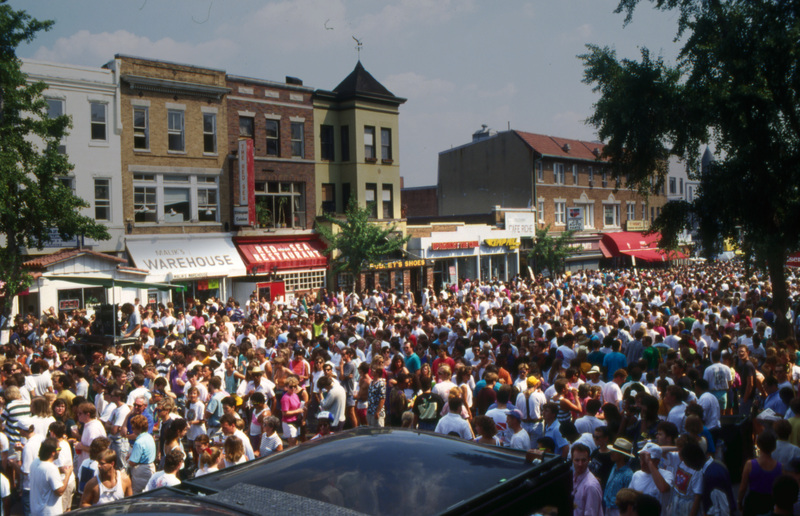}} & \raisebox{-0.5\height}{\includegraphics[width=0.116\textwidth]{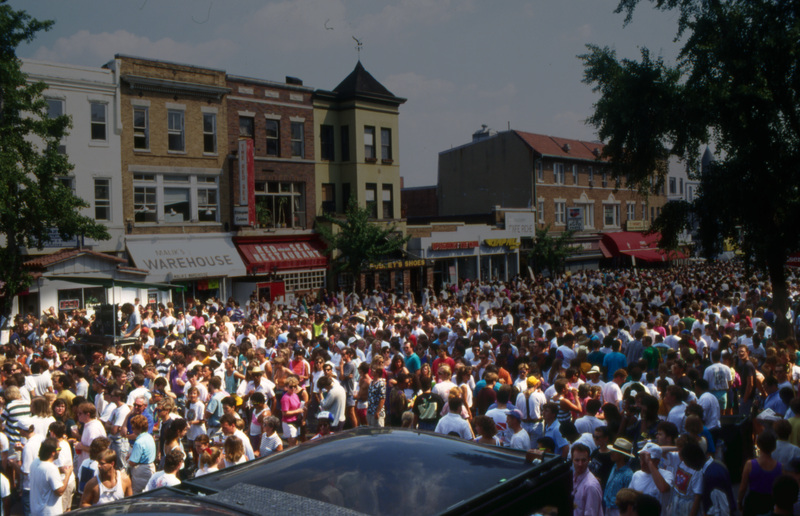}} \\
\scriptsize fixed-pattern noise\hspace{2pt} & \raisebox{-0.5\height}{\includegraphics[width=0.116\textwidth]{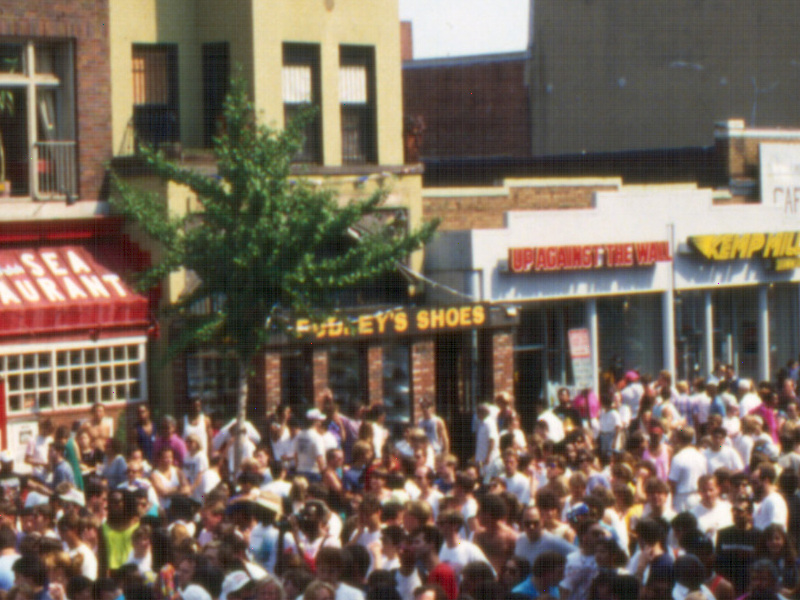}} & \raisebox{-0.5\height}{\includegraphics[width=0.116\textwidth]{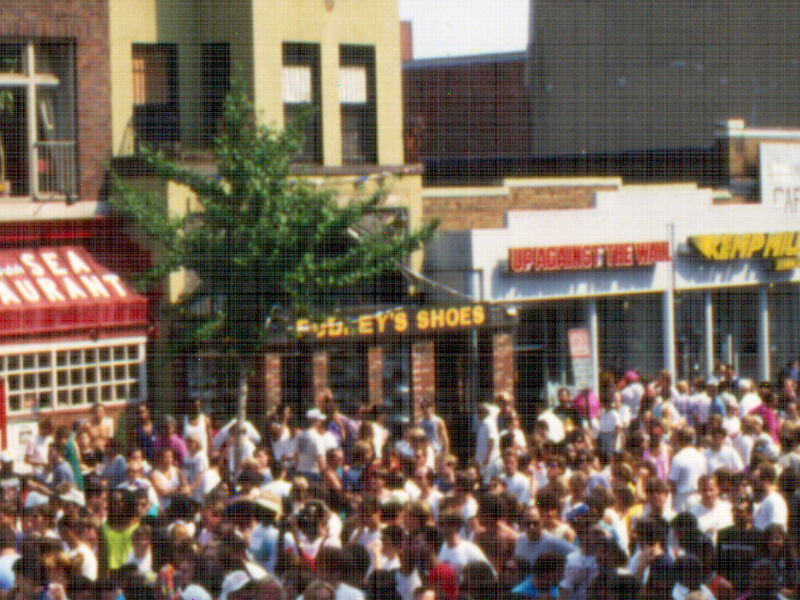}} & \raisebox{-0.5\height}{\includegraphics[width=0.116\textwidth]{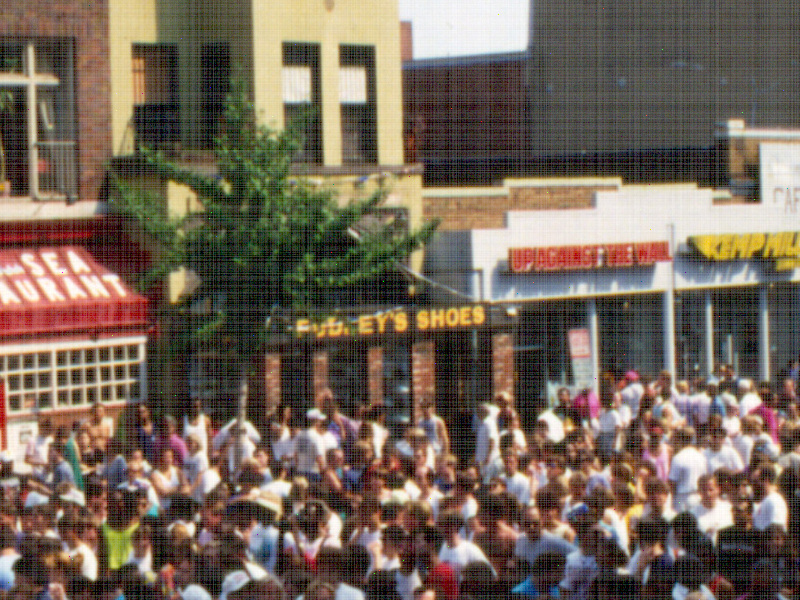}} & \scriptsize H.264 compression\hspace{2pt} & \raisebox{-0.5\height}{\includegraphics[width=0.116\textwidth]{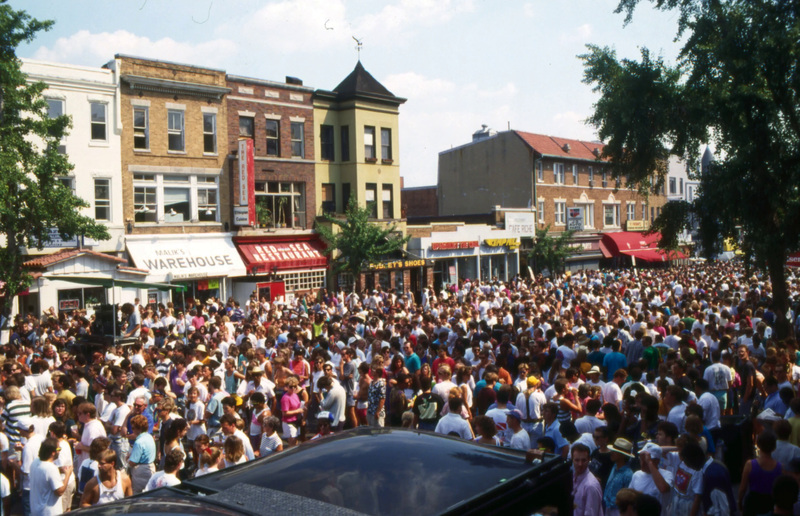}} & \raisebox{-0.5\height}{\includegraphics[width=0.116\textwidth]{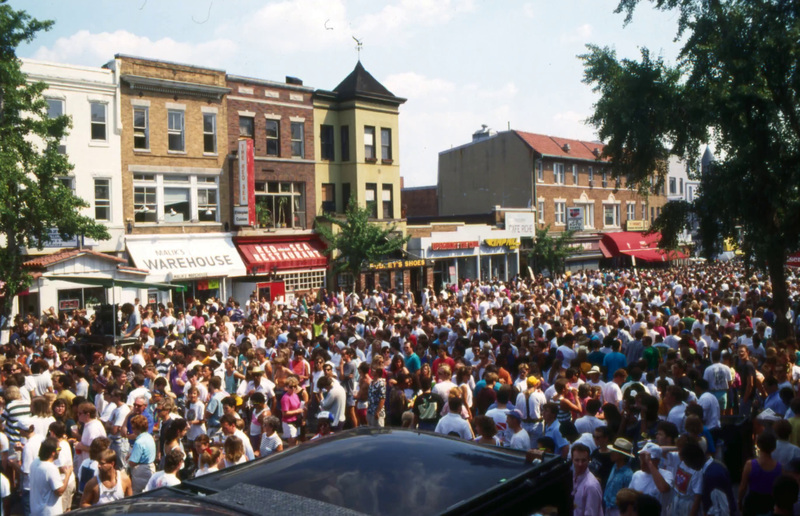}} & \raisebox{-0.5\height}{\includegraphics[width=0.116\textwidth]{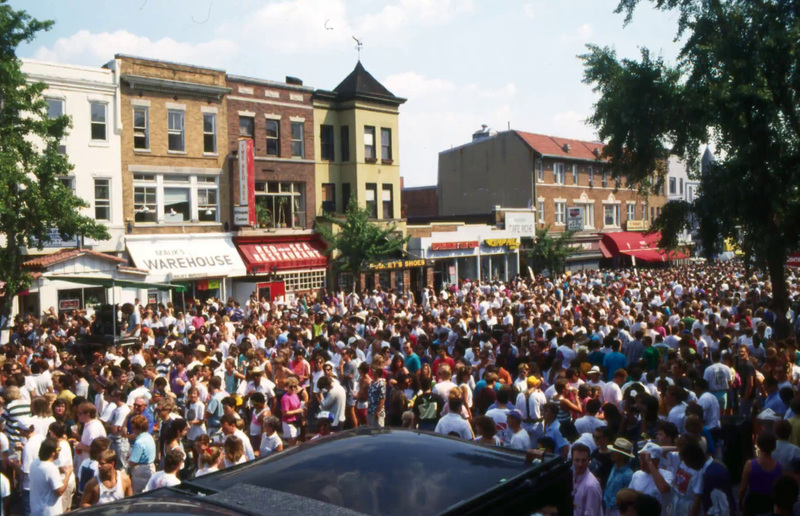}} \\
\scriptsize low light\hspace{2pt} & \raisebox{-0.5\height}{\includegraphics[width=0.116\textwidth]{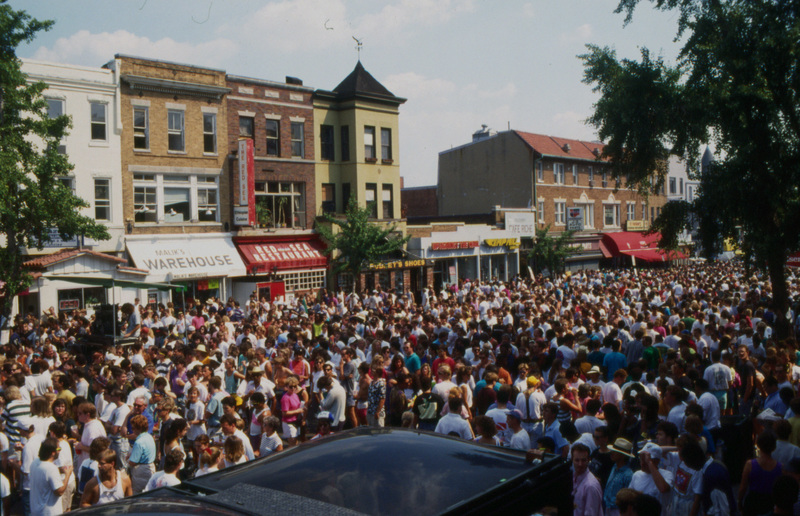}} & \raisebox{-0.5\height}{\includegraphics[width=0.116\textwidth]{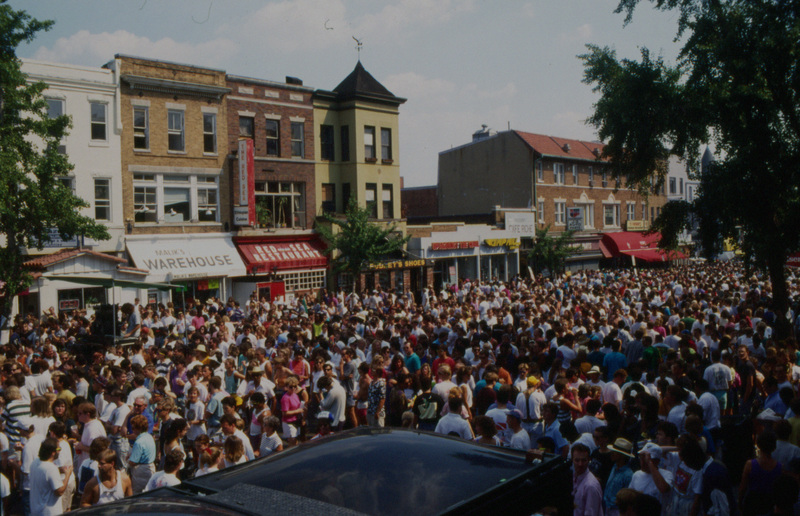}} & \raisebox{-0.5\height}{\includegraphics[width=0.116\textwidth]{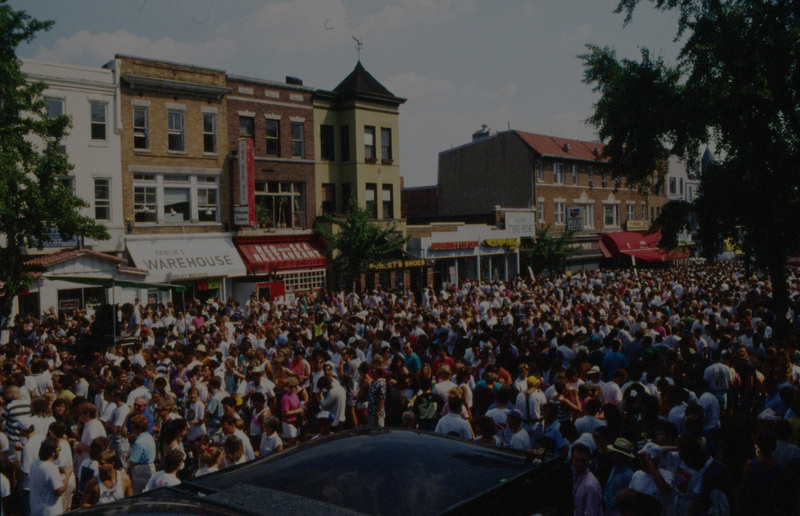}} & \scriptsize moir\'e aliasing\hspace{2pt} & \raisebox{-0.5\height}{\includegraphics[width=0.116\textwidth]{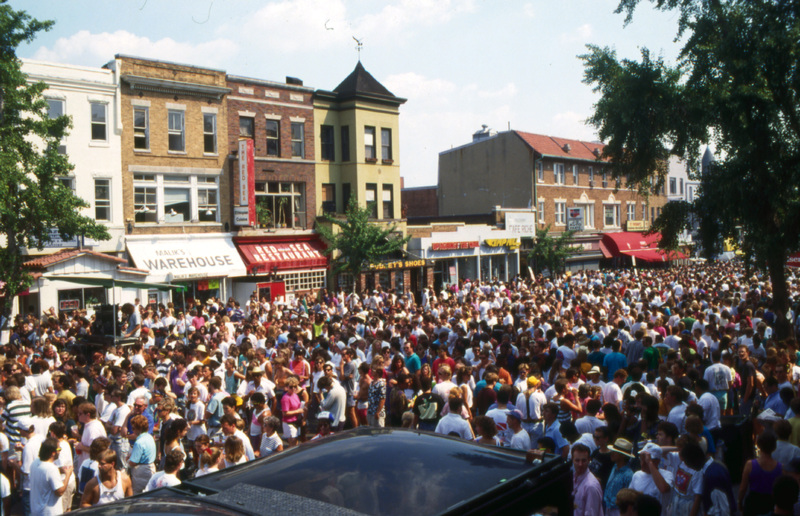}} & \raisebox{-0.5\height}{\includegraphics[width=0.116\textwidth]{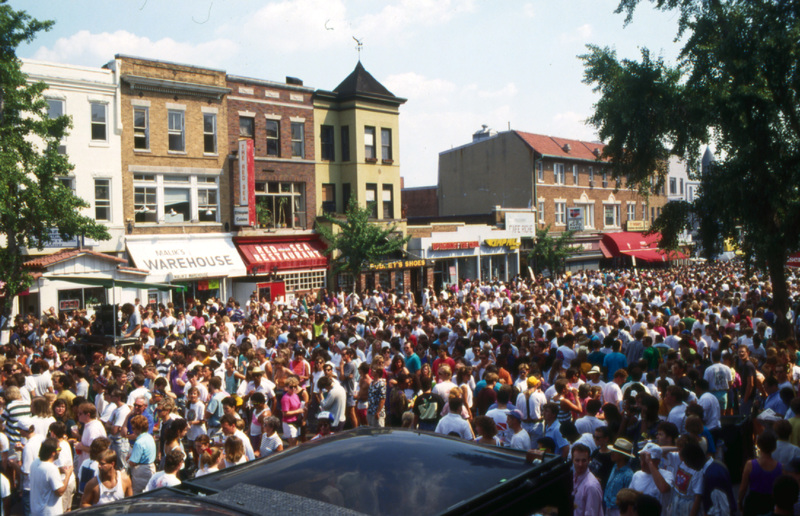}} & \raisebox{-0.5\height}{\includegraphics[width=0.116\textwidth]{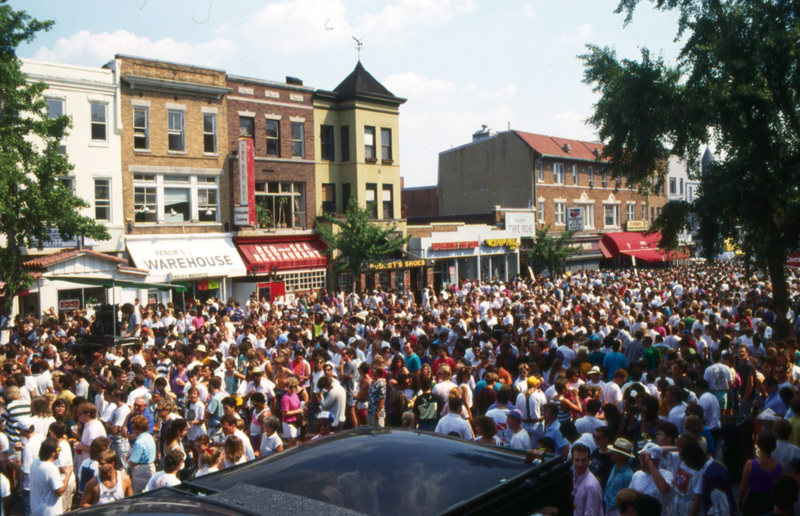}} \\
\scriptsize highlight clipping\hspace{2pt} & \raisebox{-0.5\height}{\includegraphics[width=0.116\textwidth]{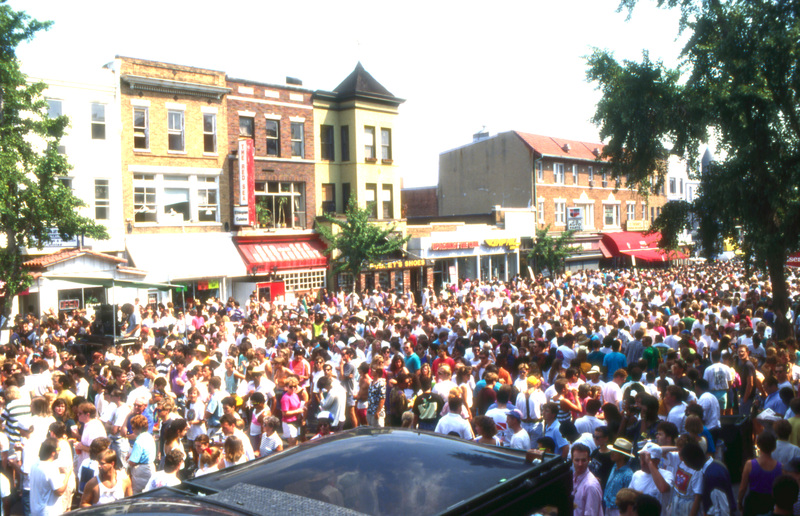}} & \raisebox{-0.5\height}{\includegraphics[width=0.116\textwidth]{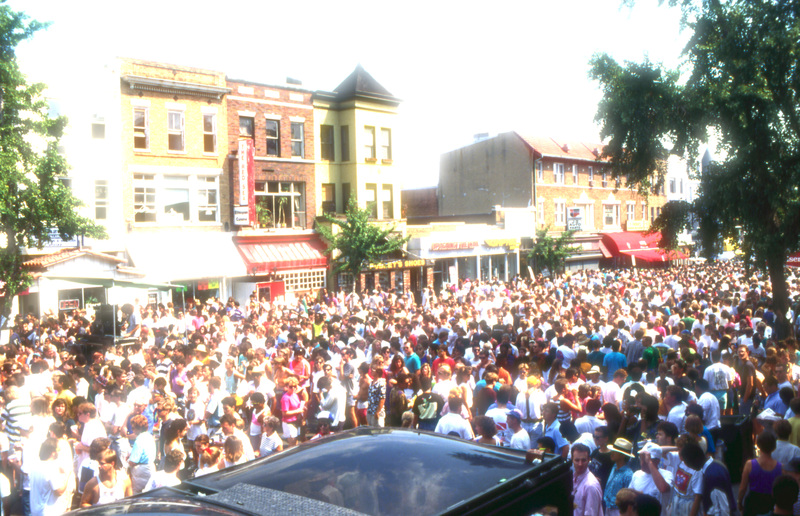}} & \raisebox{-0.5\height}{\includegraphics[width=0.116\textwidth]{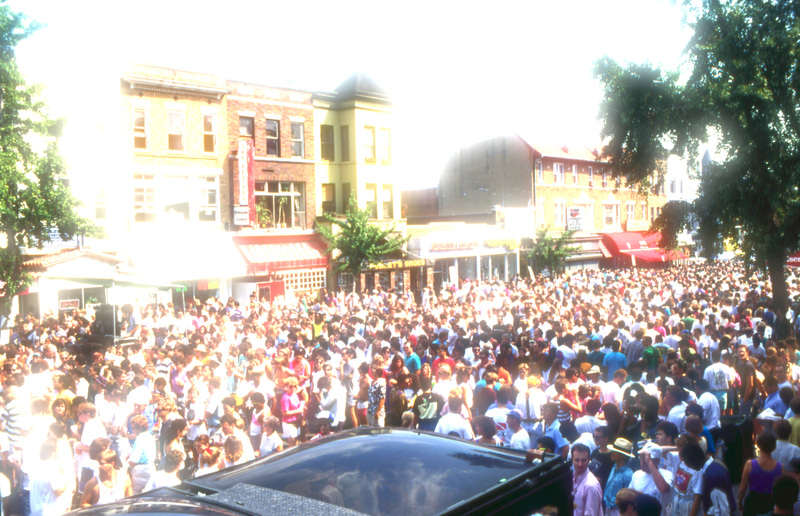}} & \scriptsize head-spacing loss\hspace{2pt} & \raisebox{-0.5\height}{\includegraphics[width=0.116\textwidth]{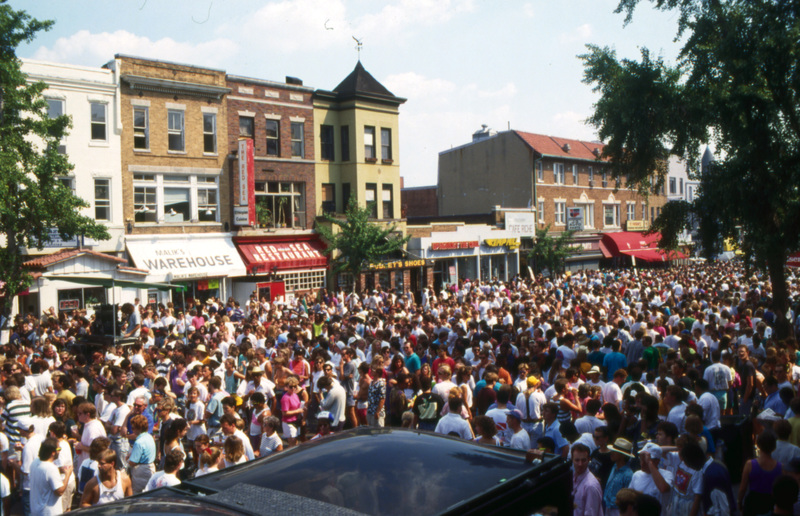}} & \raisebox{-0.5\height}{\includegraphics[width=0.116\textwidth]{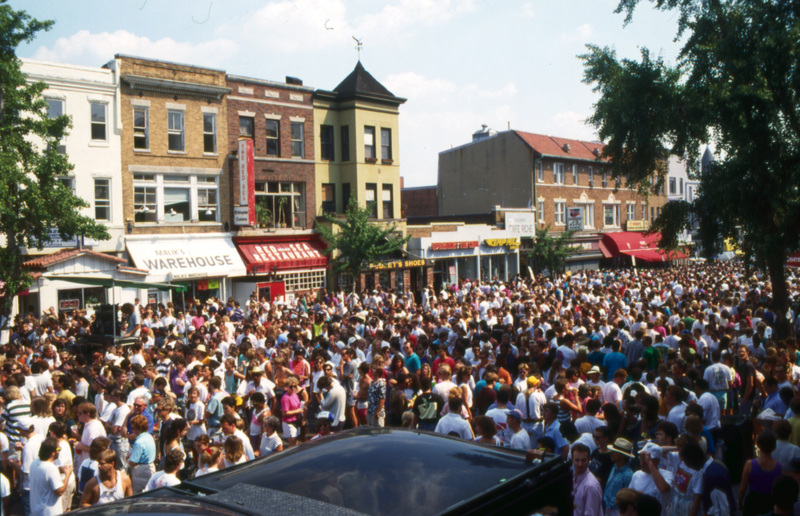}} & \raisebox{-0.5\height}{\includegraphics[width=0.116\textwidth]{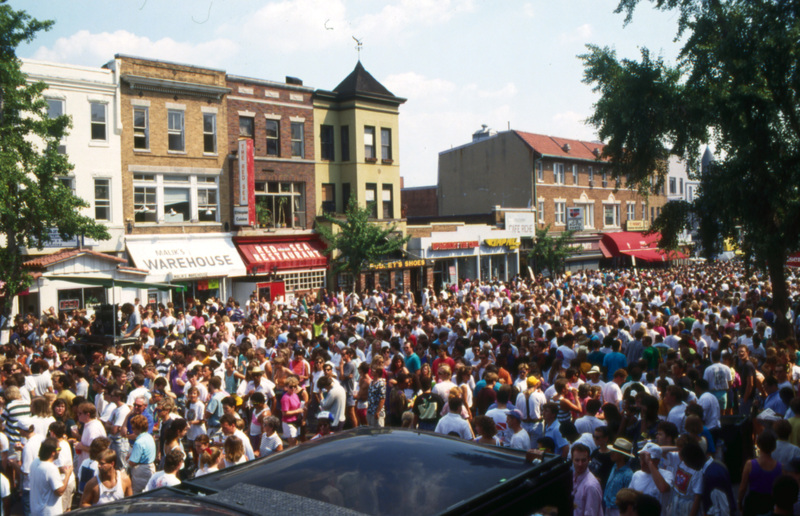}} \\
\scriptsize vignetting\hspace{2pt} & \raisebox{-0.5\height}{\includegraphics[width=0.116\textwidth]{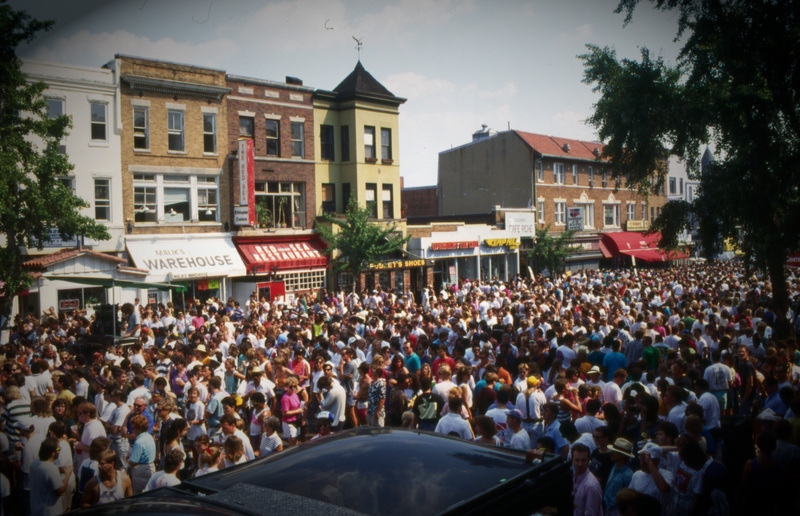}} & \raisebox{-0.5\height}{\includegraphics[width=0.116\textwidth]{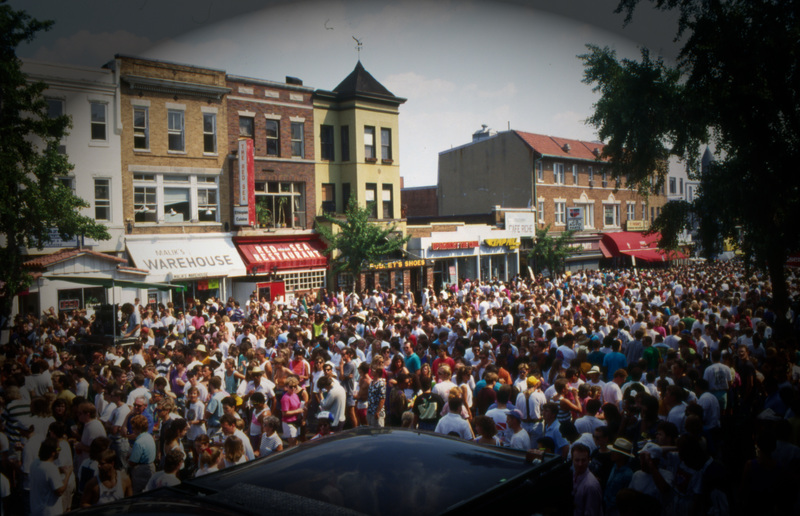}} & \raisebox{-0.5\height}{\includegraphics[width=0.116\textwidth]{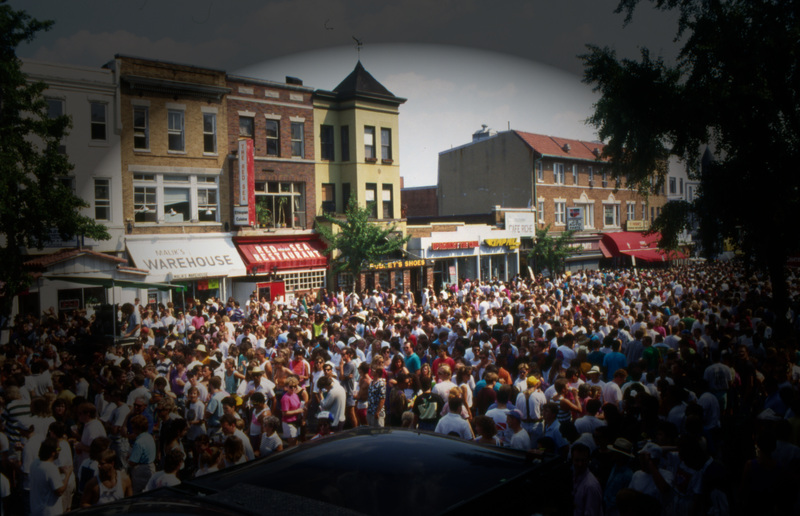}} & \scriptsize packet-loss concealment\hspace{2pt} & \raisebox{-0.5\height}{\includegraphics[width=0.116\textwidth]{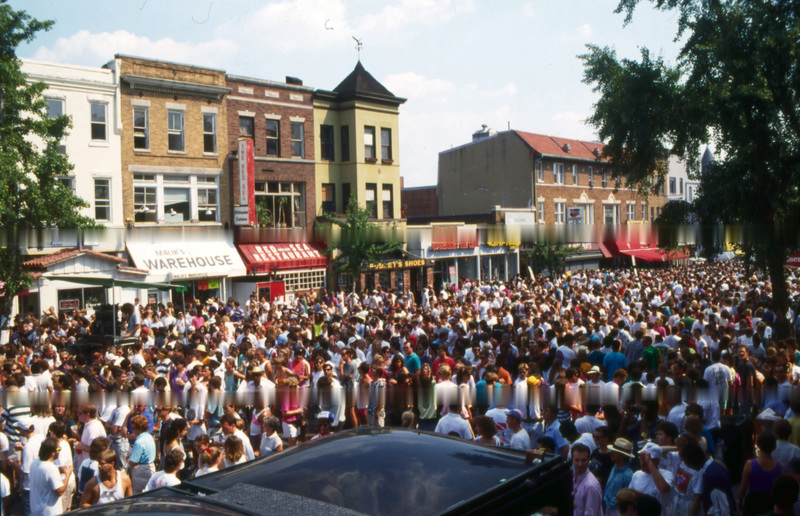}} & \raisebox{-0.5\height}{\includegraphics[width=0.116\textwidth]{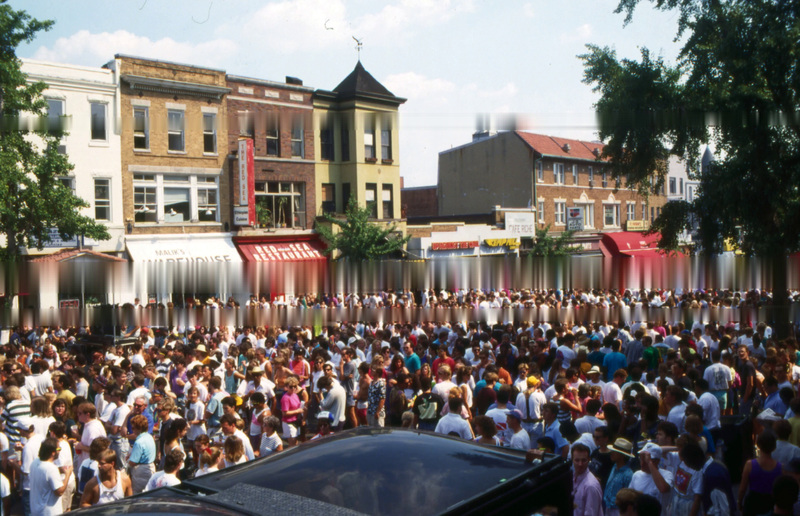}} & \raisebox{-0.5\height}{\includegraphics[width=0.116\textwidth]{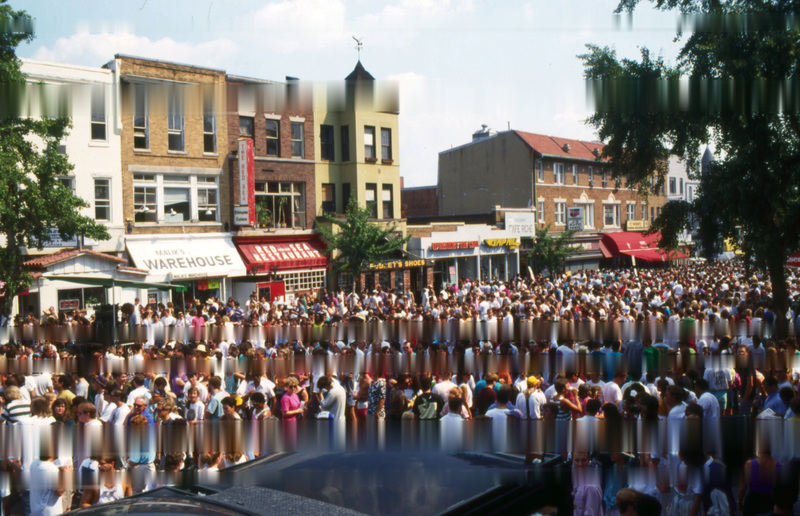}} \\
\scriptsize camera motion blur\hspace{2pt} & \raisebox{-0.5\height}{\includegraphics[width=0.116\textwidth]{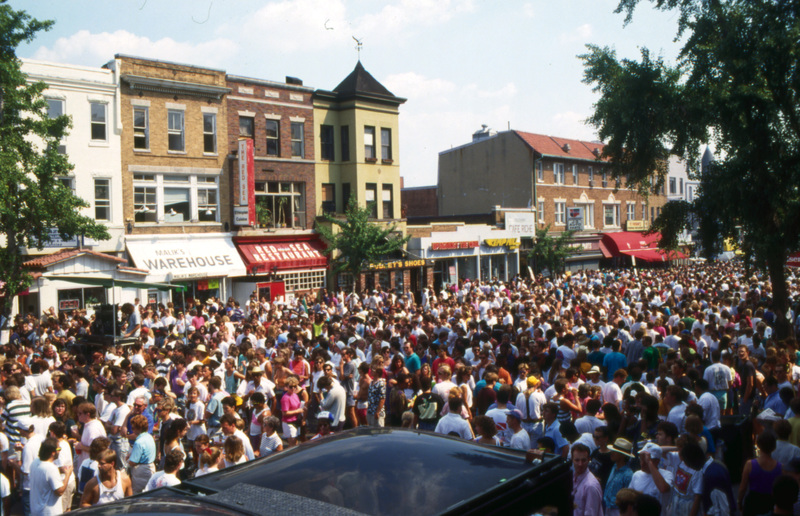}} & \raisebox{-0.5\height}{\includegraphics[width=0.116\textwidth]{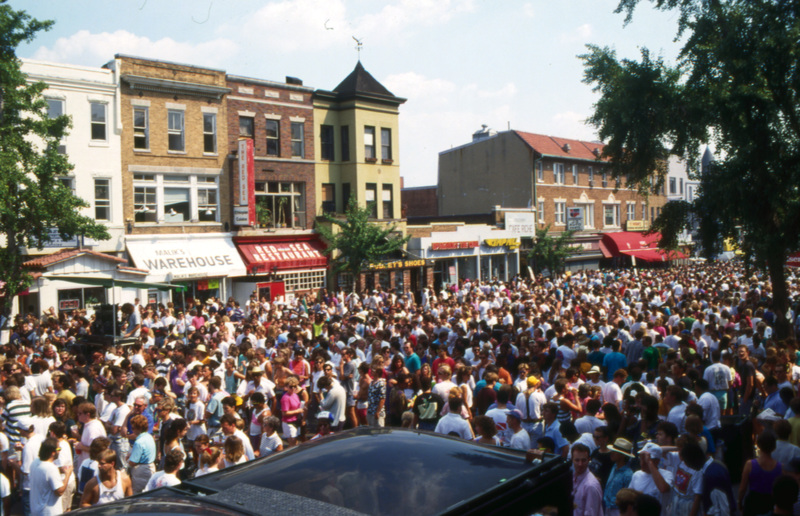}} & \raisebox{-0.5\height}{\includegraphics[width=0.116\textwidth]{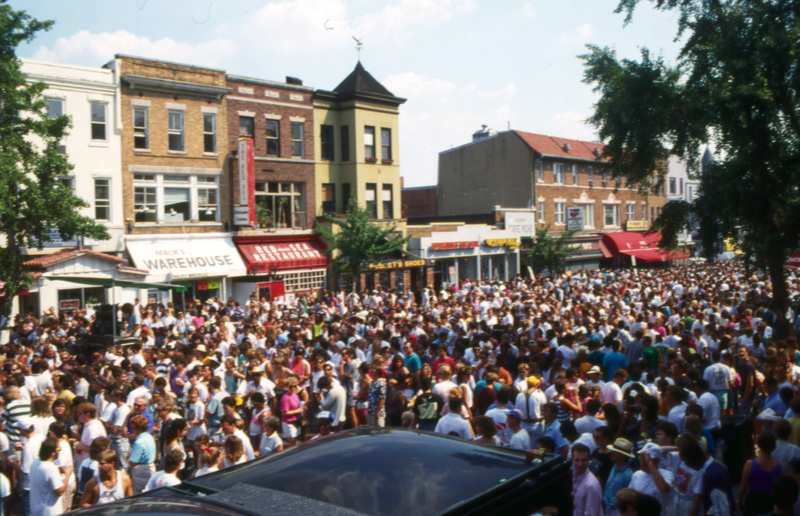}} & \scriptsize texture distractors\hspace{2pt} & \raisebox{-0.5\height}{\includegraphics[width=0.116\textwidth]{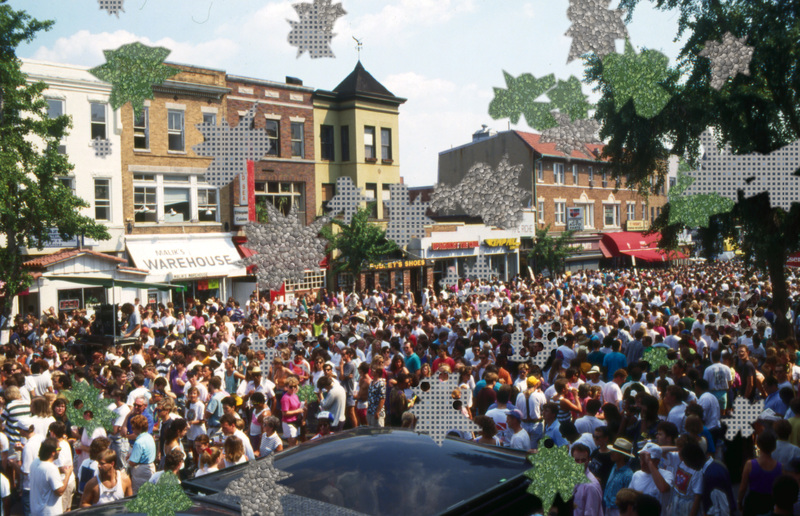}} & \raisebox{-0.5\height}{\includegraphics[width=0.116\textwidth]{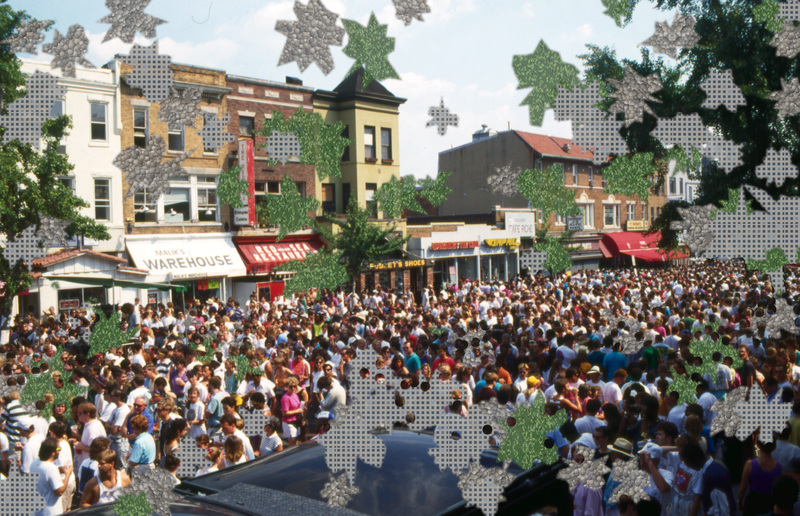}} & \raisebox{-0.5\height}{\includegraphics[width=0.116\textwidth]{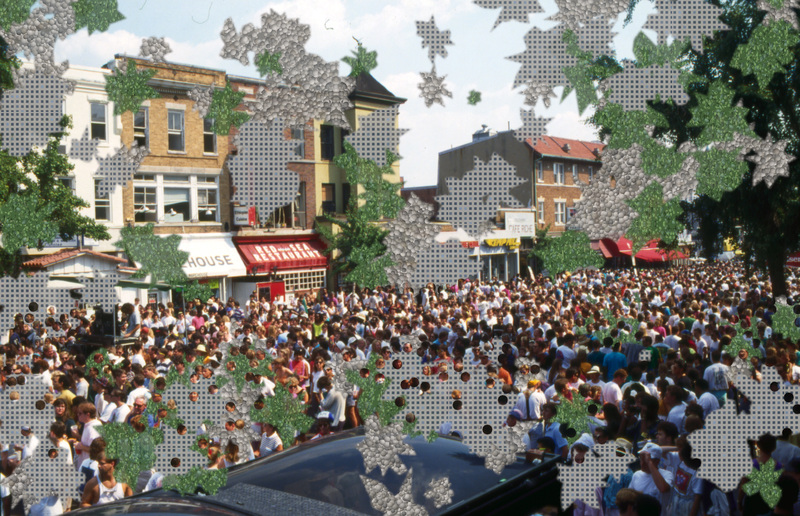}} \\
\scriptsize far-field defocus\hspace{2pt} & \raisebox{-0.5\height}{\includegraphics[width=0.116\textwidth]{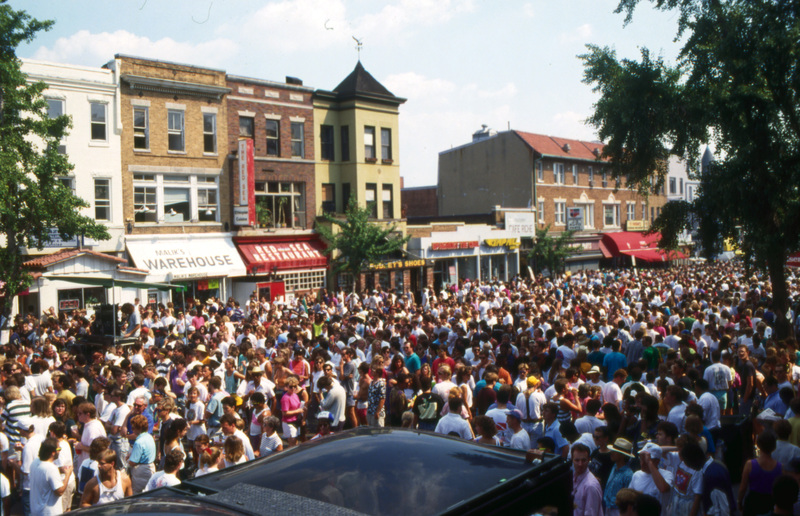}} & \raisebox{-0.5\height}{\includegraphics[width=0.116\textwidth]{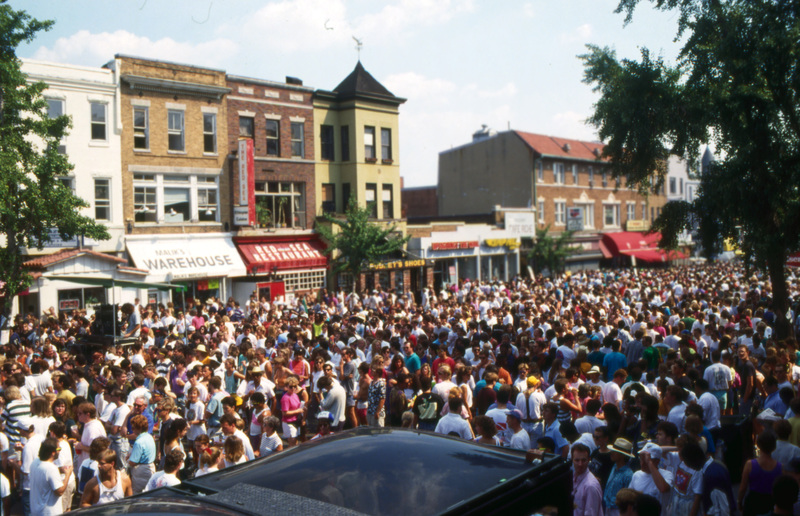}} & \raisebox{-0.5\height}{\includegraphics[width=0.116\textwidth]{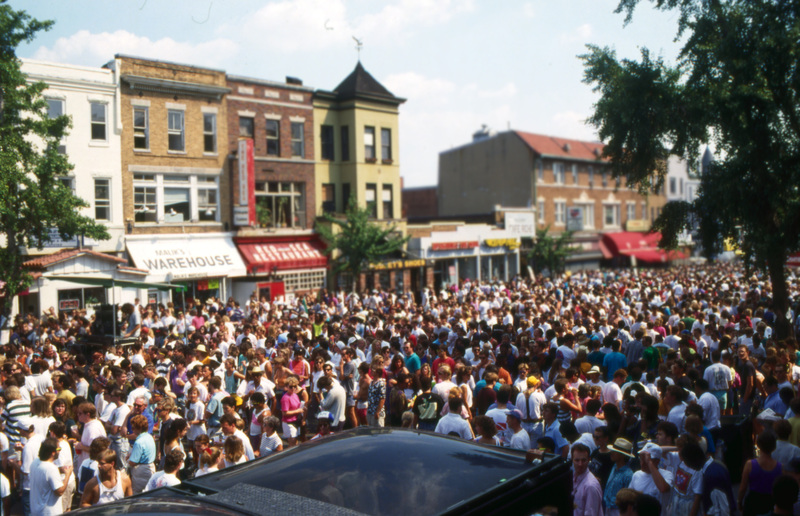}} & \scriptsize crowd-like distractors\hspace{2pt} & \raisebox{-0.5\height}{\includegraphics[width=0.116\textwidth]{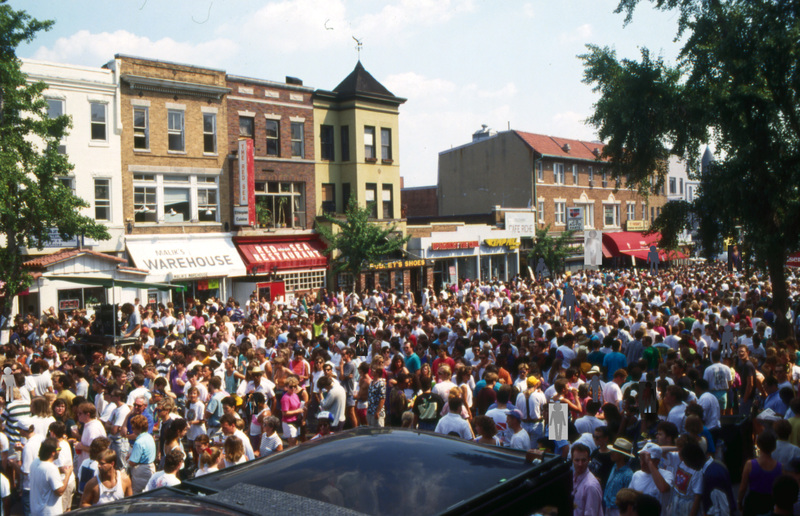}} & \raisebox{-0.5\height}{\includegraphics[width=0.116\textwidth]{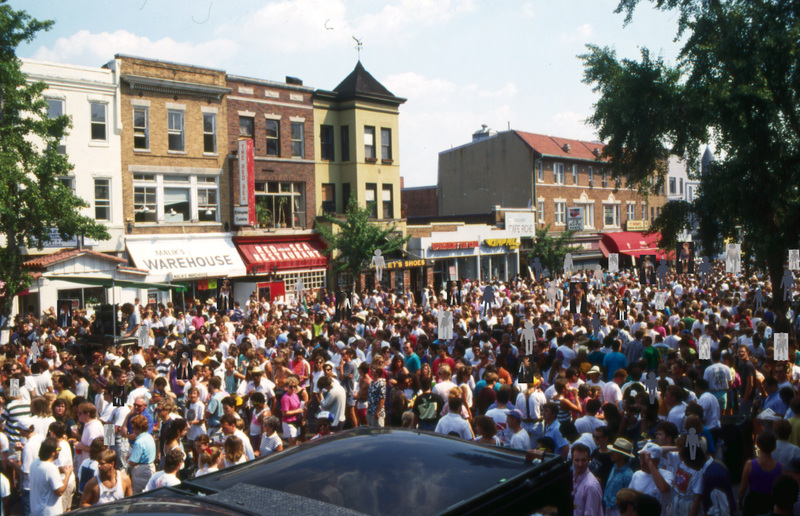}} & \raisebox{-0.5\height}{\includegraphics[width=0.116\textwidth]{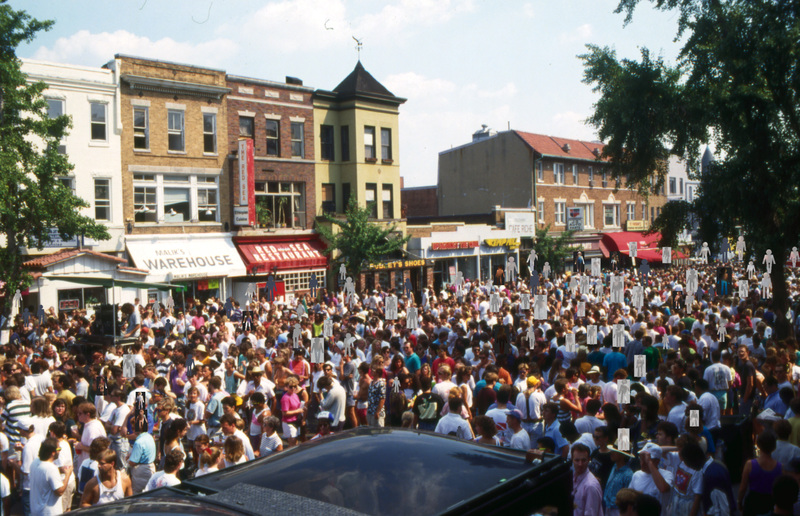}} \\
\scriptsize lens flare\hspace{2pt} & \raisebox{-0.5\height}{\includegraphics[width=0.116\textwidth]{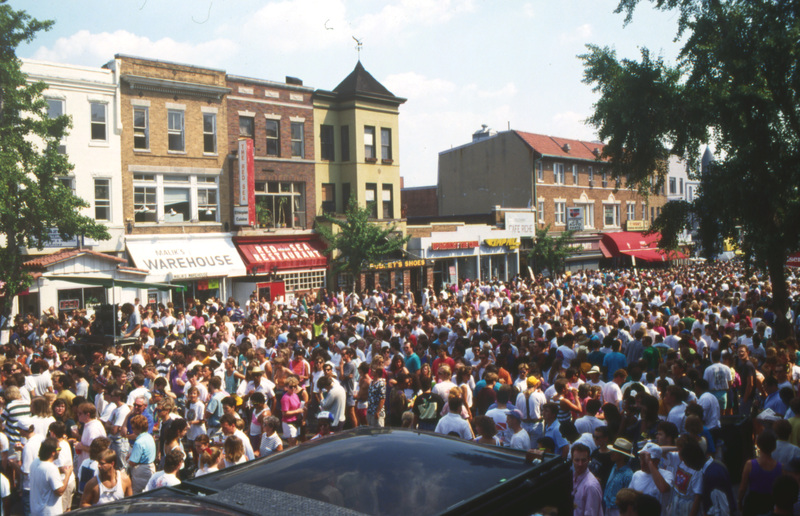}} & \raisebox{-0.5\height}{\includegraphics[width=0.116\textwidth]{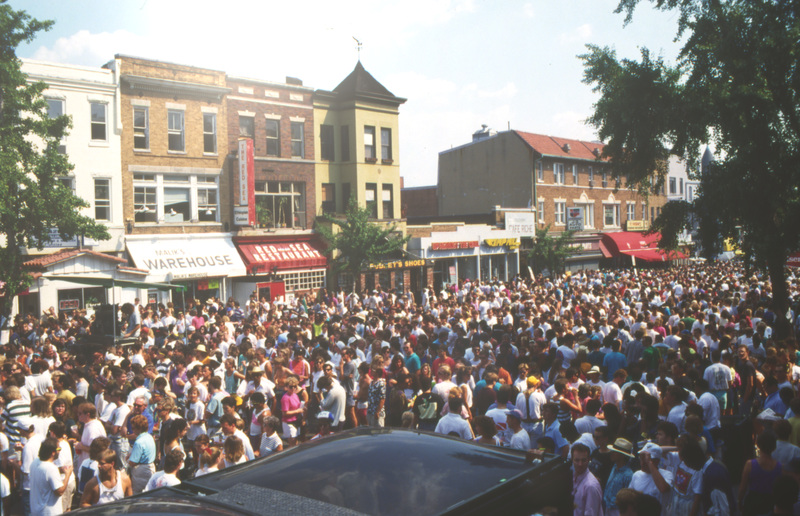}} & \raisebox{-0.5\height}{\includegraphics[width=0.116\textwidth]{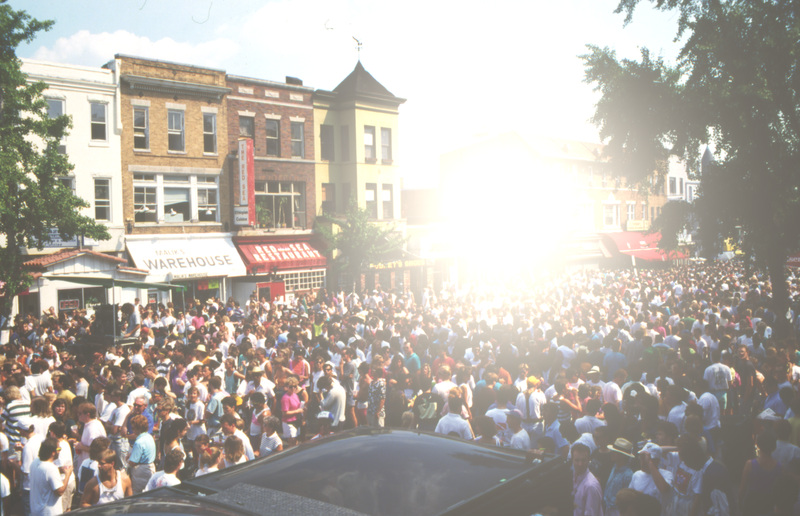}} & \scriptsize object distractors\hspace{2pt} & \raisebox{-0.5\height}{\includegraphics[width=0.116\textwidth]{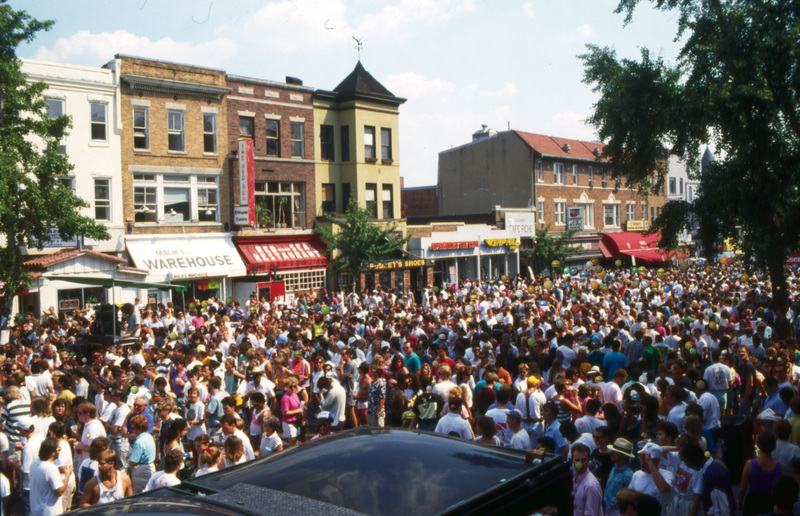}} & \raisebox{-0.5\height}{\includegraphics[width=0.116\textwidth]{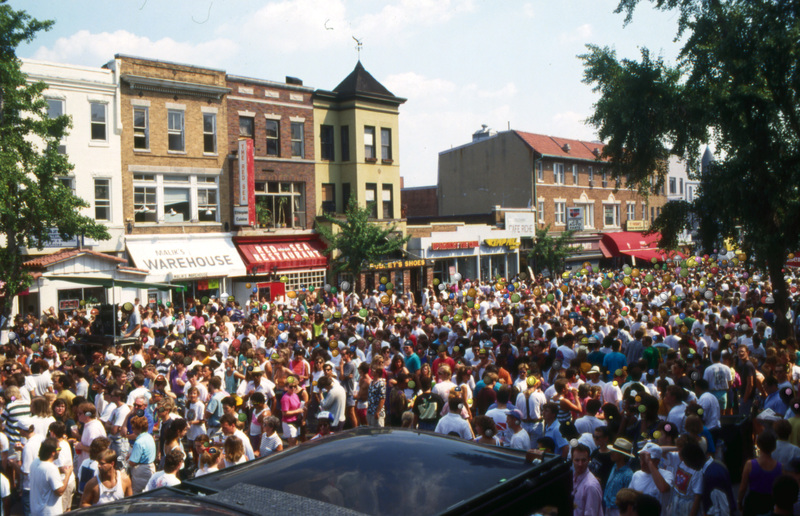}} & \raisebox{-0.5\height}{\includegraphics[width=0.116\textwidth]{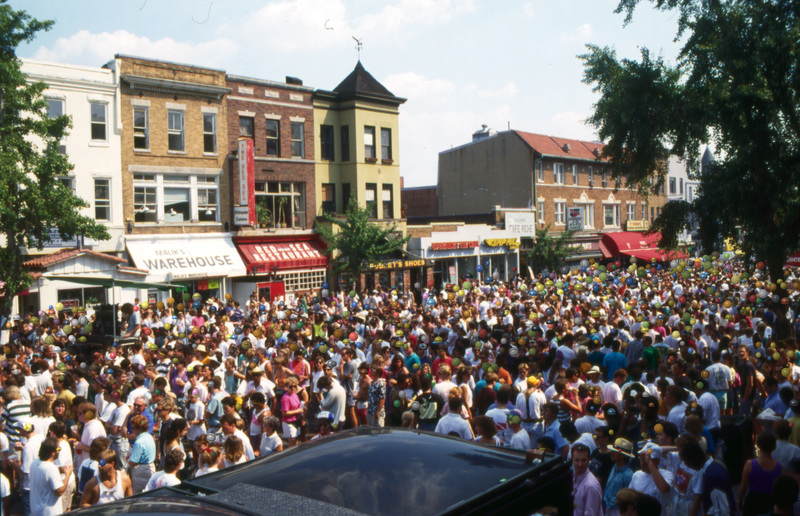}} \\
\scriptsize lens contamination\hspace{2pt} & \raisebox{-0.5\height}{\includegraphics[width=0.116\textwidth]{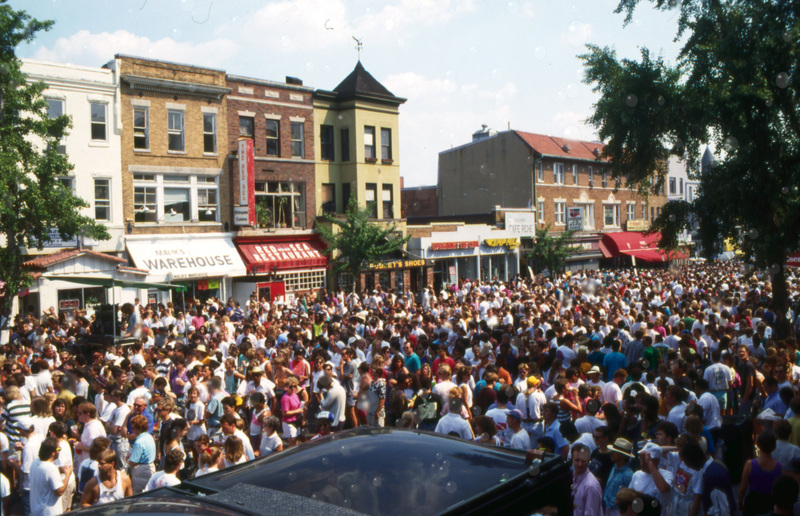}} & \raisebox{-0.5\height}{\includegraphics[width=0.116\textwidth]{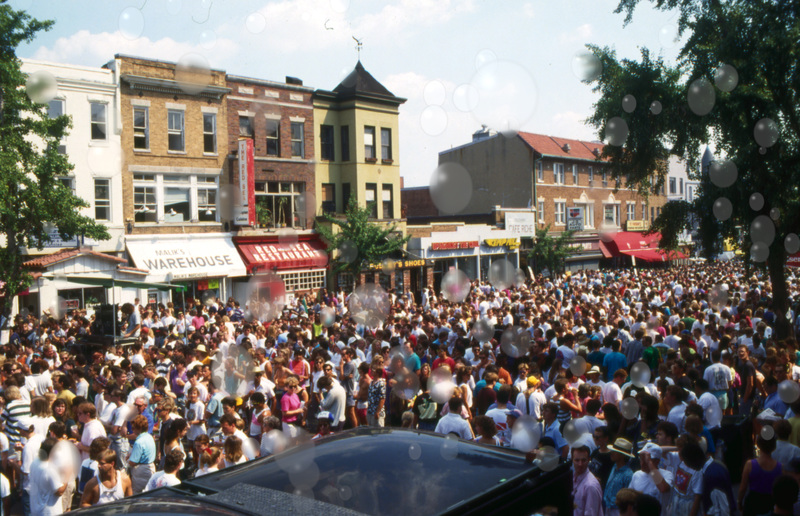}} & \raisebox{-0.5\height}{\includegraphics[width=0.116\textwidth]{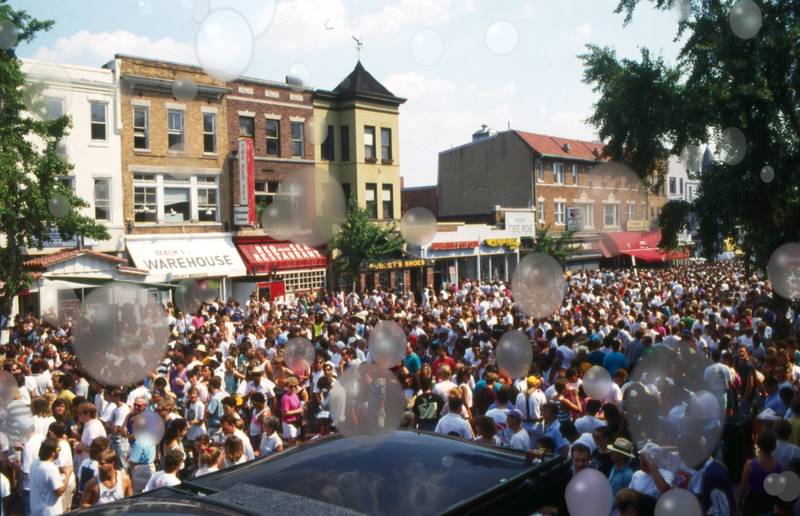}} & \scriptsize viewpoint shift\hspace{2pt} & \raisebox{-0.5\height}{\includegraphics[width=0.116\textwidth]{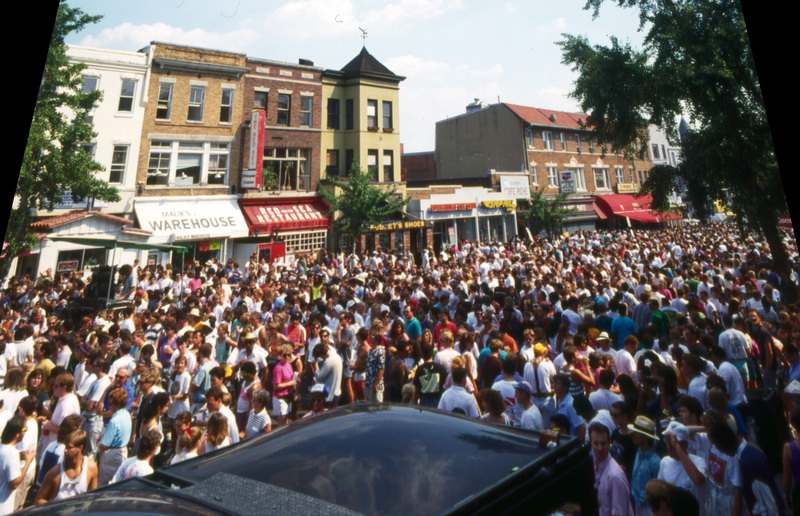}} & \raisebox{-0.5\height}{\includegraphics[width=0.116\textwidth]{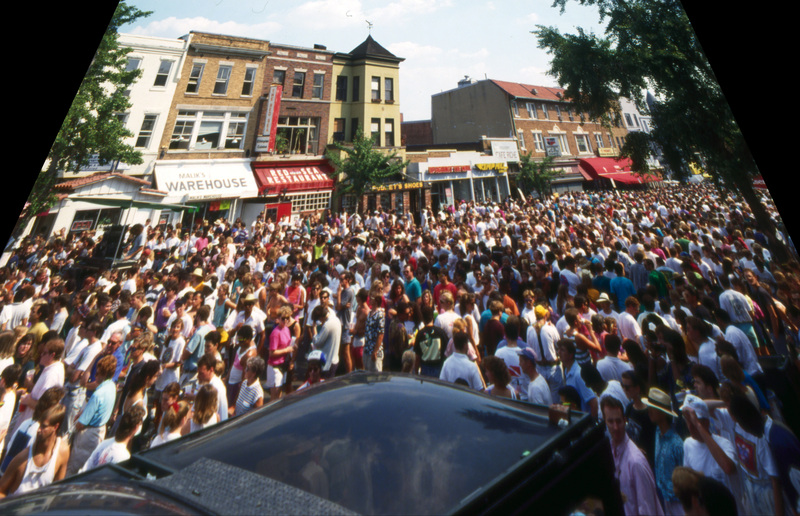}} & \raisebox{-0.5\height}{\includegraphics[width=0.116\textwidth]{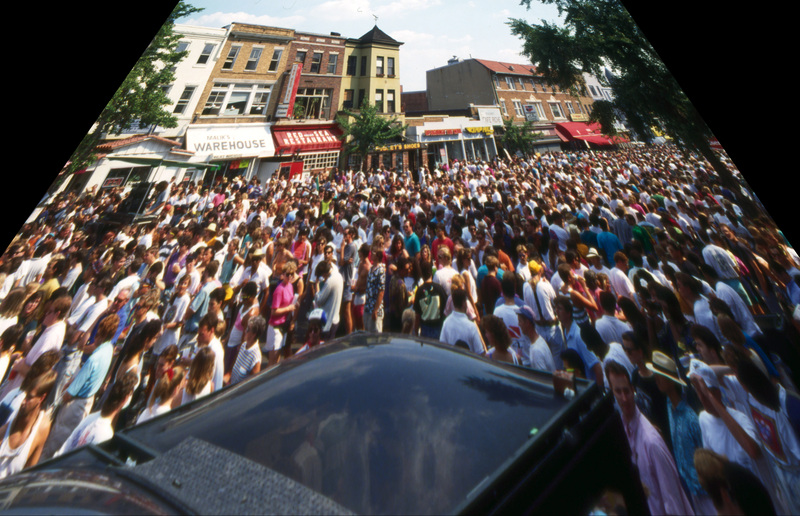}} \\
\scriptsize rain streaks\hspace{2pt} & \raisebox{-0.5\height}{\includegraphics[width=0.116\textwidth]{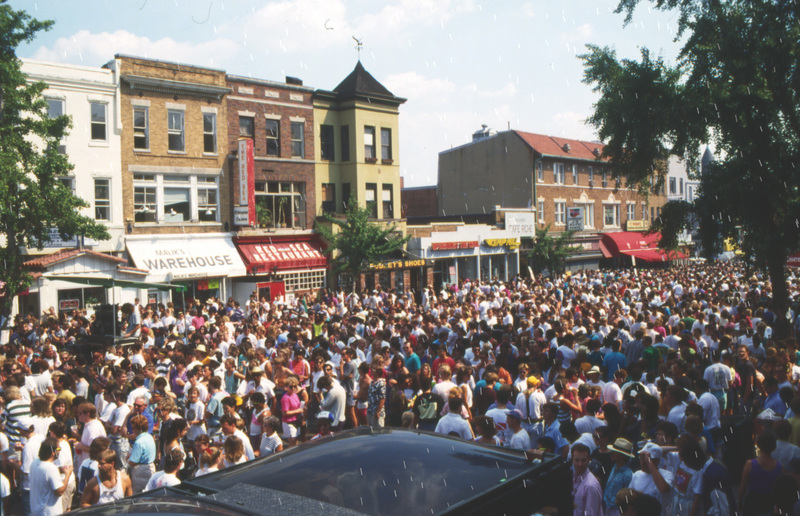}} & \raisebox{-0.5\height}{\includegraphics[width=0.116\textwidth]{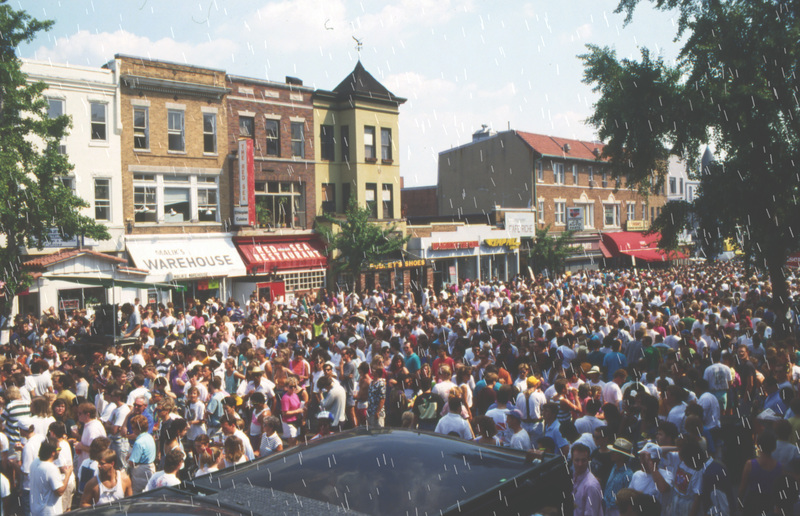}} & \raisebox{-0.5\height}{\includegraphics[width=0.116\textwidth]{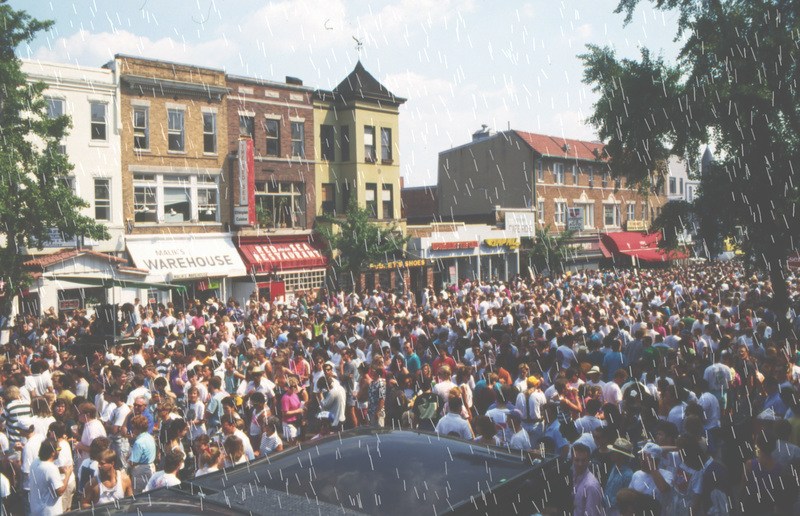}} & \scriptsize foreground occlusion\hspace{2pt} & \raisebox{-0.5\height}{\includegraphics[width=0.116\textwidth]{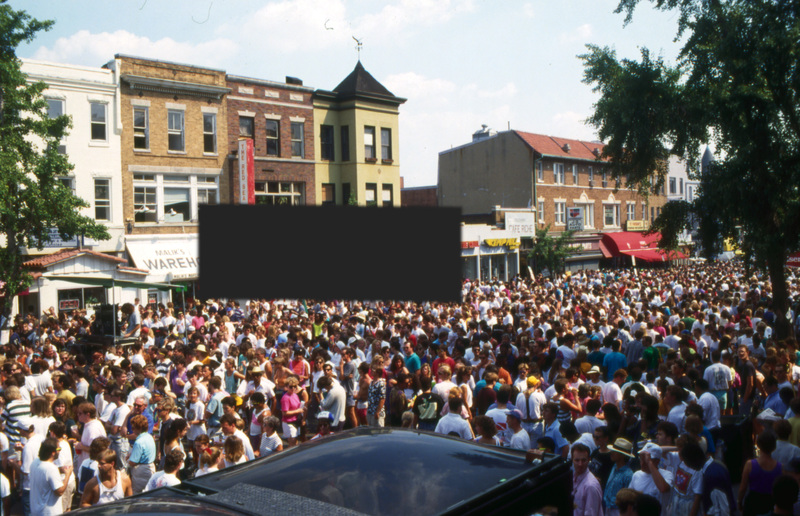}} & \raisebox{-0.5\height}{\includegraphics[width=0.116\textwidth]{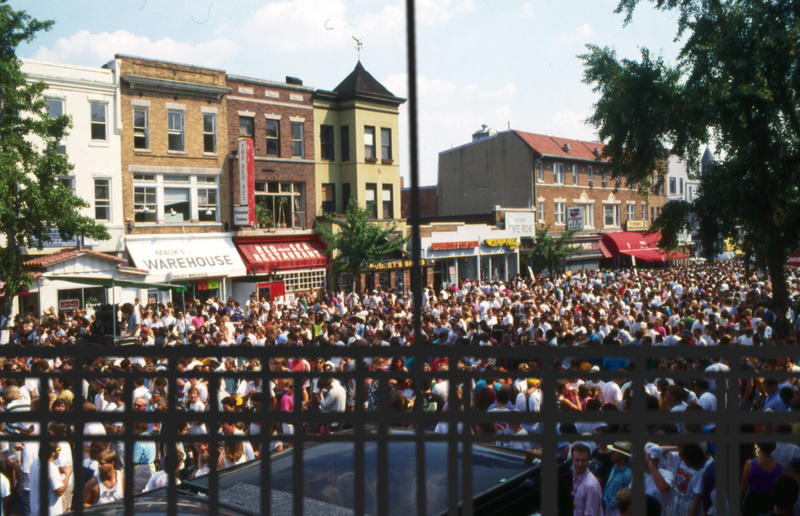}} & \raisebox{-0.5\height}{\includegraphics[width=0.116\textwidth]{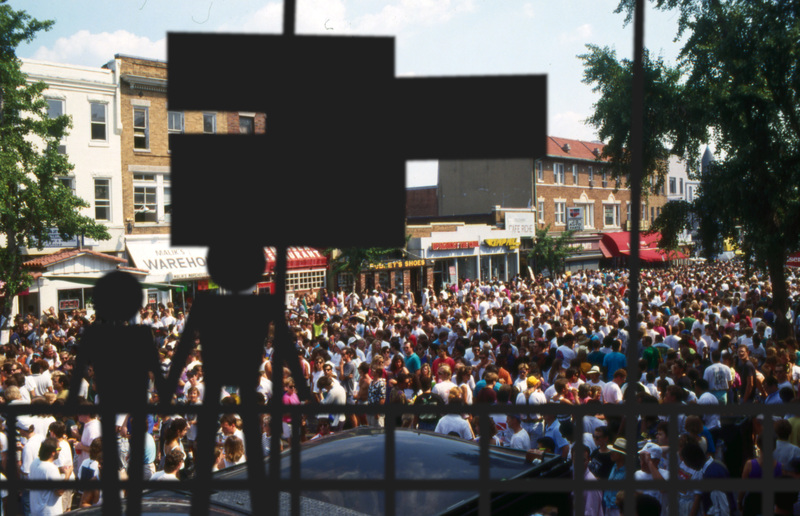}} \\
\scriptsize snow particles\hspace{2pt} & \raisebox{-0.5\height}{\includegraphics[width=0.116\textwidth]{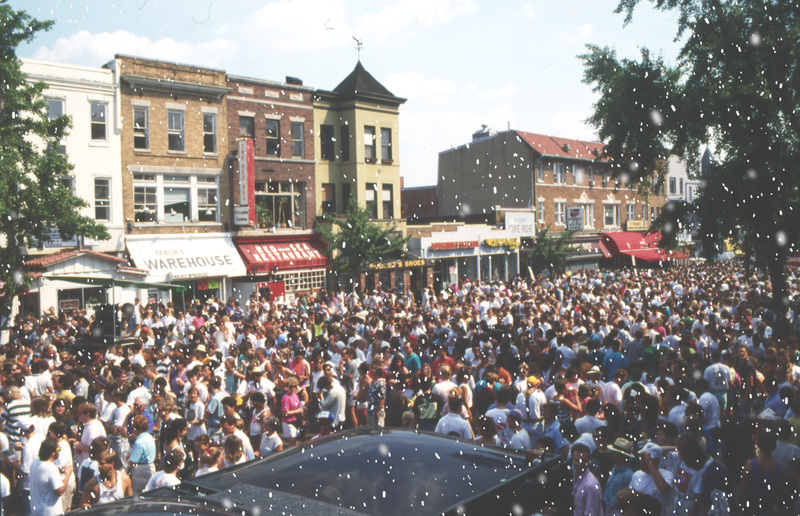}} & \raisebox{-0.5\height}{\includegraphics[width=0.116\textwidth]{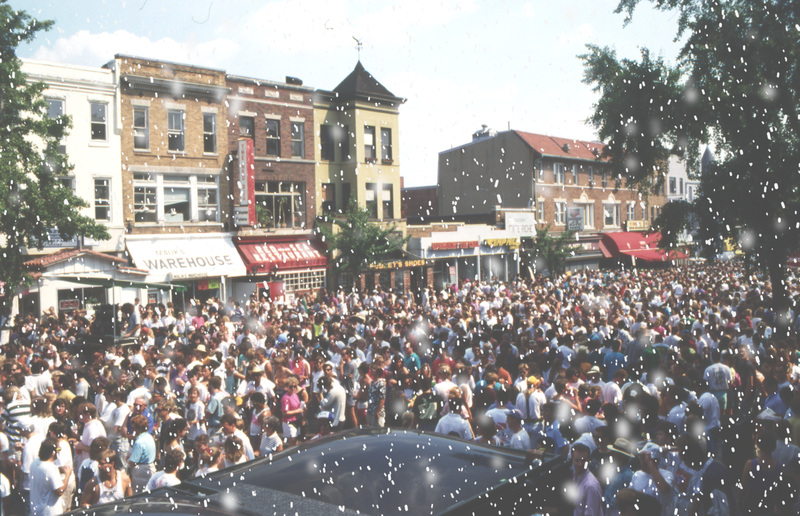}} & \raisebox{-0.5\height}{\includegraphics[width=0.116\textwidth]{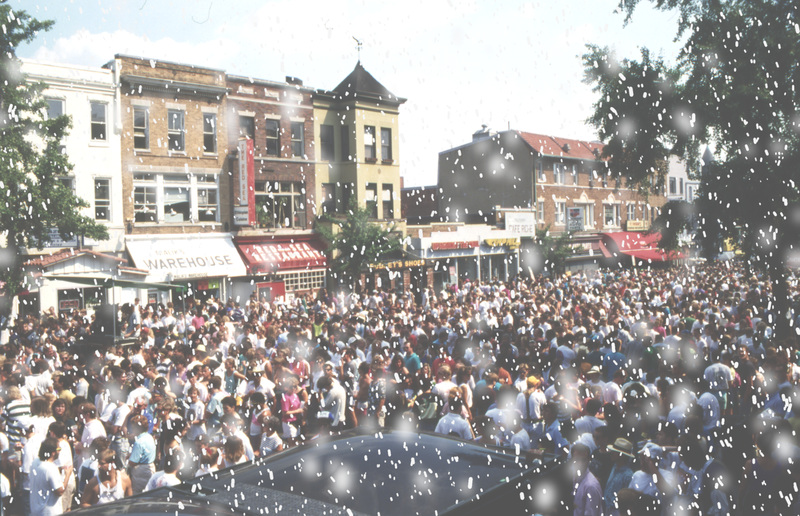}} & \scriptsize field-of-view loss\hspace{2pt} & \raisebox{-0.5\height}{\includegraphics[width=0.116\textwidth]{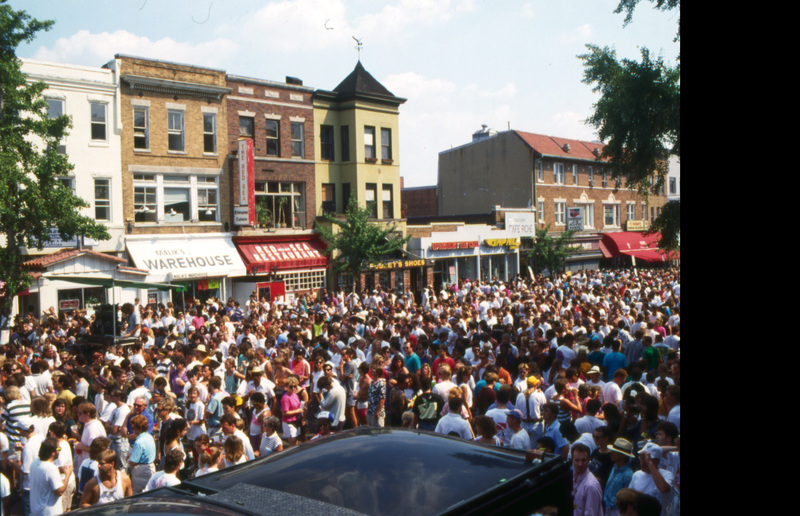}} & \raisebox{-0.5\height}{\includegraphics[width=0.116\textwidth]{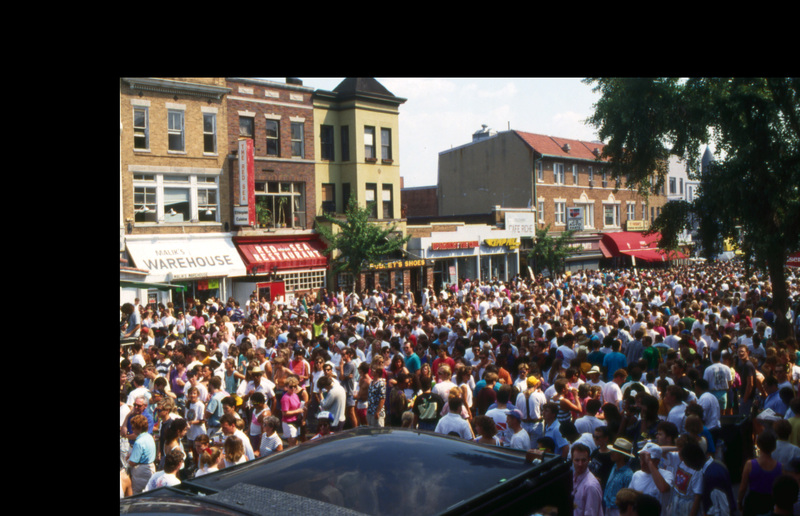}} & \raisebox{-0.5\height}{\includegraphics[width=0.116\textwidth]{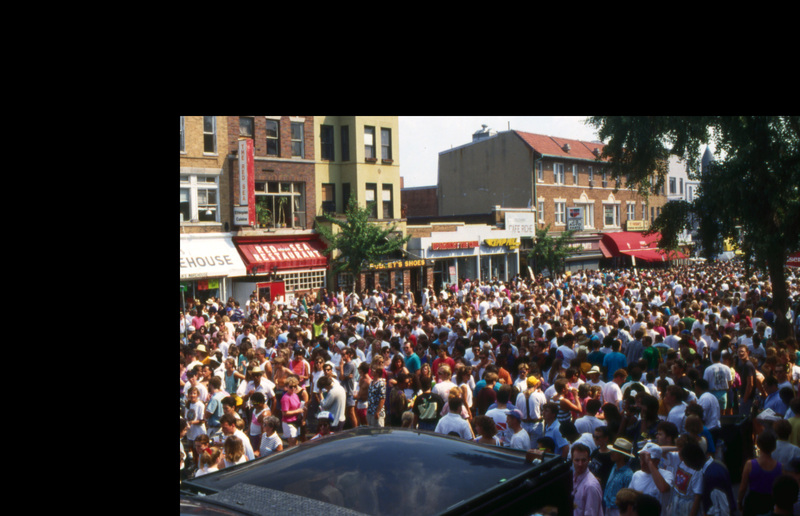}} \\
\scriptsize depth-aware haze\hspace{2pt} & \raisebox{-0.5\height}{\includegraphics[width=0.116\textwidth]{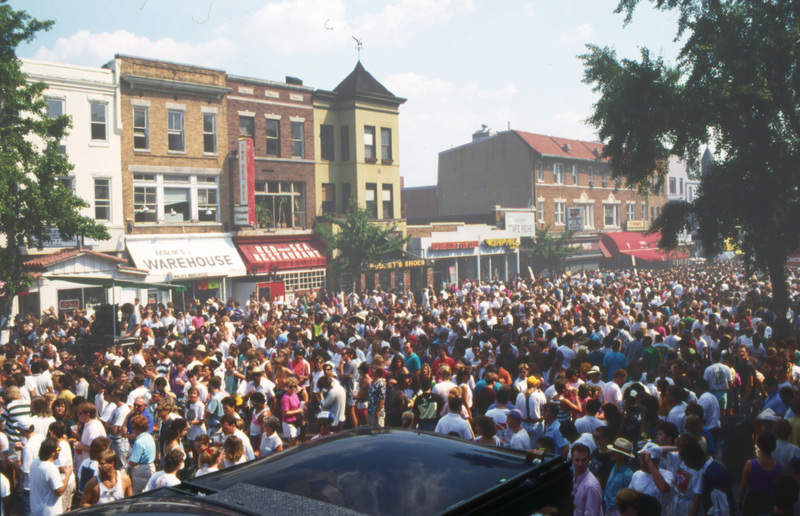}} & \raisebox{-0.5\height}{\includegraphics[width=0.116\textwidth]{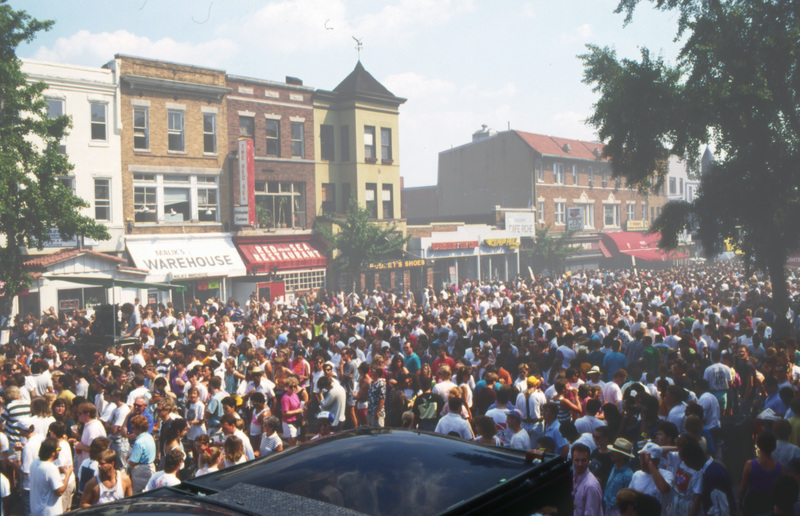}} & \raisebox{-0.5\height}{\includegraphics[width=0.116\textwidth]{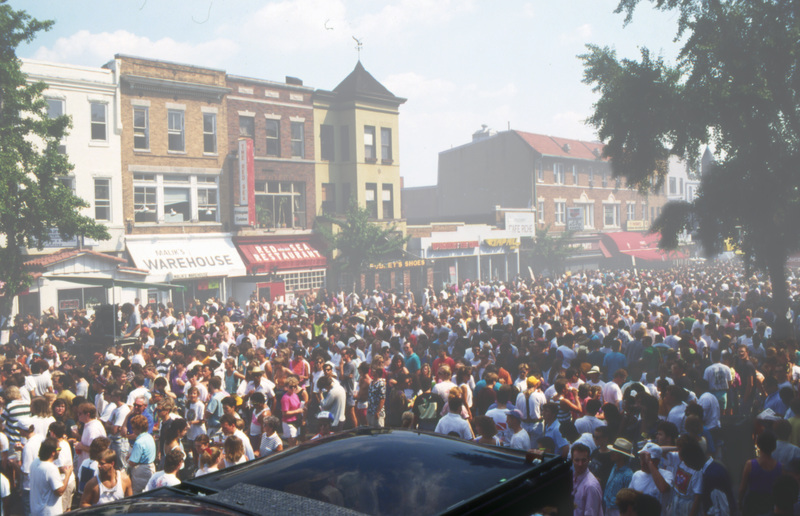}} &  &  &  &  \\
  \end{tabular}

  \caption{\textbf{Corruption visualizations across severity levels.} The same UCF-QNRF image is shown under all 23 corruptions at Light, Moderate, and High severity.}
  \label{fig:corruption-gallery-full}
\end{figure*}

\subsection{Severity response}

\Cref{fig:severity-response} reports mean FP32 relative degradation at each severity level, with one curve per corruption grouped by family. All 23 corruptions increase monotonically in the aggregate, and each family preserves this ordering across severity levels. This monotonicity applies to the aggregated response and does not imply monotonic behavior for every image or model. Corruptions differ primarily in the rate of degradation: far-field defocus increases most sharply, from $17.71\%$ to $181.89\%$, whereas several other corruptions show substantially smaller separation between severity levels.

\begin{figure*}[tp]
\centering
\includegraphics[width=\linewidth]{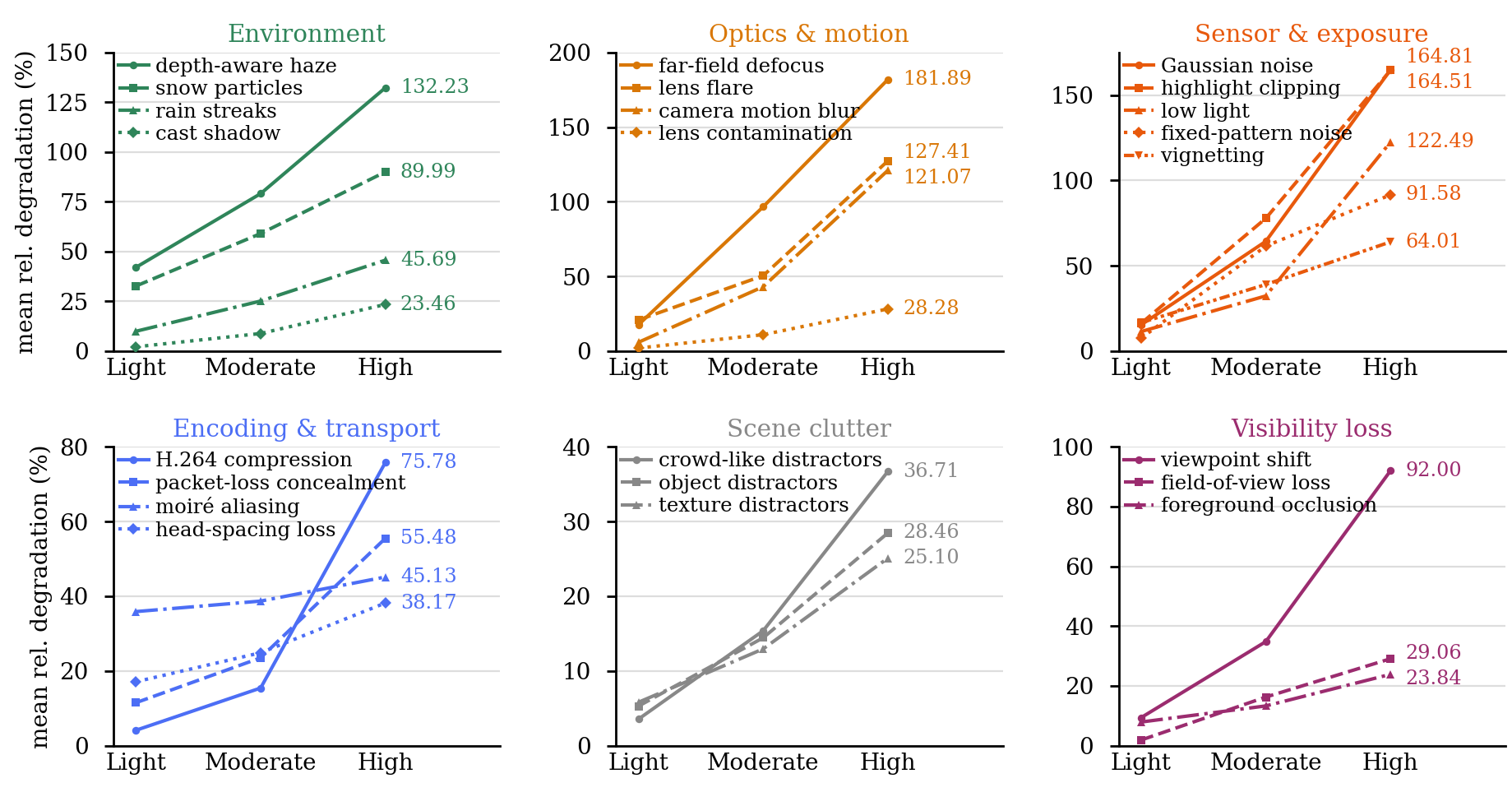}
\caption{\textbf{Severity response by corruption family.}
Mean FP32 relative degradation at Light, Moderate, and High severity, grouped by corruption family.
Aggregate degradation increases monotonically for all 23 corruptions, with far-field defocus exhibiting the steepest increase ($17.71\%$--$181.89\%$).}
\label{fig:severity-response}
\end{figure*}

\subsection{Synthetic and real weather}
\label{sec:supp-weather}

To assess weather realism, synthetic rain and snow are calibrated against labeled real-weather images from the JHU-Crowd++ training split (\cref{fig:weather-realism}). We compare real-weather images with seeded clean controls and retain two appearance statistics for rain and two for snow that distinguish weather from clean conditions. Each synthetic severity is then required to fall within the 10th--90th percentile range observed for real weather on the corresponding statistic. This checks selected appearance statistics, not whether robustness rankings or precision effects transfer to naturally captured weather.

\begin{figure}[htbp]
\centering
\begin{tikzpicture}
\begin{groupplot}[
  group style={group size=2 by 2, horizontal sep=1.15cm, vertical sep=0.95cm},
  width=0.55\linewidth, height=2.35cm,
  title style={font=\scriptsize\bfseries, yshift=-3pt},
  tick label style={font=\tiny}, grid=major, grid style={rulecol!18},
  axis on top,
]
\nextgroupplot[title={rain: rms contrast},
  xmin=0.1437, xmax=0.2743, ymin=0.4, ymax=3.6,
  ytick={1,2,3}, yticklabels={Light,Moderate,High}]
\addplot[draw=none, fill=dsC!22, forget plot] coordinates {(0.1580,0.4) (0.2600,0.4) (0.2600,3.6) (0.1580,3.6)} \closedcycle;
\draw[dsC, line width=0.7pt] (axis cs:0.2080,0.4) -- (axis cs:0.2080,3.6);
\draw[rulecol, dashed, line width=0.7pt] (axis cs:0.2490,0.4) -- (axis cs:0.2490,3.6);
\addplot[only marks, mark=*, mark size=1.7pt, draw=int8col, fill=int8col, forget plot] coordinates {(0.2150,1) (0.1900,2) (0.1700,3)};
\nextgroupplot[title={rain: saturation},
  xmin=0.0269, xmax=0.1741, ymin=0.4, ymax=3.6,
  ytick={1,2,3}, yticklabels={Light,Moderate,High}]
\addplot[draw=none, fill=dsC!22, forget plot] coordinates {(0.0430,0.4) (0.1580,0.4) (0.1580,3.6) (0.0430,3.6)} \closedcycle;
\draw[dsC, line width=0.7pt] (axis cs:0.0710,0.4) -- (axis cs:0.0710,3.6);
\draw[rulecol, dashed, line width=0.7pt] (axis cs:0.1240,0.4) -- (axis cs:0.1240,3.6);
\addplot[only marks, mark=*, mark size=1.7pt, draw=int8col, fill=int8col, forget plot] coordinates {(0.1060,1) (0.0930,2) (0.0800,3)};
\nextgroupplot[title={snow: mean luminance},
  xmin=0.1217, xmax=0.5813, ymin=0.4, ymax=3.6,
  ytick={1,2,3}, yticklabels={Light,Moderate,High}]
\addplot[draw=none, fill=dsC!22, forget plot] coordinates {(0.1900,0.4) (0.5310,0.4) (0.5310,3.6) (0.1900,3.6)} \closedcycle;
\draw[dsC, line width=0.7pt] (axis cs:0.3260,0.4) -- (axis cs:0.3260,3.6);
\draw[rulecol, dashed, line width=0.7pt] (axis cs:0.1720,0.4) -- (axis cs:0.1720,3.6);
\addplot[only marks, mark=*, mark size=1.7pt, draw=int8col, fill=int8col, forget plot] coordinates {(0.1960,1) (0.2450,2) (0.3280,3)};
\nextgroupplot[title={snow: saturation},
  xmin=0.0099, xmax=0.1481, ymin=0.4, ymax=3.6,
  ytick={1,2,3}, yticklabels={Light,Moderate,High}]
\addplot[draw=none, fill=dsC!22, forget plot] coordinates {(0.0250,0.4) (0.1330,0.4) (0.1330,3.6) (0.0250,3.6)} \closedcycle;
\draw[dsC, line width=0.7pt] (axis cs:0.0700,0.4) -- (axis cs:0.0700,3.6);
\draw[rulecol, dashed, line width=0.7pt] (axis cs:0.1270,0.4) -- (axis cs:0.1270,3.6);
\addplot[only marks, mark=*, mark size=1.7pt, draw=int8col, fill=int8col, forget plot] coordinates {(0.1060,1) (0.0880,2) (0.0720,3)};
\end{groupplot}
\end{tikzpicture}

\vspace{1pt}
{\scriptsize \textcolor{dsC}{$\blacksquare$}\,real weather [p10,\,p90] \quad \textcolor{dsC}{\textbf{|}}\,real weather median \quad \textcolor{rulecol}{\textbf{--\,--}}\,real \emph{clean} median \quad \textcolor{int8col}{$\bullet$}\,synthetic}
\caption{\textbf{Synthetic weather calibration.}
All rain and snow severity levels fall within the interdecile ranges of selected appearance statistics from real-weather images in JHU-Crowd++.}
\label{fig:weather-realism}
\end{figure}
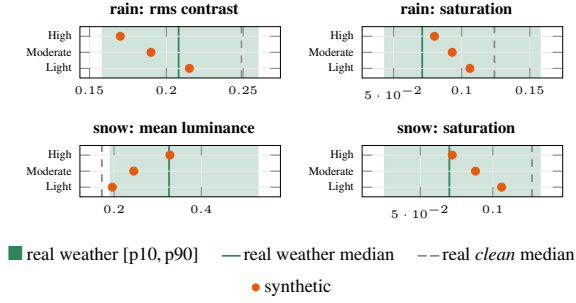

\subsection{Corruption redundancy}

Corruptions are potentially redundant when they induce similar relative rankings across the evaluated models. We therefore represent each corruption by its FP32 response profile, obtained by averaging relative degradation over the three severity levels for every model--dataset pair, and quantify pairwise similarity using Spearman correlation $\rho$.

\Cref{fig:redundancy} shows the resulting correlation matrix, ordered by hierarchical clustering using distance $1-\rho$. Among the 253 corruption pairs, only one exceeds $\rho=0.90$ and seven more fall in $[0.80,0.90)$. Correlations are higher within families than across families (median $0.659$ versus $0.506$), although the strongest associations cross family boundaries: lens flare and depth-aware haze reach $\rho=0.944$, while vignetting and cast shadow reach $\rho=0.890$.

We retain these correlated corruptions because similar response profiles do not imply equivalent failure mechanisms. The correlations describe model-level sensitivity rather than visual similarity or co-occurrence. Moreover, highly correlated corruptions can arise at different stages of the capture pipeline and require different interventions. For example, haze originates in the scene path, whereas lens flare arises within the optical system. Combining such corruptions would obscure mechanistic distinctions that are important for diagnosis and mitigation.

\input{components/supplement/figures/redundancy-matrix}

\section{Model Reproduction and Quantization}

\subsection{Reproduction of Published Clean Accuracy}
\label{sec:supp-roster}

\Cref{tab:repro-audit} compares our clean FP32 performance on each full test split with the corresponding published results. Each model was trained to reproduce its reported accuracy as closely as possible. Published references are available for 84 of the 104 model--dataset combinations; the remaining cases largely involve UCF-QNRF or JHU-Crowd++, which post-date several methods.

\begin{table*}[t]
\centering
\caption{\textbf{Comparison with published results.}
Clean FP32 MAE and RMSE for 26 models on the full test splits, compared with published values.
\better{Green} indicates lower error than reported, while \worse{red} marks relative deviations exceeding the 30\% reproduction threshold.
Dashes indicate unavailable published results.}
\label{tab:repro-audit}
\scriptsize
\setlength{\tabcolsep}{4.5pt}
\renewcommand{\arraystretch}{0.92}
\begin{tabular}{@{}l rr@{\hspace{12pt}}rr@{\hspace{12pt}}rr@{\hspace{12pt}}rr@{}}
\toprule
 & \multicolumn{2}{c}{ShanghaiTech~A} & \multicolumn{2}{c}{ShanghaiTech~B} & \multicolumn{2}{c}{UCF-QNRF} & \multicolumn{2}{c}{JHU-Crowd++} \\
\cmidrule(l{3pt}r{3pt}){2-3}\cmidrule(l{3pt}r{3pt}){4-5}\cmidrule(l{3pt}r{3pt}){6-7}\cmidrule(l{3pt}r{3pt}){8-9}
Model & Ours & Paper & Ours & Paper & Ours & Paper & Ours & Paper \\
\midrule
\multicolumn{9}{@{}l}{\textbf{CNN-based}} \\
\quad FFNet-MobileNetV3 & 70.68/118.90 & 64.30/109.30 & 12.64/23.91 & -- & 107.94/198.67 & -- & 65.14/233.29 & -- \\
\quad ZIP-Pico & 79.88/125.90 & 71.18/109.60 & 9.73/15.35 & 8.23/12.62 & 111.62/173.88 & 96.29/161.82 & 75.96/285.50 & -- \\
\quad ZIP-Nano & 70.49/109.17 & 58.86/94.63 & 8.50/13.69 & 7.74/12.14 & \worse{114.44}/\worse{196.29} & 86.46/147.64 & 73.53/275.65 & -- \\
\quad MobileCount & \better{86.66}/\better{139.69} & 89.40/146.00 & 9.24/\better{14.94} & 9.00/15.40 & \better{114.69}/\better{189.84} & 131.10/222.60 & 137.04/397.79 & -- \\
\quad ZIP-Tiny & 62.90/102.19 & 56.36/86.09 & 7.52/11.64 & 6.67/9.90 & 91.66/149.67 & 76.02/129.40 & 62.38/257.62 & -- \\
\quad ZIP-Small & 71.50/114.00 & 55.17/88.99 & 7.26/\worse{12.14} & 5.83/9.21 & 76.89/131.85 & 73.32/125.09 & 56.80/251.92 & -- \\
\quad FFNet-ConvNeXt-T & 54.92/88.57 & 48.30/80.50 & 6.78/11.51 & 6.10/9.00 & 84.13/144.82 & -- & 66.09/289.80 & -- \\
\quad MPCount & 61.19/104.02 & -- & 6.52/10.90 & -- & 110.08/203.75 & -- & 59.21/246.71 & -- \\
\quad STEERER-VGG19 & 62.29/103.30 & 55.60/87.30 & 7.90/11.60 & 6.80/10.70 & 86.44/145.71 & 76.70/135.10 & 58.79/247.29 & 55.40/221.40 \\
\quad APGCC & 57.85/96.33 & 48.80/76.70 & \worse{8.33}/\worse{13.12} & 5.60/8.70 & \worse{110.39}/175.93 & 80.10/136.60 & \worse{72.74}/289.72 & 54.30/225.90 \\
\quad P2PNet & 57.76/89.61 & 52.74/85.06 & 7.98/12.48 & 6.25/9.90 & 100.97/186.98 & 85.32/154.50 & 66.88/253.31 & -- \\
\quad CSRNet & \better{67.41}/\better{110.67} & 68.20/115.00 & 11.38/16.17 & 10.60/16.00 & \better{95.79}/168.94 & 110.00/-- & \better{68.86}/284.64 & 85.00/-- \\
\quad DM-Count & 65.72/111.63 & 59.70/95.70 & 7.83/12.78 & 7.40/11.80 & 91.53/161.48 & 85.60/148.30 & 62.38/261.20 & -- \\
\quad GCHF & 59.19/\better{94.26} & 57.50/94.30 & 7.10/11.66 & 6.90/11.00 & 81.82/143.67 & 80.30/137.60 & 60.42/254.21 & 57.00/235.70 \\
\quad CHF & \better{57.22}/96.23 & 57.50/94.30 & 7.27/12.33 & 6.90/11.00 & 84.17/142.46 & 80.30/137.60 & 60.36/251.77 & 57.00/235.70 \\
\quad MCNN & \better{99.36}/\better{154.14} & 110.20/173.20 & \better{18.29}/\better{32.76} & 26.40/41.30 & \better{177.76}/\better{274.84} & 277.00/426.00 & \better{118.12}/\better{363.14} & 188.90/483.40 \\
\quad STEERER-HRNet & \better{54.41}/\better{85.22} & 54.50/86.90 & 6.30/10.66 & 5.80/8.50 & 79.50/135.82 & 74.30/128.30 & 54.45/\better{231.87} & 54.30/238.30 \\
\quad FIDTM & 66.87/125.98 & 57.00/-- & 8.48/14.31 & 6.90/-- & 107.35/182.24 & 89.00/-- & 69.21/296.55 & 66.60/-- \\
\multicolumn{9}{@{}l}{\textbf{Transformer}} \\
\quad CCTrans-Weak & 75.60/119.14 & 64.40/95.40 & 8.26/13.58 & 7.00/11.50 & 109.74/202.17 & 92.10/158.90 & 61.76/247.98 & -- \\
\quad CCTrans-Full & 57.32/94.76 & 52.30/84.90 & 7.51/12.17 & 6.20/9.90 & 92.29/167.50 & 82.80/142.30 & 56.82/246.40 & -- \\
\quad Transcrowd-Token & 70.77/\better{113.37} & 69.00/116.50 & \better{9.05}/\better{17.13} & 10.60/19.70 & \better{91.41}/\better{166.97} & 98.90/176.10 & \better{62.94}/\better{259.82} & 76.40/319.80 \\
\quad Transcrowd-GAP & 67.04/106.54 & 66.10/105.10 & 10.21/\worse{21.44} & 9.30/16.10 & \better{91.36}/\better{162.29} & 97.20/168.50 & \better{68.00}/\better{284.99} & 74.90/295.60 \\
\quad CCST & 63.56/100.36 & 62.80/94.10 & \better{8.14}/\better{12.55} & 8.30/13.40 & 96.27/177.25 & 93.70/166.90 & 61.42/\better{234.39} & 61.40/239.30 \\
\multicolumn{9}{@{}l}{\textbf{Hybrid}} \\
\quad PET & 52.35/82.45 & 49.34/78.77 & 6.78/10.01 & 6.19/9.69 & 100.53/\worse{191.21} & 79.53/144.32 & 61.89/252.49 & 58.50/238.00 \\
\quad Gramformer & 60.25/94.70 & 54.70/87.10 & 7.19/11.75 & -- & 80.63/134.75 & 76.70/129.50 & 55.91/235.87 & 53.10/228.10 \\
\multicolumn{9}{@{}l}{\textbf{Vision-language}} \\
\quad CLIP-EBC-RN50 & 56.36/89.81 & 54.00/-- & 6.71/10.68 & 6.00/-- & 91.15/163.74 & 80.50/-- & 60.08/261.57 & -- \\
\bottomrule
\end{tabular}

\end{table*}

\subsection{INT8 calibration}
\label{sec:supp-int8-calibration}

We apply post-training INT8 quantization to the exported ONNX graph without retraining or fine-tuning. Calibration images are drawn from the training split by count-stratified sampling: images whose ground-truth counts lie within the training split's interquartile range are sorted by count, divided into ten equally populated strata, and sampled equally from each. The largest images and those with the most extreme aspect ratios are added to cover the input shapes encountered at inference.

A calibration pass over this fixed subset collects activation histograms, from which symmetric quantization scales are determined after selecting clipping thresholds. Thresholds are optimized independently for each model--dataset configuration by searching a logarithmic range of tail-mass values and refining the optimum by bisection until further changes fall below the measured build-to-build variation. Each threshold is selected to minimize the clean MAE of the INT8 engine on the corresponding calibration images. The resulting calibration parameters are then used to build the deployment engine without further calibration.

\section{Detailed Robustness Evaluation}
\label{sec:supp-robustness}

\Cref{tab:per-cell-SHA,tab:per-cell-SHB,tab:per-cell-QNRF,tab:per-cell-JHU} provide the complete robustness results for every model--dataset--precision configuration, including clean MAE, mean corrupted MAE, and relative degradation. RD is computed relative to the corresponding clean MAE at the same precision.

\begin{table}[htbp]
\centering
\caption{\textbf{Robustness on ShanghaiTech~A.}
Clean MAE, mcMAE, and RD across numerical precisions.
RD is relative to the corresponding clean baseline;
$\Delta_{\mathrm{RD}}=\mathrm{RD}_{\mathrm{INT8}}-\mathrm{RD}_{\mathrm{FP32}}$
is reported in percentage points.}
\label{tab:per-cell-SHA}
\footnotesize
\setlength{\tabcolsep}{3.2pt}
\renewcommand{\arraystretch}{0.9}
\resizebox{\columnwidth}{!}{%
\begin{tabular}{@{}l rrr rr rrr r@{}}
\toprule
& \multicolumn{3}{c}{clean MAE} & \multicolumn{2}{c}{mcMAE}
& \multicolumn{3}{c}{RD (\%)} & \\
\cmidrule(lr){2-4}\cmidrule(lr){5-6}\cmidrule(lr){7-9}
Model & FP32 & FP16 & INT8 & FP32 & INT8 & FP32 & FP16 & INT8 & $\Delta_{\rm RD}$ \\
\midrule
APGCC & 57.9 & 57.9 & 57.3 & 75.4 & 76.1 & +30.4 & +30.2 & +32.7 & +2.3 \\
CHF & 57.2 & 57.3 & 55.7 & 74.8 & 73.6 & +30.7 & +30.7 & +32.2 & +1.5 \\
GCHF & 59.2 & 59.2 & 58.1 & 73.4 & 72.5 & +24.0 & +24.0 & +24.8 & +0.7 \\
CCST & 63.6 & 63.6 & 63.2 & 72.3 & 72.0 & +13.8 & +13.7 & +13.9 & +0.2 \\
CCTrans-Full & 57.3 & 57.4 & 59.7 & 70.5 & 73.2 & +23.0 & +22.8 & +22.6 & -0.3 \\
CCTrans-Weak & 75.6 & 75.5 & 75.9 & 85.9 & 86.6 & +13.6 & +13.8 & +14.0 & +0.4 \\
CLIP-EBC-RN50 & 56.4 & 56.4 & 57.2 & 69.4 & 70.1 & +23.1 & +23.1 & +22.6 & -0.5 \\
CSRNet & 67.4 & 67.4 & 68.2 & 87.2 & 87.9 & +29.4 & +29.4 & +28.9 & -0.6 \\
DM-Count & 65.7 & 65.7 & 66.0 & 81.0 & 81.1 & +23.3 & +23.3 & +22.9 & -0.3 \\
FFNet-ConvNeXt-T & 54.9 & 54.9 & 56.5 & 70.2 & 71.6 & +27.8 & +27.8 & +26.8 & -1.1 \\
FFNet-MobileNetV3 & 70.7 & 70.8 & 68.8 & 96.0 & 94.7 & +35.7 & +35.7 & +37.6 & +1.9 \\
FIDTM & 66.9 & 66.5 & 67.3 & 80.4 & 80.6 & +20.2 & +20.3 & +19.7 & -0.4 \\
Gramformer & 60.3 & 60.3 & 60.4 & 74.7 & 74.8 & +24.0 & +24.0 & +23.8 & -0.2 \\
MCNN & 99.4 & 99.4 & 98.4 & 124.5 & 128.7 & +25.3 & +25.3 & +30.7 & +5.4 \\
MobileCount & 86.7 & 86.6 & 88.3 & 108.3 & 108.0 & +25.0 & +25.1 & +22.3 & -2.6 \\
MPCount & 61.2 & 61.2 & 61.7 & 70.9 & 71.4 & +15.9 & +15.8 & +15.7 & -0.1 \\
P2PNet & 57.8 & 57.8 & 57.5 & 80.7 & 82.1 & +39.7 & +39.7 & +42.9 & +3.2 \\
PET & 52.4 & 52.3 & 54.6 & 70.8 & 69.9 & +35.3 & +35.5 & +27.9 & -7.3 \\
STEERER-HRNet & 54.4 & 54.4 & 59.1 & 68.9 & 71.3 & +26.6 & +26.6 & +20.6 & -6.0 \\
STEERER-VGG19 & 62.3 & 62.4 & 59.6 & 79.6 & 75.9 & +27.9 & +27.9 & +27.2 & -0.6 \\
Transcrowd-GAP & 67.0 & 67.0 & 68.4 & 77.1 & 78.9 & +15.0 & +15.1 & +15.2 & +0.2 \\
Transcrowd-Token & 70.8 & 70.7 & 73.1 & 80.4 & 83.8 & +13.7 & +13.7 & +14.7 & +1.0 \\
ZIP-Nano & 70.5 & 70.6 & 62.3 & 86.1 & 82.9 & +22.1 & +22.0 & +33.0 & +11.0 \\
ZIP-Pico & 79.9 & 79.9 & 74.9 & 100.6 & 92.4 & +25.9 & +26.0 & +23.2 & -2.7 \\
ZIP-Small & 71.5 & 71.6 & 64.6 & 77.8 & 72.2 & +8.8 & +8.9 & +11.7 & +2.9 \\
ZIP-Tiny & 62.9 & 62.9 & 61.3 & 78.7 & 75.6 & +25.2 & +25.1 & +23.3 & -1.9 \\
\bottomrule
\end{tabular}}

\end{table}

\begin{table}[htbp]
\centering
\caption{\textbf{Robustness on ShanghaiTech~B.}
Clean MAE, mcMAE, and RD across numerical precisions.
RD is relative to the corresponding clean baseline;
$\Delta_{\mathrm{RD}}=\mathrm{RD}_{\mathrm{INT8}}-\mathrm{RD}_{\mathrm{FP32}}$
is reported in percentage points.}
\label{tab:per-cell-SHB}
\footnotesize
\setlength{\tabcolsep}{3.2pt}
\renewcommand{\arraystretch}{0.9}
\resizebox{\columnwidth}{!}{%
\begin{tabular}{@{}l rrr rr rrr r@{}}
\toprule
& \multicolumn{3}{c}{clean MAE} & \multicolumn{2}{c}{mcMAE}
& \multicolumn{3}{c}{RD (\%)} & \\
\cmidrule(lr){2-4}\cmidrule(lr){5-6}\cmidrule(lr){7-9}
Model & FP32 & FP16 & INT8 & FP32 & INT8 & FP32 & FP16 & INT8 & $\Delta_{\rm RD}$ \\
\midrule
APGCC & 8.3 & 8.3 & 8.6 & 18.5 & 18.8 & +122.5 & +122.9 & +118.9 & -3.6 \\
CHF & 7.3 & 7.3 & 7.9 & 17.4 & 16.8 & +139.4 & +139.4 & +112.8 & -26.6 \\
GCHF & 7.1 & 7.1 & 8.3 & 14.2 & 14.6 & +100.2 & +100.3 & +76.8 & -23.4 \\
CCST & 8.1 & 8.1 & 8.3 & 13.8 & 13.7 & +69.4 & +69.7 & +65.4 & -4.0 \\
CCTrans-Full & 7.5 & 7.5 & 8.1 & 12.5 & 13.7 & +66.0 & +66.2 & +68.8 & +2.7 \\
CCTrans-Weak & 8.3 & 8.3 & 8.7 & 14.5 & 14.9 & +75.0 & +75.1 & +71.8 & -3.2 \\
CLIP-EBC-RN50 & 6.7 & 6.7 & 6.9 & 17.2 & 17.9 & +156.1 & +156.4 & +159.7 & +3.5 \\
CSRNet & 11.4 & 11.4 & 11.1 & 26.5 & 26.1 & +133.1 & +133.1 & +134.5 & +1.4 \\
DM-Count & 7.8 & 7.8 & 7.9 & 16.3 & 15.9 & +108.1 & +108.1 & +99.7 & -8.4 \\
FFNet-ConvNeXt-T & 6.8 & 6.8 & 7.8 & 12.9 & 13.8 & +89.8 & +89.6 & +77.6 & -12.1 \\
FFNet-MobileNetV3 & 12.6 & 12.8 & 11.4 & 33.3 & 29.0 & +163.2 & +159.5 & +155.0 & -8.1 \\
FIDTM & 8.5 & 9.0 & 7.5 & 18.8 & 17.2 & +121.8 & +99.0 & +131.3 & +9.6 \\
Gramformer & 7.2 & 7.2 & 8.0 & 17.4 & 17.9 & +141.7 & +141.8 & +123.6 & -18.1 \\
MCNN & 18.3 & 18.3 & 18.2 & 41.9 & 52.4 & +129.3 & +129.3 & +187.4 & +58.1 \\
MobileCount & 9.2 & 9.2 & 9.2 & 26.7 & 25.9 & +188.9 & +188.9 & +180.4 & -8.5 \\
MPCount & 6.5 & 6.5 & 6.4 & 12.1 & 11.9 & +86.0 & +85.8 & +84.6 & -1.4 \\
P2PNet & 8.0 & 8.0 & 8.1 & 18.7 & 18.4 & +134.2 & +134.4 & +126.2 & -8.0 \\
PET & 6.8 & 6.7 & 6.9 & 17.9 & 18.4 & +163.8 & +165.3 & +166.4 & +2.6 \\
STEERER-HRNet & 6.3 & 6.3 & 6.2 & 14.0 & 13.4 & +122.6 & +122.3 & +114.9 & -7.7 \\
STEERER-VGG19 & 7.9 & 7.9 & 8.6 & 15.0 & 15.5 & +90.4 & +90.1 & +80.6 & -9.8 \\
Transcrowd-GAP & 10.2 & 10.2 & 10.4 & 15.3 & 15.3 & +50.3 & +50.3 & +46.6 & -3.7 \\
Transcrowd-Token & 9.1 & 9.1 & 9.2 & 15.0 & 14.9 & +65.8 & +65.8 & +61.0 & -4.8 \\
ZIP-Nano & 8.5 & 8.5 & 8.9 & 17.6 & 19.1 & +106.8 & +106.6 & +114.6 & +7.8 \\
ZIP-Pico & 9.7 & 9.7 & 9.8 & 19.7 & 23.4 & +102.7 & +104.6 & +139.9 & +37.2 \\
ZIP-Small & 7.3 & 7.3 & 8.8 & 12.3 & 14.2 & +69.4 & +68.9 & +60.6 & -8.8 \\
ZIP-Tiny & 7.5 & 7.5 & 8.2 & 14.4 & 15.8 & +91.2 & +91.2 & +91.3 & +0.0 \\
\bottomrule
\end{tabular}}

\end{table}

\begin{table}[htbp]
\centering
\caption{\textbf{Robustness on UCF-QNRF.}
Clean MAE, mcMAE, and RD across numerical precisions.
RD is relative to the corresponding clean baseline;
$\Delta_{\mathrm{RD}}=\mathrm{RD}_{\mathrm{INT8}}-\mathrm{RD}_{\mathrm{FP32}}$
is reported in percentage points.}
\label{tab:per-cell-QNRF}
\footnotesize
\setlength{\tabcolsep}{3.2pt}
\renewcommand{\arraystretch}{0.9}
\resizebox{\columnwidth}{!}{%
\begin{tabular}{@{}l rrr rr rrr r@{}}
\toprule
& \multicolumn{3}{c}{clean MAE} & \multicolumn{2}{c}{mcMAE}
& \multicolumn{3}{c}{RD (\%)} & \\
\cmidrule(lr){2-4}\cmidrule(lr){5-6}\cmidrule(lr){7-9}
Model & FP32 & FP16 & INT8 & FP32 & INT8 & FP32 & FP16 & INT8 & $\Delta_{\rm RD}$ \\
\midrule
APGCC & 110.4 & 110.4 & 105.5 & 136.1 & 131.0 & +23.3 & +23.2 & +24.1 & +0.9 \\
CHF & 84.2 & 84.1 & 85.7 & 107.0 & 107.0 & +27.1 & +27.1 & +24.8 & -2.3 \\
GCHF & 81.8 & 81.8 & 84.1 & 104.9 & 105.9 & +28.2 & +28.1 & +25.9 & -2.3 \\
CCST & 96.3 & 96.2 & 96.7 & 106.5 & 107.5 & +10.6 & +10.7 & +11.2 & +0.6 \\
CCTrans-Full & 92.3 & 92.3 & 91.2 & 99.5 & 101.2 & +7.8 & +7.8 & +10.9 & +3.1 \\
CCTrans-Weak & 109.7 & 109.8 & 105.3 & 119.1 & 115.9 & +8.6 & +8.5 & +10.0 & +1.5 \\
CLIP-EBC-RN50 & 91.2 & 91.2 & 91.7 & 116.5 & 117.0 & +27.8 & +27.6 & +27.6 & -0.2 \\
CSRNet & 95.8 & 95.9 & 96.7 & 120.8 & 120.1 & +26.1 & +26.1 & +24.2 & -1.9 \\
DM-Count & 91.5 & 91.5 & 91.8 & 110.0 & 109.5 & +20.1 & +20.2 & +19.3 & -0.8 \\
FFNet-ConvNeXt-T & 84.1 & 84.1 & 89.4 & 99.4 & 105.8 & +18.2 & +18.2 & +18.3 & +0.1 \\
FFNet-MobileNetV3 & 107.9 & 108.1 & 109.3 & 133.1 & 130.7 & +23.3 & +23.6 & +19.6 & -3.8 \\
FIDTM & 107.4 & 106.8 & 107.5 & 129.1 & 127.9 & +20.2 & +20.0 & +19.0 & -1.3 \\
Gramformer & 80.6 & 80.7 & 85.8 & 106.3 & 111.0 & +31.9 & +31.7 & +29.3 & -2.6 \\
MCNN & 177.8 & 177.8 & 174.7 & 210.6 & 204.9 & +18.5 & +18.5 & +17.3 & -1.2 \\
MobileCount & 114.7 & 114.7 & 116.3 & 134.8 & 135.8 & +17.6 & +17.5 & +16.8 & -0.7 \\
MPCount & 110.1 & 110.1 & 108.8 & 141.2 & 144.1 & +28.3 & +28.3 & +32.4 & +4.1 \\
P2PNet & 101.0 & 101.1 & 102.0 & 117.7 & 118.4 & +16.6 & +16.6 & +16.2 & -0.4 \\
PET & 100.5 & 100.6 & 98.7 & 114.1 & 110.9 & +13.5 & +13.5 & +12.4 & -1.0 \\
STEERER-HRNet & 79.5 & 79.6 & 76.7 & 93.9 & 91.7 & +18.1 & +18.0 & +19.6 & +1.5 \\
STEERER-VGG19 & 86.4 & 86.1 & 91.2 & 104.0 & 106.8 & +20.3 & +20.3 & +17.1 & -3.2 \\
Transcrowd-GAP & 91.4 & 91.3 & 96.1 & 102.0 & 106.3 & +11.6 & +11.7 & +10.6 & -1.0 \\
Transcrowd-Token & 91.4 & 91.4 & 95.4 & 102.3 & 106.1 & +12.0 & +12.0 & +11.3 & -0.7 \\
ZIP-Nano & 114.4 & 114.3 & 122.9 & 135.9 & 142.7 & +18.8 & +18.8 & +16.2 & -2.6 \\
ZIP-Pico & 111.6 & 111.5 & 115.4 & 133.9 & 138.5 & +20.0 & +20.0 & +20.1 & +0.1 \\
ZIP-Small & 76.9 & 76.8 & 77.8 & 88.4 & 91.4 & +15.0 & +15.2 & +17.6 & +2.6 \\
ZIP-Tiny & 91.7 & 91.8 & 88.2 & 109.2 & 106.1 & +19.1 & +19.2 & +20.4 & +1.3 \\
\bottomrule
\end{tabular}}

\end{table}

\begin{table}[htbp]
\centering
\caption{\textbf{Robustness on JHU-Crowd++.}
Clean MAE, mcMAE, and RD across numerical precisions.
RD is relative to the corresponding clean baseline;
$\Delta_{\mathrm{RD}}=\mathrm{RD}_{\mathrm{INT8}}-\mathrm{RD}_{\mathrm{FP32}}$
is reported in percentage points.}
\label{tab:per-cell-JHU}
\footnotesize
\setlength{\tabcolsep}{3.2pt}
\renewcommand{\arraystretch}{0.9}
\resizebox{\columnwidth}{!}{%
\begin{tabular}{@{}l rrr rr rrr r@{}}
\toprule
& \multicolumn{3}{c}{clean MAE} & \multicolumn{2}{c}{mcMAE}
& \multicolumn{3}{c}{RD (\%)} & \\
\cmidrule(lr){2-4}\cmidrule(lr){5-6}\cmidrule(lr){7-9}
Model & FP32 & FP16 & INT8 & FP32 & INT8 & FP32 & FP16 & INT8 & $\Delta_{\rm RD}$ \\
\midrule
APGCC & 72.7 & 72.8 & 73.4 & 80.9 & 81.2 & +11.3 & +11.2 & +10.7 & -0.5 \\
CHF & 60.4 & 60.4 & 60.3 & 72.2 & 72.8 & +19.6 & +19.7 & +20.6 & +1.0 \\
GCHF & 60.4 & 60.3 & 60.5 & 74.7 & 75.1 & +23.7 & +23.6 & +24.1 & +0.5 \\
CCST & 61.4 & 61.4 & 61.2 & 67.5 & 67.9 & +9.9 & +9.9 & +11.0 & +1.1 \\
CCTrans-Full & 56.8 & 56.8 & 59.1 & 63.8 & 67.8 & +12.3 & +12.3 & +14.7 & +2.4 \\
CCTrans-Weak & 61.8 & 61.8 & 62.4 & 68.8 & 72.3 & +11.4 & +11.3 & +15.9 & +4.5 \\
CLIP-EBC-RN50 & 60.1 & 60.1 & 60.3 & 69.6 & 68.9 & +15.9 & +15.7 & +14.3 & -1.6 \\
CSRNet & 68.9 & 68.9 & 68.8 & 86.2 & 85.8 & +25.2 & +25.2 & +24.8 & -0.4 \\
DM-Count & 62.4 & 62.4 & 62.4 & 73.5 & 73.0 & +17.9 & +17.9 & +17.1 & -0.8 \\
FFNet-ConvNeXt-T & 66.1 & 66.1 & 59.5 & 75.4 & 70.8 & +14.0 & +13.9 & +19.0 & +5.0 \\
FFNet-MobileNetV3 & 65.1 & 65.1 & 65.9 & 75.4 & 75.7 & +15.8 & +15.6 & +14.9 & -0.9 \\
FIDTM & 69.2 & 68.7 & 70.9 & 82.8 & 85.4 & +19.6 & +19.7 & +20.5 & +0.9 \\
Gramformer & 55.9 & 55.9 & 59.3 & 71.8 & 73.5 & +28.4 & +28.4 & +24.1 & -4.3 \\
MCNN & 118.1 & 118.1 & 117.9 & 144.0 & 142.8 & +21.9 & +21.9 & +21.1 & -0.8 \\
MobileCount & 137.0 & 137.0 & 140.0 & 186.2 & 192.3 & +35.9 & +35.9 & +37.4 & +1.5 \\
MPCount & 59.2 & 59.2 & 57.8 & 66.9 & 67.7 & +12.9 & +12.9 & +17.0 & +4.1 \\
P2PNet & 66.9 & 66.9 & 65.8 & 77.5 & 77.2 & +15.9 & +15.9 & +17.3 & +1.4 \\
PET & 61.9 & 61.9 & 61.1 & 70.9 & 68.4 & +14.6 & +14.7 & +11.9 & -2.7 \\
STEERER-HRNet & 54.4 & 54.4 & 54.3 & 64.0 & 63.9 & +17.6 & +17.5 & +17.7 & +0.1 \\
STEERER-VGG19 & 58.8 & 58.8 & 63.1 & 72.8 & 75.6 & +23.9 & +23.9 & +19.8 & -4.0 \\
Transcrowd-GAP & 68.0 & 68.0 & 67.3 & 78.4 & 77.4 & +15.3 & +15.4 & +15.0 & -0.3 \\
Transcrowd-Token & 62.9 & 62.9 & 64.1 & 69.2 & 70.9 & +9.9 & +9.9 & +10.5 & +0.6 \\
ZIP-Nano & 73.5 & 73.5 & 70.0 & 79.2 & 79.8 & +7.7 & +7.7 & +14.0 & +6.3 \\
ZIP-Pico & 76.0 & 75.9 & 91.4 & 85.0 & 95.8 & +11.9 & +11.9 & +4.8 & -7.0 \\
ZIP-Small & 56.8 & 56.8 & 55.9 & 67.2 & 65.9 & +18.4 & +18.3 & +17.7 & -0.6 \\
ZIP-Tiny & 62.4 & 62.4 & 61.5 & 79.8 & 73.1 & +27.9 & +27.9 & +18.7 & -9.1 \\
\bottomrule
\end{tabular}}

\end{table}

\Cref{tab:corruption-difficulty} reports corruption-level relative degradation across datasets and precisions, providing the full ranking of all 23 corruptions.

\begin{table}[htbp]
\centering
\caption{\textbf{Corruption difficulty.}
Mean relative degradation (RD) across 26 models, ranked by the four-dataset FP32 average.
F32/I8 denote FP32/INT8 precision; $^\dagger$ marks geometry-altering corruptions evaluated using kept-head ground truth.}
\label{tab:corruption-difficulty}
\scriptsize
\setlength{\tabcolsep}{3.5pt}
\renewcommand{\arraystretch}{0.92}
\begin{tabular}{@{}l rr@{\hspace{12pt}}rr@{\hspace{12pt}}rr@{\hspace{12pt}}rr@{}}
\toprule
& \multicolumn{2}{c}{SHA} & \multicolumn{2}{c}{SHB} & \multicolumn{2}{c}{QNRF} & \multicolumn{2}{c}{JHU} \\
\cmidrule(lr){2-3}\cmidrule(lr){4-5}\cmidrule(lr){6-7}\cmidrule(lr){8-9}
Corruption & F32 & I8 & F32 & I8 & F32 & I8 & F32 & I8 \\
\midrule
far-field defocus & 22 & 24 & 320 & 303 & 29 & 26 & 24 & 23 \\
highlight clipping & 40 & 42 & 227 & 223 & 49 & 48 & 30 & 30 \\
depth-aware haze & 39 & 37 & 252 & 244 & 28 & 27 & 18 & 16 \\
Gaussian noise & 49 & 51 & 183 & 182 & 35 & 38 & 58 & 59 \\
lens flare & 23 & 22 & 206 & 196 & 26 & 25 & 10 & 9 \\
snow particles & 54 & 49 & 115 & 109 & 37 & 36 & 36 & 37 \\
camera motion blur & 17 & 20 & 170 & 163 & 19 & 16 & 20 & 19 \\
low light & 36 & 36 & 138 & 131 & 20 & 20 & 28 & 30 \\
fixed-pattern noise & 32 & 32 & 111 & 107 & 42 & 41 & 29 & 31 \\
viewpoint shift$^\dagger$ & 24 & 25 & 110 & 104 & 22 & 21 & 26 & 26 \\
vignetting & 9 & 10 & 134 & 123 & 9 & 8 & 7 & 6 \\
moir\'e aliasing & 44 & 45 & 82 & 79 & 9 & 9 & 24 & 25 \\
H.264 compression & 22 & 24 & 70 & 67 & 13 & 13 & 22 & 22 \\
packet-loss concealment$^\dagger$ & 26 & 24 & 63 & 61 & 19 & 20 & 13 & 14 \\
rain streaks & 17 & 15 & 58 & 55 & 19 & 18 & 14 & 14 \\
head-spacing loss & 27 & 29 & 67 & 64 & 6 & 5 & 7 & 7 \\
crowd-like distractors$^\dagger$ & 3 & 4 & 60 & 61 & 7 & 7 & 5 & 5 \\
object distractors$^\dagger$ & 11 & 9 & 29 & 28 & 16 & 17 & 8 & 8 \\
field-of-view loss$^\dagger$ & 11 & 15 & 49 & 117 & 2 & 3 & 1 & 0 \\
foreground occlusion$^\dagger$ & 17 & 14 & 22 & 22 & 10 & 11 & 11 & 7 \\
texture distractors & 15 & 18 & 18 & 18 & 18 & 19 & 6 & 8 \\
lens contamination$^\dagger$ & 5 & 7 & 42 & 39 & 4 & 3 & 4 & 5 \\
cast shadow & 8 & 6 & 29 & 25 & 6 & 4 & 3 & 3 \\
\bottomrule
\end{tabular}

\end{table}

\Cref{tab:mce-reference} lists the CSRNet reference errors used to normalize mCE.
\begin{table}[t]
\centering
\caption{\textbf{mCE reference errors.}
Mean CSRNet MAE across three severity levels at FP32 for each corruption and dataset, using kept-annotation normalization where applicable.
Each model's mCE is normalized against the CSRNet reference at the same dataset and precision.
FP16 and INT8 references are computed analogously but omitted for brevity.
The clean row reports uncorrupted MAE.}
\label{tab:mce-reference}
\scriptsize
\renewcommand{\arraystretch}{0.9}
\setlength{\tabcolsep}{4pt}
\begin{tabular}{@{}l rrrr@{}}
\toprule
Corruption & SHA & SHB & QNRF & JHU \\
\midrule
\multicolumn{1}{@{}l}{\textit{clean (no corruption)}} & 67.4 & 11.4 & 95.8 & 68.9 \\
\midrule
\multicolumn{5}{@{}l}{\textbf{Sensor \& exposure}} \\
\quad fixed-pattern noise & 102.6 & 28.5 & 141.0 & 103.9 \\
\quad Gaussian noise & 99.8 & 33.4 & 155.6 & 117.5 \\
\quad highlight clipping & 121.1 & 38.3 & 179.5 & 109.4 \\
\quad low light & 93.0 & 38.0 & 141.5 & 101.0 \\
\quad vignetting & 81.5 & 33.8 & 114.3 & 79.6 \\
\multicolumn{5}{@{}l}{\textbf{Optics \& motion}} \\
\quad camera motion blur & 93.2 & 33.9 & 106.4 & 89.5 \\
\quad far-field defocus & 86.3 & 50.2 & 126.1 & 88.7 \\
\quad lens contamination & 73.9 & 17.2 & 103.3 & 74.9 \\
\quad lens flare & 88.7 & 42.2 & 131.5 & 83.1 \\
\multicolumn{5}{@{}l}{\textbf{Environment}} \\
\quad cast shadow & 74.6 & 17.9 & 103.9 & 73.7 \\
\quad depth-aware haze & 93.9 & 54.7 & 130.1 & 88.0 \\
\quad rain streaks & 79.1 & 27.3 & 119.6 & 81.2 \\
\quad snow particles & 80.2 & 35.5 & 128.2 & 93.9 \\
\multicolumn{5}{@{}l}{\textbf{Encoding \& transport}} \\
\quad H.264 compression & 92.0 & 20.6 & 113.4 & 92.4 \\
\quad moir\'e aliasing & 105.2 & 20.3 & 108.0 & 85.7 \\
\quad packet-loss concealment & 76.1 & 17.3 & 111.3 & 83.9 \\
\quad head-spacing loss & 96.9 & 23.2 & 102.0 & 75.6 \\
\multicolumn{5}{@{}l}{\textbf{Scene clutter}} \\
\quad crowd-like distractors & 68.7 & 12.4 & 103.8 & 73.3 \\
\quad object distractors & 75.8 & 11.5 & 111.3 & 69.5 \\
\quad texture distractors & 76.4 & 11.9 & 102.9 & 76.8 \\
\multicolumn{5}{@{}l}{\textbf{Visibility loss}} \\
\quad field-of-view loss & 74.9 & 12.6 & 95.4 & 66.9 \\
\quad foreground occlusion & 73.3 & 12.8 & 103.9 & 73.6 \\
\quad viewpoint shift & 99.6 & 16.4 & 146.1 & 100.9 \\
\bottomrule
\end{tabular}
\vspace{-0.7em}
\end{table}

\subsection{Output representation}

\Cref{tab:representation-family} summarizes relative degradation by output representation and corruption family. Because the groups differ in size, architecture, and training procedure, the results characterize the evaluated models.

\begin{table}[t]
\centering
\caption{\textbf{FP32 relative degradation by output representation.}
Median RD across models within each output representation, reported separately for each corruption family.
\textbf{Bold} indicates the lowest median within each family.
Results are descriptive; the direct-count group comprises two Transcrowd variants.}
\label{tab:representation-family}
\scriptsize
\setlength{\tabcolsep}{5pt}
\renewcommand{\arraystretch}{1.05}
\begin{tabular}{@{}l r rrrrrr@{}}
\toprule
Output representation & $n$ & \rotatebox{90}{Environment} & \rotatebox{90}{Optics \& motion} & \rotatebox{90}{Sensor \& exposure} & \rotatebox{90}{Encoding \& transport} & \rotatebox{90}{Scene clutter} & \rotatebox{90}{Visibility loss} \\
\midrule
\textbf{D}~Density map & 15 & 38.2 & 60.4 & 68.7 & 38.0 & 14.9 & 22.4 \\
\textbf{B}~Blockwise count map & 5 & 44.2 & 56.9 & 53.6 & 26.7 & 19.9 & 25.5 \\
\textbf{P}~Localization-oriented & 4 & 51.0 & 66.3 & 86.8 & 32.0 & \textbf{11.9} & \textbf{21.8} \\
\textbf{C}~Direct count & 2 & \textbf{27.7} & \textbf{37.3} & \textbf{20.8} & \textbf{18.2} & 17.4 & 22.5 \\
\bottomrule
\end{tabular}
\end{table}

\subsection{Crowd-size analysis}

\Cref{tab:crowd-size-detail} reports absolute and population-normalized corruption error by crowd-size quartile and dataset, including the mean, median, and 90th percentile of the per-image change. Median and 90th-percentile values are included to characterize the heavy upper tail produced by occasional large prediction errors.

\begin{table}[htbp]
\centering
\caption{\textbf{FP32 robustness by crowd-size quartile.}
S/M/D/VD denote sparse, medium, dense, and very dense scenes.
Statistics are aggregated across all models, corruptions, and severity levels.
Median and 90th-percentile values characterize typical and upper-tail behavior, respectively.}
\label{tab:crowd-size-detail}
\scriptsize
\setlength{\tabcolsep}{1.0pt}
\renewcommand{\arraystretch}{0.84}
\begin{tabular}{@{}llrrrrrr@{}}
\toprule
Data & Q & clean & mean $\Delta$ & med. & p90 & RD & $\Delta$nMAE \\
\midrule
\multirow{4}{*}{SHA} & S & 24.1 & 6.3 & 2.0 & 27.9 & 26.2 & 4.4 \\
 & M & 43.8 & 7.7 & 2.9 & 42.8 & 17.7 & 3.2 \\
 & D & 60.0 & 13.0 & 5.0 & 66.1 & 21.7 & 3.4 \\
 & VD & 125.8 & 33.4 & 12.0 & 142.9 & 26.5 & 3.7 \\
\addlinespace[1pt]
\multirow{4}{*}{SHB} & S & 2.7 & 3.8 & 1.3 & 12.0 & 141.1 & 10.1 \\
 & M & 5.2 & 5.5 & 2.0 & 19.0 & 105.0 & 8.0 \\
 & D & 8.7 & 10.0 & 3.7 & 33.2 & 114.9 & 8.0 \\
 & VD & 16.2 & 18.6 & 6.9 & 59.0 & 115.2 & 7.7 \\
\addlinespace[1pt]
\multirow{4}{*}{QNRF} & S & 27.5 & 5.8 & 1.7 & 27.7 & 21.0 & 3.6 \\
 & M & 43.1 & 11.3 & 3.8 & 46.0 & 26.3 & 3.6 \\
 & D & 93.5 & 15.8 & 5.3 & 76.0 & 16.8 & 2.6 \\
 & VD & 218.7 & 41.9 & 11.2 & 182.3 & 19.1 & 2.5 \\
\addlinespace[1pt]
\multirow{4}{*}{JHU} & S & 11.3 & 3.8 & 0.3 & 7.7 & 34.1 & 16.7 \\
 & M & 18.2 & 4.3 & 1.0 & 15.7 & 23.6 & 6.0 \\
 & D & 37.1 & 6.8 & 2.0 & 32.7 & 18.4 & 3.8 \\
 & VD & 197.3 & 34.4 & 8.2 & 142.4 & 17.4 & 3.5 \\
\bottomrule
\end{tabular}

\end{table}

\subsection{Localization analysis}

\Cref{tab:localization-family,tab:localization-degradation} provide complementary views of localization robustness for APGCC, P2PNet, PET, and FIDTM, the four models with native point outputs. \Cref{tab:localization-family} summarizes degradation by corruption family, while \cref{tab:localization-degradation} reports the corresponding dataset-level results. Localization metrics are computed within each dataset using its matching convention before aggregation, since raw $F_1$ values are not directly comparable across datasets.

\begin{table}[htbp]
\centering
\caption{\textbf{Localization degradation by dataset.}
Clean $F_1$ scores and mean corruption-induced degradation across four native-point models.
FP32 precision and recall changes are reported separately; FP16 and INT8 exhibit similar $F_1$ degradation.}
\label{tab:localization-degradation}
\scriptsize
\setlength{\tabcolsep}{3.0pt}
\renewcommand{\arraystretch}{0.92}
\begin{tabular}{@{}l c ccc cc@{}}
\toprule
& clean & \multicolumn{3}{c}{$F_1$ degradation} & \multicolumn{2}{c}{FP32 drop} \\
\cmidrule(lr){3-5}\cmidrule(l){6-7}
Dataset & $F_1$ & FP32 & FP16 & INT8 & precision & recall \\
\midrule
SHA & 0.726 & 0.034 & 0.033 & 0.032 & 0.030 & 0.036 \\
SHB & 0.826 & 0.051 & 0.051 & 0.050 & 0.003 & 0.088 \\
QNRF & 0.739 & 0.027 & 0.027 & 0.027 & 0.014 & 0.037 \\
JHU & 0.685 & 0.029 & 0.029 & 0.029 & 0.005 & 0.047 \\
\bottomrule
\end{tabular}

\end{table}

\begin{table}[htbp]
\centering
\caption{\textbf{Localization degradation by corruption family.}
Mean within-dataset $F_1$ degradation across four native-point models.
FP32 precision and recall changes indicate that most corruption families primarily reduce recall.}
\label{tab:localization-family}
\scriptsize
\setlength{\tabcolsep}{3.4pt}
\renewcommand{\arraystretch}{0.92}
\begin{tabular}{@{}l ccc cc@{}}
\toprule
& \multicolumn{3}{c}{$F_1$ degradation} & \multicolumn{2}{c}{FP32 drop} \\
\cmidrule(lr){2-4}\cmidrule(l){5-6}
Family & FP32 & FP16 & INT8 & precision & recall \\
\midrule
Sensor \& exposure & 0.054 & 0.054 & 0.052 & 0.015 & 0.085 \\
Optics \& motion & 0.034 & 0.034 & 0.034 & -0.001 & 0.062 \\
Environment & 0.033 & 0.033 & 0.033 & 0.018 & 0.043 \\
Encoding \& transport & 0.030 & 0.030 & 0.030 & 0.007 & 0.050 \\
Scene clutter & 0.026 & 0.026 & 0.025 & 0.030 & 0.021 \\
Visibility loss & 0.023 & 0.023 & 0.023 & 0.011 & 0.033 \\
\bottomrule
\end{tabular}

\end{table}

\section{Precision-Dependent Robustness}
\label{sec:supp-precision}

\Cref{tab:precision-sensitivity,fig:int8-sensitivity} report corruption-specific $\drd$ relative to the FP16 control envelope, highlighting where INT8 produces the largest shifts in relative degradation. For each model--dataset pair, $\drd$ is measured against FP32, with RD normalized by the clean error at the same precision. This isolates changes in corruption response from changes in clean accuracy.

\begin{table}[htbp]
\centering
\caption{\textbf{Precision sensitivity by corruption.} $\Delta_{\rm RD}$ across 104 paired model--dataset evaluations.}
\label{tab:precision-sensitivity}
\footnotesize
\renewcommand{\arraystretch}{0.9}
\setlength{\tabcolsep}{2.5pt}
\resizebox{\columnwidth}{!}{%
\begin{tabular}{@{}l rr r rr l@{}}
\toprule
& \multicolumn{1}{c}{FP16} & \multicolumn{4}{c}{INT8} & \\
\cmidrule(lr){2-2}\cmidrule(lr){3-6}
Corruption & med $\Delta_{\rm RD}$ & med $\Delta_{\rm RD}$ & med $|\Delta|$ & p10 & p90 & largest comparison \\
\midrule
\multicolumn{7}{@{}l}{\textit{Sensor \& exposure}}\\
\quad Gaussian noise & -0.03 & +0.58 & 5.91 & -13.2 & +20.4 & ZIP-Small\,/\,JHU \\
\quad vignetting & +0.01 & -0.55 & 2.80 & -16.1 & +4.6 & ZIP-Small\,/\,SHB \\
\quad highlight clipping & +0.01 & -0.52 & 3.66 & -12.8 & +9.9 & ZIP-Pico\,/\,SHB \\
\quad low light & -0.01 & -0.23 & 3.67 & -14.1 & +8.5 & GCHF\,/\,SHB \\
\quad fixed-pattern noise & -0.02 & -0.21 & 3.30 & -11.5 & +8.8 & ZIP-Nano\,/\,SHB \\
\multicolumn{7}{@{}l}{\textit{Optics \& motion}}\\
\quad lens flare & +0.01 & -1.39 & 2.87 & -15.6 & +3.9 & ZIP-Pico\,/\,SHB \\
\quad far-field defocus & -0.01 & -1.08 & 6.01 & -23.0 & +10.3 & ZIP-Small\,/\,SHB \\
\quad camera motion blur & +0.00 & -1.04 & 3.54 & -16.7 & +6.6 & CHF\,/\,SHB \\
\quad lens contamination & -0.00 & -0.18 & 1.84 & -6.4 & +4.5 & GCHF\,/\,SHB \\
\multicolumn{7}{@{}l}{\textit{Environment}}\\
\quad snow particles & -0.01 & -1.34 & 3.69 & -14.7 & +6.5 & ZIP-Pico\,/\,SHB \\
\quad depth-aware haze & -0.00 & -0.88 & 3.86 & -20.4 & +5.2 & ZIP-Pico\,/\,SHB \\
\quad rain streaks & -0.01 & -0.71 & 2.33 & -8.6 & +4.2 & ZIP-Pico\,/\,SHB \\
\quad cast shadow & +0.00 & -0.23 & 1.07 & -9.4 & +2.7 & FFNet-MobileNetV3\,/\,QNRF \\
\multicolumn{7}{@{}l}{\textit{Encoding \& transport}}\\
\quad head-spacing loss & +0.01 & -0.88 & 2.93 & -9.4 & +7.4 & ZIP-Pico\,/\,SHB \\
\quad moir\'e aliasing & -0.00 & -0.24 & 2.90 & -9.4 & +8.7 & CCTrans-Full\,/\,SHB \\
\quad packet-loss concealment & -0.01 & +0.13 & 2.73 & -9.3 & +5.1 & ZIP-Small\,/\,SHB \\
\quad H.264 compression & -0.01 & -0.01 & 3.82 & -9.2 & +9.9 & CCTrans-Full\,/\,SHB \\
\multicolumn{7}{@{}l}{\textit{Scene clutter}}\\
\quad texture distractors & -0.01 & +0.80 & 2.05 & -3.8 & +6.5 & MPCount\,/\,QNRF \\
\quad crowd-like distractors & +0.00 & +0.45 & 1.71 & -2.6 & +5.4 & CCTrans-Full\,/\,SHB \\
\quad object distractors & -0.01 & +0.34 & 3.52 & -11.0 & +9.7 & CCTrans-Full\,/\,SHB \\
\multicolumn{7}{@{}l}{\textit{Visibility loss}}\\
\quad viewpoint shift & -0.01 & -0.39 & 3.25 & -10.1 & +7.6 & GCHF\,/\,SHB \\
\quad field-of-view loss & -0.01 & +0.30 & 1.93 & -4.4 & +8.7 & MCNN\,/\,SHB \\
\quad foreground occlusion & -0.00 & +0.19 & 1.41 & -4.2 & +3.5 & ZIP-Tiny\,/\,JHU \\
\bottomrule
\end{tabular}}

\end{table}

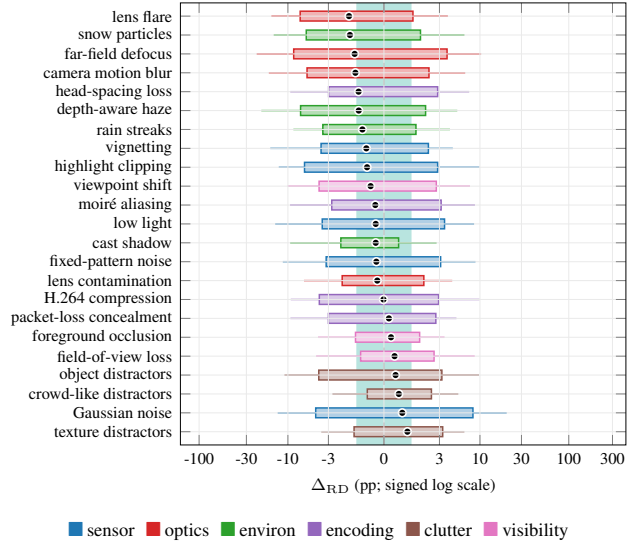
\begin{figure}[htbp]
\centering
\begin{tikzpicture}
\begin{axis}[
  width=0.99\linewidth, height=8.2cm,
  xmin=-2.207, xmax=2.624, ymin=0.3, ymax=23.7,
  xlabel={$\Delta_{\rm RD}$ (pp; signed log scale)},
  xlabel style={font=\scriptsize}, tick label style={font=\scriptsize},
  xtick={-2.0043,-1.4914,-1.0414,-0.6021,0.0000,0.6021,1.0414,1.4914,2.0043,2.4786},
  xticklabels={-100,-30,-10,-3,0,3,10,30,100,300},
  ytick={23,22,21,20,19,18,17,16,15,14,13,12,11,10,9,8,7,6,5,4,3,2,1},
  yticklabels={{lens flare},{snow particles},{far-field defocus},{camera motion blur},{head-spacing loss},{depth-aware haze},{rain streaks},{vignetting},{highlight clipping},{viewpoint shift},{moir\'e aliasing},{low light},{cast shadow},{fixed-pattern noise},{lens contamination},{H.264 compression},{packet-loss concealment},{foreground occlusion},{field-of-view loss},{object distractors},{crowd-like distractors},{Gaussian noise},{texture distractors}},
  yticklabel style={font=\scriptsize},
  grid=major, grid style={rulecol!18}, axis on top,
]
\addplot[draw=none, fill=fp16col!30, forget plot] coordinates {(-0.2993,0.3) (0.2993,0.3) (0.2993,23.7) (-0.2993,23.7)} \closedcycle;
\draw[rulecol, dashed] (axis cs:0,0.3) -- (axis cs:0,23.7);
\draw[grpB!85, line width=0.6pt] (-1.2206,23) -- (0.6936,23);
\fill[grpB!35] (-0.9083,22.74) rectangle (0.3148,23.26);
\draw[grpB, line width=0.6pt] (-0.9083,22.74) rectangle (0.3148,23.26);
\fill[white] (-0.3786,23) circle (2.0pt);
\fill[black] (-0.3786,23) circle (1.3pt);
\draw[grpC!85, line width=0.6pt] (-1.1954,22) -- (0.8731,22);
\fill[grpC!35] (-0.8423,21.74) rectangle (0.3964,22.26);
\draw[grpC, line width=0.6pt] (-0.8423,21.74) rectangle (0.3964,22.26);
\fill[white] (-0.3687,22) circle (2.0pt);
\fill[black] (-0.3687,22) circle (1.3pt);
\draw[grpB!85, line width=0.6pt] (-1.3800,21) -- (1.0542,21);
\fill[grpB!35] (-0.9786,20.74) rectangle (0.6846,21.26);
\draw[grpB, line width=0.6pt] (-0.9786,20.74) rectangle (0.6846,21.26);
\fill[white] (-0.3176,21) circle (2.0pt);
\fill[black] (-0.3176,21) circle (1.3pt);
\draw[grpB!85, line width=0.6pt] (-1.2486,20) -- (0.8816,20);
\fill[grpB!35] (-0.8338,19.74) rectangle (0.4886,20.26);
\draw[grpB, line width=0.6pt] (-0.8338,19.74) rectangle (0.4886,20.26);
\fill[white] (-0.3094,20) circle (2.0pt);
\fill[black] (-0.3094,20) circle (1.3pt);
\draw[grpD!85, line width=0.6pt] (-1.0158,19) -- (0.9244,19);
\fill[grpD!35] (-0.6001,18.74) rectangle (0.5824,19.26);
\draw[grpD, line width=0.6pt] (-0.6001,18.74) rectangle (0.5824,19.26);
\fill[white] (-0.2748,19) circle (2.0pt);
\fill[black] (-0.2748,19) circle (1.3pt);
\draw[grpC!85, line width=0.6pt] (-1.3298,18) -- (0.7957,18);
\fill[grpC!35] (-0.9039,17.74) rectangle (0.4518,18.26);
\draw[grpC, line width=0.6pt] (-0.9039,17.74) rectangle (0.4518,18.26);
\fill[white] (-0.2733,18) circle (2.0pt);
\fill[black] (-0.2733,18) circle (1.3pt);
\draw[grpC!85, line width=0.6pt] (-0.9822,17) -- (0.7164,17);
\fill[grpC!35] (-0.6624,16.74) rectangle (0.3473,17.26);
\draw[grpC, line width=0.6pt] (-0.6624,16.74) rectangle (0.3473,17.26);
\fill[white] (-0.2335,17) circle (2.0pt);
\fill[black] (-0.2335,17) circle (1.3pt);
\draw[grpA!85, line width=0.6pt] (-1.2328,16) -- (0.7476,16);
\fill[grpA!35] (-0.6819,15.74) rectangle (0.4822,16.26);
\draw[grpA, line width=0.6pt] (-0.6819,15.74) rectangle (0.4822,16.26);
\fill[white] (-0.1908,16) circle (2.0pt);
\fill[black] (-0.1908,16) circle (1.3pt);
\draw[grpA!85, line width=0.6pt] (-1.1405,15) -- (1.0357,15);
\fill[grpA!35] (-0.8612,14.74) rectangle (0.5823,15.26);
\draw[grpA, line width=0.6pt] (-0.8612,14.74) rectangle (0.5823,15.26);
\fill[white] (-0.1830,15) circle (2.0pt);
\fill[black] (-0.1830,15) circle (1.3pt);
\draw[grpF!85, line width=0.6pt] (-1.0455,14) -- (0.9331,14);
\fill[grpF!35] (-0.7032,13.74) rectangle (0.5680,14.26);
\draw[grpF, line width=0.6pt] (-0.7032,13.74) rectangle (0.5680,14.26);
\fill[white] (-0.1442,14) circle (2.0pt);
\fill[black] (-0.1442,14) circle (1.3pt);
\draw[grpD!85, line width=0.6pt] (-1.0181,13) -- (0.9869,13);
\fill[grpD!35] (-0.5656,12.74) rectangle (0.6205,13.26);
\draw[grpD, line width=0.6pt] (-0.5656,12.74) rectangle (0.6205,13.26);
\fill[white] (-0.0936,13) circle (2.0pt);
\fill[black] (-0.0936,13) circle (1.3pt);
\draw[grpA!85, line width=0.6pt] (-1.1791,12) -- (0.9778,12);
\fill[grpA!35] (-0.6689,11.74) rectangle (0.6574,12.26);
\draw[grpA, line width=0.6pt] (-0.6689,11.74) rectangle (0.6574,12.26);
\fill[white] (-0.0888,12) circle (2.0pt);
\fill[black] (-0.0888,12) circle (1.3pt);
\draw[grpC!85, line width=0.6pt] (-1.0168,11) -- (0.5712,11);
\fill[grpC!35] (-0.4673,10.74) rectangle (0.1592,11.26);
\draw[grpC, line width=0.6pt] (-0.4673,10.74) rectangle (0.1592,11.26);
\fill[white] (-0.0887,11) circle (2.0pt);
\fill[black] (-0.0887,11) circle (1.3pt);
\draw[grpA!85, line width=0.6pt] (-1.0967,10) -- (0.9931,10);
\fill[grpA!35] (-0.6264,9.74) rectangle (0.6149,10.26);
\draw[grpA, line width=0.6pt] (-0.6264,9.74) rectangle (0.6149,10.26);
\fill[white] (-0.0822,10) circle (2.0pt);
\fill[black] (-0.0822,10) circle (1.3pt);
\draw[grpB!85, line width=0.6pt] (-0.8671,9) -- (0.7408,9);
\fill[grpB!35] (-0.4527,8.74) rectangle (0.4329,9.26);
\draw[grpB, line width=0.6pt] (-0.4527,8.74) rectangle (0.4329,9.26);
\fill[white] (-0.0722,9) circle (2.0pt);
\fill[black] (-0.0722,9) circle (1.3pt);
\draw[grpD!85, line width=0.6pt] (-1.0102,8) -- (1.0370,8);
\fill[grpD!35] (-0.7006,7.74) rectangle (0.5929,8.26);
\draw[grpD, line width=0.6pt] (-0.7006,7.74) rectangle (0.5929,8.26);
\fill[white] (-0.0050,8) circle (2.0pt);
\fill[black] (-0.0050,8) circle (1.3pt);
\draw[grpD!85, line width=0.6pt] (-1.0144,7) -- (0.7865,7);
\fill[grpD!35] (-0.6022,6.74) rectangle (0.5629,7.26);
\draw[grpD, line width=0.6pt] (-0.6022,6.74) rectangle (0.5629,7.26);
\fill[white] (0.0540,7) circle (2.0pt);
\fill[black] (0.0540,7) circle (1.3pt);
\draw[grpF!85, line width=0.6pt] (-0.7146,6) -- (0.6573,6);
\fill[grpF!35] (-0.3083,5.74) rectangle (0.3873,6.26);
\draw[grpF, line width=0.6pt] (-0.3083,5.74) rectangle (0.3873,6.26);
\fill[white] (0.0766,6) circle (2.0pt);
\fill[black] (0.0766,6) circle (1.3pt);
\draw[grpF!85, line width=0.6pt] (-0.7357,5) -- (0.9846,5);
\fill[grpF!35] (-0.2528,4.74) rectangle (0.5438,5.26);
\draw[grpF, line width=0.6pt] (-0.2528,4.74) rectangle (0.5438,5.26);
\fill[white] (0.1152,5) circle (2.0pt);
\fill[black] (0.1152,5) circle (1.3pt);
\draw[grpE!85, line width=0.6pt] (-1.0789,4) -- (1.0312,4);
\fill[grpE!35] (-0.7065,3.74) rectangle (0.6284,4.26);
\draw[grpE, line width=0.6pt] (-0.7065,3.74) rectangle (0.6284,4.26);
\fill[white] (0.1263,4) circle (2.0pt);
\fill[black] (0.1263,4) circle (1.3pt);
\draw[grpE!85, line width=0.6pt] (-0.5585,3) -- (0.8068,3);
\fill[grpE!35] (-0.1803,2.74) rectangle (0.5151,3.26);
\draw[grpE, line width=0.6pt] (-0.1803,2.74) rectangle (0.5151,3.26);
\fill[white] (0.1618,3) circle (2.0pt);
\fill[black] (0.1618,3) circle (1.3pt);
\draw[grpA!85, line width=0.6pt] (-1.1523,2) -- (1.3314,2);
\fill[grpA!35] (-0.7399,1.74) rectangle (0.9653,2.26);
\draw[grpA, line width=0.6pt] (-0.7399,1.74) rectangle (0.9653,2.26);
\fill[white] (0.1988,2) circle (2.0pt);
\fill[black] (0.1988,2) circle (1.3pt);
\draw[grpE!85, line width=0.6pt] (-0.6795,1) -- (0.8727,1);
\fill[grpE!35] (-0.3239,0.74) rectangle (0.6375,1.26);
\draw[grpE, line width=0.6pt] (-0.3239,0.74) rectangle (0.6375,1.26);
\fill[white] (0.2546,1) circle (2.0pt);
\fill[black] (0.2546,1) circle (1.3pt);
\end{axis}
\end{tikzpicture}

\vspace{1pt}
{\scriptsize \textcolor{grpA}{$\blacksquare$}\,sensor\hspace{4pt}\textcolor{grpB}{$\blacksquare$}\,optics\hspace{4pt}\textcolor{grpC}{$\blacksquare$}\,environ\hspace{4pt}\textcolor{grpD}{$\blacksquare$}\,encoding\hspace{4pt}\textcolor{grpE}{$\blacksquare$}\,clutter\hspace{4pt}\textcolor{grpF}{$\blacksquare$}\,visibility}
\caption{\textbf{Corruption-specific INT8 robustness shifts.}
Distributions of $\Delta_{\rm RD}$ across 104 model--dataset evaluations per corruption.
Dots indicate medians, boxes the interquartile range (IQR), and whiskers the 10th--90th percentiles.
The shaded band denotes the FP16 control range ($\pm 0.99$ percentage points); colors distinguish corruption families.}
\label{fig:int8-sensitivity}
\end{figure}

\section{Deployment Performance}
\label{sec:supp-deployment}
\Cref{tab:edge-full-SHA,tab:edge-full-SHB,tab:edge-full-QNRF,tab:edge-full-JHU} 
report energy per image and throughput
for every model--precision configuration on each dataset.
Runtime decomposition is reported separately in
\cref{fig:runtime-stages-detail}.

\begin{table}[htbp]
\centering
\caption{\textbf{Deployment performance on ShanghaiTech~A.}
Energy and throughput at FP32, FP16, and INT8 precision.
\worse{Red} highlights INT8 regressions relative to FP16.}

\label{tab:edge-full-SHA}
\scriptsize
\setlength{\tabcolsep}{2.35pt}
\renewcommand{\arraystretch}{0.9}
\begin{tabular}{@{}l r rr rr rr@{}}
\toprule
& & \multicolumn{2}{c}{FP32} & \multicolumn{2}{c}{FP16} & \multicolumn{2}{c}{INT8} \\
\cmidrule(lr){3-4}\cmidrule(lr){5-6}\cmidrule(lr){7-8}
Model & Par. & J/img & img/s & J/img & img/s & J/img & img/s \\
& (M) & & & & & & \\
\midrule
\multicolumn{8}{@{}l}{\emph{CNN-based}} \\
\quad APGCC & 18.7 & 2.49 & 9.6 & 1.17 & 18.3 & 0.66 & 23.6 \\
\quad CHF & 21.5 & 2.36 & 10.4 & 1.07 & 20.7 & 0.52 & 29.1 \\
\quad GCHF & 21.5 & 2.37 & 10.4 & 1.07 & 20.7 & 0.52 & 29.1 \\
\quad CSRNet & 16.3 & 2.35 & 11.1 & 1.03 & 23.8 & 0.47 & 35.5 \\
\quad DM-Count & 21.5 & 2.36 & 11.0 & 1.02 & 23.5 & 0.46 & 35.5 \\
\quad FFNet-ConvNeXt-T & 34.6 & 1.81 & 10.8 & 1.06 & 15.0 & \worse{1.10} & \worse{14.4} \\
\quad FFNet-MobileNetV3 & 3.3 & 0.52 & 22.3 & 0.43 & 24.3 & \worse{0.45} & \worse{23.0} \\
\quad FIDTM & 66.6 & 3.83 & 5.8 & 1.60 & 13.4 & 1.07 & 17.0 \\
\quad MCNN & 0.1 & 0.48 & 33.6 & 0.28 & 51.5 & \worse{0.35} & \worse{43.5} \\
\quad MobileCount & 3.4 & 0.32 & 43.9 & 0.22 & 54.6 & 0.20 & 56.4 \\
\quad MPCount & 33.2 & 5.80 & 3.1 & 3.09 & 5.0 & 2.21 & 5.7 \\
\quad P2PNet & 21.6 & 2.39 & 10.2 & 1.09 & 19.7 & 0.54 & 26.8 \\
\quad STEERER-HRNet & 64.6 & 14.47 & 1.5 & 6.31 & 3.2 & 4.36 & 3.7 \\
\quad STEERER-VGG19 & 30.6 & 13.00 & 1.8 & 5.72 & 3.5 & 3.29 & 4.4 \\
\quad ZIP-Nano & 3.4 & 0.48 & 28.1 & 0.33 & 33.8 & 0.31 & 35.1 \\
\quad ZIP-Pico & 0.8 & 0.33 & 33.7 & 0.27 & 37.2 & \worse{0.28} & \worse{35.6} \\
\quad ZIP-Small & 33.6 & 2.86 & 7.9 & 1.53 & 12.7 & 1.33 & \worse{12.5} \\
\quad ZIP-Tiny & 10.5 & 0.83 & 19.6 & 0.48 & 28.1 & 0.39 & 30.4 \\
\multicolumn{8}{@{}l}{\emph{Transformer}} \\
\quad CCST & 300.7 & 129.95 & 0.2 & 40.74 & 0.8 & 20.39 & 1.4 \\
\quad CCTrans-Full & 104.6 & 4.63 & 5.1 & 2.10 & 10.2 & 1.75 & 10.8 \\
\quad CCTrans-Weak & 104.6 & 4.64 & 5.1 & 2.09 & 10.3 & 1.84 & 10.7 \\
\quad Transcrowd-GAP & 90.4 & 6.19 & 4.4 & 2.14 & 12.0 & 1.59 & 13.7 \\
\quad Transcrowd-Token & 86.9 & 6.20 & 4.4 & 2.13 & 11.9 & 1.60 & 13.5 \\
\multicolumn{8}{@{}l}{\emph{Hybrid}} \\
\quad Gramformer & 29.0 & 4.06 & 5.6 & 2.01 & 10.9 & 1.34 & 13.9 \\
\quad PET & 20.9 & 6.25 & 4.2 & 2.75 & 8.9 & 1.80 & 11.2 \\
\multicolumn{8}{@{}l}{\emph{Vision--language}} \\
\quad CLIP-EBC-RN50 & 135.5 & 19.05 & 1.6 & 4.51 & 6.2 & 1.96 & 11.7 \\
\bottomrule
\end{tabular}

\end{table}

\begin{table}[htbp]
\centering
\caption{\textbf{Deployment performance on ShanghaiTech~B.}
Energy and throughput at FP32, FP16, and INT8 precision.
\worse{Red} highlights INT8 regressions relative to FP16.}

\label{tab:edge-full-SHB}
\scriptsize
\setlength{\tabcolsep}{2.35pt}
\renewcommand{\arraystretch}{0.9}
\begin{tabular}{@{}l r rr rr rr@{}}
\toprule
& & \multicolumn{2}{c}{FP32} & \multicolumn{2}{c}{FP16} & \multicolumn{2}{c}{INT8} \\
\cmidrule(lr){3-4}\cmidrule(lr){5-6}\cmidrule(lr){7-8}
Model & Par. & J/img & img/s & J/img & img/s & J/img & img/s \\
& (M) & & & & & & \\
\midrule
\multicolumn{8}{@{}l}{\emph{CNN-based}} \\
\quad APGCC & 18.7 & 3.35 & 7.3 & 1.53 & 14.1 & 0.86 & 18.1 \\
\quad CHF & 21.5 & 3.54 & 7.1 & 1.59 & 14.5 & 0.73 & 20.9 \\
\quad GCHF & 21.5 & 3.54 & 7.1 & 1.59 & 14.6 & 0.73 & 20.9 \\
\quad CSRNet & 16.3 & 3.52 & 7.6 & 1.51 & 16.8 & 0.67 & 25.6 \\
\quad DM-Count & 21.5 & 3.48 & 7.6 & 1.52 & 16.7 & 0.64 & 26.0 \\
\quad FFNet-ConvNeXt-T & 34.6 & 2.73 & 7.7 & 1.55 & 11.3 & \worse{1.58} & \worse{11.1} \\
\quad FFNet-MobileNetV3 & 3.3 & 0.79 & 15.7 & 0.61 & 18.0 & \worse{0.62} & \worse{17.5} \\
\quad FIDTM & 66.6 & 5.52 & 4.1 & 2.27 & 10.0 & 1.50 & 12.7 \\
\quad MCNN & 0.1 & 0.70 & 24.2 & 0.40 & 38.0 & \worse{0.43} & \worse{34.7} \\
\quad MobileCount & 3.4 & 0.47 & 31.7 & 0.32 & 39.9 & 0.27 & 41.9 \\
\quad MPCount & 33.2 & 7.76 & 2.2 & 4.01 & 3.8 & 2.83 & 4.3 \\
\quad P2PNet & 21.6 & 3.17 & 7.7 & 1.46 & 14.9 & 0.72 & 20.3 \\
\quad STEERER-HRNet & 64.6 & 5.19 & 4.2 & 2.27 & 8.9 & 1.52 & 10.7 \\
\quad STEERER-VGG19 & 30.6 & 4.51 & 5.1 & 2.03 & 10.0 & 1.15 & 12.6 \\
\quad ZIP-Nano & 3.4 & 0.61 & 21.7 & 0.43 & 25.8 & 0.40 & 27.1 \\
\quad ZIP-Pico & 0.8 & 0.43 & 25.9 & 0.35 & 28.5 & 0.35 & 28.7 \\
\quad ZIP-Small & 33.6 & 3.64 & 6.2 & 1.96 & 9.9 & 1.71 & \worse{9.8} \\
\quad ZIP-Tiny & 10.5 & 1.06 & 15.6 & 0.61 & 22.0 & 0.51 & 23.6 \\
\multicolumn{8}{@{}l}{\emph{Transformer}} \\
\quad CCST & 300.7 & 130.05 & 0.2 & 41.27 & 0.8 & 20.53 & 1.4 \\
\quad CCTrans-Full & 104.6 & 8.40 & 3.1 & 3.39 & 7.1 & 2.83 & 7.3 \\
\quad CCTrans-Weak & 104.6 & 8.40 & 3.1 & 3.36 & 7.1 & 2.97 & 7.3 \\
\quad Transcrowd-GAP & 90.4 & 5.71 & 4.7 & 2.31 & 10.9 & 1.66 & 12.3 \\
\quad Transcrowd-Token & 86.9 & 5.74 & 4.8 & 2.32 & 11.0 & 1.67 & 12.4 \\
\multicolumn{8}{@{}l}{\emph{Hybrid}} \\
\quad Gramformer & 29.0 & 5.44 & 4.6 & 2.58 & 9.1 & 1.61 & 11.5 \\
\quad PET & 20.9 & 7.49 & 3.5 & 3.22 & 7.6 & 2.18 & 9.3 \\
\multicolumn{8}{@{}l}{\emph{Vision--language}} \\
\quad CLIP-EBC-RN50 & 135.5 & 24.05 & 1.2 & 5.75 & 4.8 & 2.53 & 9.1 \\
\bottomrule
\end{tabular}

\end{table}

\begin{table}[htbp]
\centering
\caption{\textbf{Deployment performance on UCF-QNRF.}
Energy and throughput at FP32, FP16, and INT8 precision.
\worse{Red} highlights INT8 regressions relative to FP16.}

\label{tab:edge-full-QNRF}
\scriptsize
\setlength{\tabcolsep}{2.35pt}
\renewcommand{\arraystretch}{0.9}
\begin{tabular}{@{}l r rr rr rr@{}}
\toprule
& & \multicolumn{2}{c}{FP32} & \multicolumn{2}{c}{FP16} & \multicolumn{2}{c}{INT8} \\
\cmidrule(lr){3-4}\cmidrule(lr){5-6}\cmidrule(lr){7-8}
Model & Par. & J/img & img/s & J/img & img/s & J/img & img/s \\
& (M) & & & & & & \\
\midrule
\multicolumn{8}{@{}l}{\emph{CNN-based}} \\
\quad APGCC & 18.7 & 12.35 & 1.6 & 6.46 & 2.4 & 4.28 & 2.8 \\
\quad CHF & 21.5 & 14.75 & 1.5 & 7.29 & 2.4 & 4.19 & 2.9 \\
\quad GCHF & 21.5 & 14.74 & 1.5 & 7.23 & 2.4 & 4.16 & 2.9 \\
\quad CSRNet & 16.3 & 12.85 & 1.7 & 6.34 & 2.7 & 3.65 & 3.4 \\
\quad DM-Count & 21.5 & 12.70 & 1.7 & 6.29 & 2.7 & 3.59 & 3.4 \\
\quad FFNet-ConvNeXt-T & 34.6 & 24.85 & 0.8 & 14.28 & 1.2 & \worse{14.43} & \worse{1.2} \\
\quad FFNet-MobileNetV3 & 3.3 & 7.63 & 1.6 & 5.81 & 1.9 & \worse{5.93} & \worse{1.8} \\
\quad FIDTM & 66.6 & 19.74 & 1.0 & 8.92 & 2.0 & 6.47 & 2.3 \\
\quad MCNN & 0.1 & 3.79 & 3.3 & 2.89 & 3.8 & \worse{3.14} & \worse{3.6} \\
\quad MobileCount & 3.4 & 3.08 & 3.6 & 2.57 & 4.0 & 2.46 & 4.0 \\
\quad MPCount & 33.2 & 49.79 & 0.3 & 28.85 & 0.5 & 21.89 & 0.5 \\
\quad P2PNet & 21.6 & 7.07 & 2.5 & 4.02 & 3.5 & 2.80 & 3.9 \\
\quad STEERER-HRNet & 64.6 & 32.43 & 0.6 & 14.84 & 1.2 & 10.60 & 1.4 \\
\quad STEERER-VGG19 & 30.6 & 16.52 & 1.2 & 8.21 & 2.0 & 5.49 & 2.2 \\
\quad ZIP-Nano & 3.4 & 6.22 & 2.1 & 4.49 & 2.5 & 4.21 & 2.5 \\
\quad ZIP-Pico & 0.8 & 4.13 & 2.6 & 3.75 & 2.7 & 3.62 & 2.7 \\
\quad ZIP-Small & 33.6 & 32.80 & 0.7 & 16.80 & 1.2 & 14.81 & \worse{1.1} \\
\quad ZIP-Tiny & 10.5 & 7.95 & 1.8 & 5.24 & 2.3 & 4.44 & 2.5 \\
\multicolumn{8}{@{}l}{\emph{Transformer}} \\
\quad CCST & 300.7 & 130.90 & 0.2 & 41.85 & 0.7 & 21.28 & 1.2 \\
\quad CCTrans-Full & 104.6 & 75.74 & 0.3 & 30.94 & 0.8 & 25.88 & 0.8 \\
\quad CCTrans-Weak & 104.6 & 75.42 & 0.3 & 30.65 & 0.8 & 26.97 & 0.8 \\
\quad Transcrowd-GAP & 90.4 & 26.40 & 1.0 & 9.52 & 2.4 & 7.26 & 2.7 \\
\quad Transcrowd-Token & 86.9 & 25.78 & 1.0 & 9.46 & 2.4 & 7.17 & 2.7 \\
\multicolumn{8}{@{}l}{\emph{Hybrid}} \\
\quad Gramformer & 29.0 & 39.77 & 0.6 & 20.43 & 1.0 & 14.10 & 1.3 \\
\quad PET & 20.9 & 16.93 & 1.3 & 8.33 & 2.3 & 6.01 & 2.6 \\
\multicolumn{8}{@{}l}{\emph{Vision--language}} \\
\quad CLIP-EBC-RN50 & 135.5 & 184.48 & 0.2 & 44.69 & 0.6 & 20.03 & 1.1 \\
\bottomrule
\end{tabular}

\end{table}

\begin{table}[htbp]
\centering
\caption{\textbf{Deployment performance on JHU-Crowd++.}
Energy and throughput at FP32, FP16, and INT8 precision.
\worse{Red} highlights INT8 regressions relative to FP16.}
\label{tab:edge-full-JHU}
\scriptsize
\setlength{\tabcolsep}{2.35pt}
\renewcommand{\arraystretch}{0.9}
\begin{tabular}{@{}l r rr rr rr@{}}
\toprule
& & \multicolumn{2}{c}{FP32} & \multicolumn{2}{c}{FP16} & \multicolumn{2}{c}{INT8} \\
\cmidrule(lr){3-4}\cmidrule(lr){5-6}\cmidrule(lr){7-8}
Model & Par. & J/img & img/s & J/img & img/s & J/img & img/s \\
& (M) & & & & & & \\
\midrule
\multicolumn{8}{@{}l}{\emph{CNN-based}} \\
\quad APGCC & 18.7 & 4.98 & 4.5 & 2.40 & 7.8 & 1.46 & 9.5 \\
\quad CHF & 21.5 & 6.94 & 3.5 & 3.17 & 6.6 & 1.61 & 8.8 \\
\quad GCHF & 21.5 & 7.01 & 3.5 & 3.18 & 6.6 & 1.59 & 9.0 \\
\quad CSRNet & 16.3 & 4.75 & 5.0 & 2.16 & 9.3 & 1.14 & 12.3 \\
\quad DM-Count & 21.5 & 4.82 & 4.9 & 2.25 & 9.2 & 1.14 & 12.2 \\
\quad FFNet-ConvNeXt-T & 34.6 & 6.57 & 3.2 & 3.76 & 4.6 & \worse{3.80} & \worse{4.6} \\
\quad FFNet-MobileNetV3 & 3.3 & 1.99 & 6.2 & 1.51 & 7.2 & \worse{1.55} & \worse{7.1} \\
\quad FIDTM & 66.6 & 7.75 & 2.7 & 3.34 & 5.9 & 2.34 & 7.0 \\
\quad MCNN & 0.1 & 1.19 & 11.8 & 0.84 & 15.2 & \worse{0.93} & \worse{14.0} \\
\quad MobileCount & 3.4 & 0.91 & 13.9 & 0.71 & 15.9 & 0.65 & 16.3 \\
\quad MPCount & 33.2 & 14.86 & 1.1 & 8.48 & 1.8 & 6.19 & 1.9 \\
\quad P2PNet & 21.6 & 4.79 & 4.7 & 2.32 & 7.9 & 1.31 & 9.9 \\
\quad STEERER-HRNet & 64.6 & 15.79 & 1.4 & 7.11 & 2.8 & 4.81 & 3.3 \\
\quad STEERER-VGG19 & 30.6 & 8.33 & 2.7 & 3.78 & 5.0 & 2.32 & 6.0 \\
\quad ZIP-Nano & 3.4 & 1.13 & 10.1 & 0.96 & 10.9 & 0.89 & 11.3 \\
\quad ZIP-Pico & 0.8 & 0.98 & 10.8 & 0.89 & 11.3 & 0.86 & 11.4 \\
\quad ZIP-Small & 33.6 & 6.48 & 3.4 & 3.49 & 5.1 & 3.39 & \worse{4.8} \\
\quad ZIP-Tiny & 10.5 & 1.86 & 7.8 & 1.25 & 9.7 & 1.06 & 10.4 \\
\multicolumn{8}{@{}l}{\emph{Transformer}} \\
\quad CCST & 300.7 & 130.31 & 0.2 & 41.14 & 0.8 & 20.61 & 1.4 \\
\quad CCTrans-Full & 104.6 & 20.23 & 1.3 & 8.25 & 2.9 & 6.92 & 3.0 \\
\quad CCTrans-Weak & 104.6 & 20.14 & 1.3 & 8.14 & 2.9 & 7.16 & 3.0 \\
\quad Transcrowd-GAP & 90.4 & 15.79 & 1.7 & 5.50 & 4.7 & 4.13 & 5.3 \\
\quad Transcrowd-Token & 86.9 & 15.67 & 1.7 & 5.48 & 4.7 & 4.10 & 5.3 \\
\multicolumn{8}{@{}l}{\emph{Hybrid}} \\
\quad Gramformer & 29.0 & 12.10 & 1.9 & 6.17 & 3.5 & 4.17 & 4.4 \\
\quad PET & 20.9 & 12.62 & 2.0 & 5.70 & 4.1 & 3.86 & 4.9 \\
\multicolumn{8}{@{}l}{\emph{Vision--language}} \\
\quad CLIP-EBC-RN50 & 135.5 & 49.95 & 0.6 & 12.04 & 2.3 & 5.36 & 4.2 \\
\bottomrule
\end{tabular}

\end{table}

\subsection{Runtime decomposition}

\Cref{fig:runtime-stages-detail} decomposes end-to-end inference time into host preprocessing, GPU execution, host--device transfer, dispatch, and engine residual. These components account for the full measured latency.

For most models, host preprocessing dominates end-to-end latency and therefore limits the application-level benefit of faster engine execution. The observed INT8 slowdowns, however, are concentrated in GPU execution rather than transfer or dispatch, indicating that the regressions arise within the compiled engine.

\begin{figure}[htbp]
\centering
\begin{subfigure}{\linewidth}
\centering
\includegraphics[width=\linewidth]{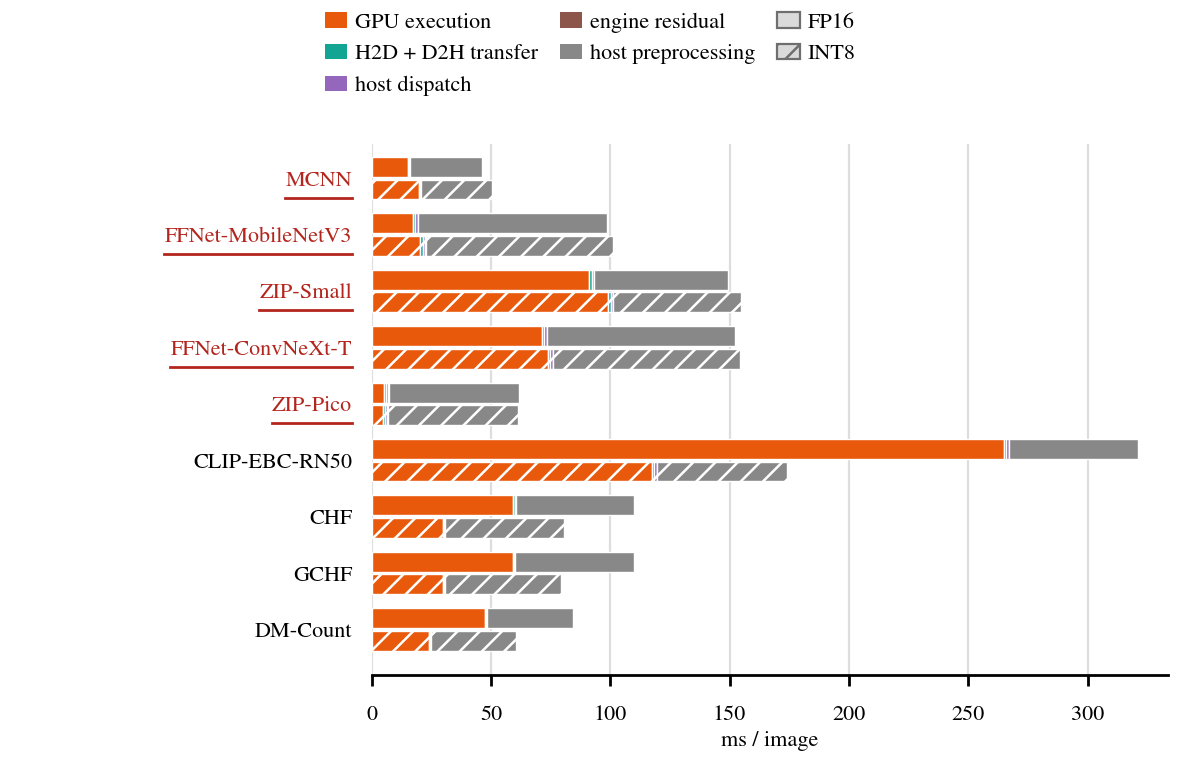}
\caption{End-to-end per image.}
\label{fig:runtime-stages-detail-a}
\end{subfigure}\\[3pt]
\begin{subfigure}{\linewidth}
\centering
\includegraphics[width=\linewidth]{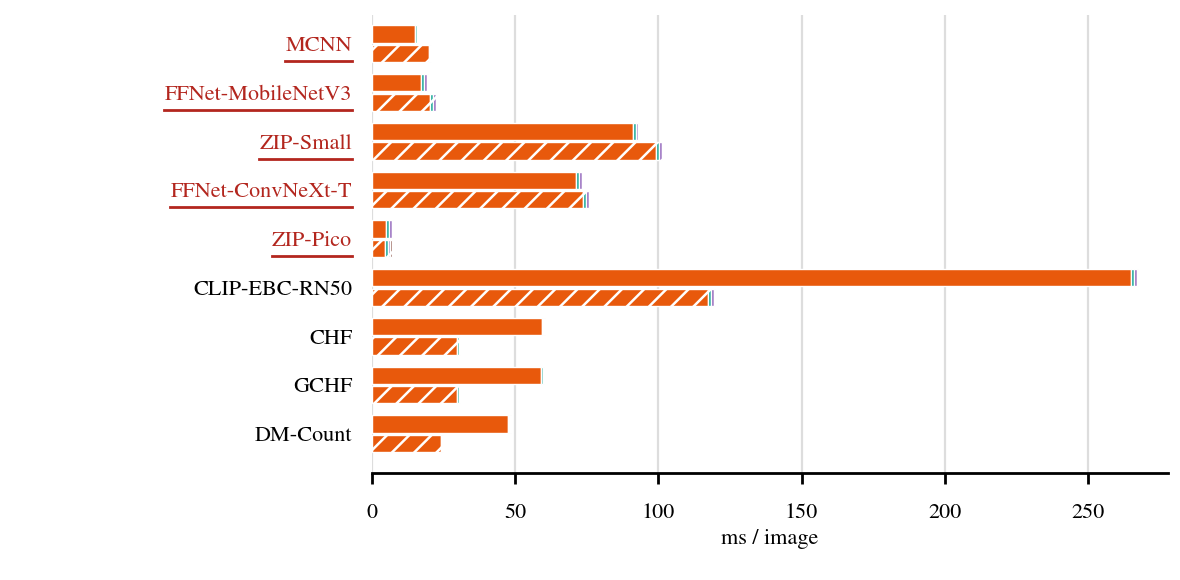}
\caption{Engine only, host preprocessing removed.}
\label{fig:runtime-stages-detail-b}
\end{subfigure}
\caption{\textbf{Runtime decomposition.}
Median stage times across four datasets for five models exhibiting INT8 slowdowns and four faster controls.
Models with slowdowns are highlighted in \textcolor{worsecol}{red}; hatched bars indicate INT8.
The observed slowdowns arise primarily during GPU execution rather than transfer or dispatch.}
\label{fig:runtime-stages-detail}
\end{figure}

\Cref{fig:coverage-risk} complements this analysis with engine-level characteristics relevant to INT8 performance: INT8 MAC coverage, reformat operations per GMAC, convolution alignment, and the GPU share of engine time.

\input{components/supplement/figures/coverage-risk-heatmap}

\subsection{Structural diagnostics}

\Cref{tab:layer-census} summarizes the compiled-engine structure for each model, including total plan layers, arithmetic layers, layers assigned to INT8, and the overall precision mix. The four models that do not improve throughput under INT8 relative to FP16---ZIP-Small, FFNet-ConvNeXt-T, FFNet-MobileNetV3, and MCNN---occupy four of the six lowest positions in INT8 compute coverage, at $52.3\%$, $57.8\%$, $50.8\%$, and $76.9\%$, respectively, compared with a median above $90\%$ across the evaluated models.

\begin{table}[t]
\centering
\caption{\textbf{INT8 layer census.}
Dataset-median A100 build-plan statistics for each model.
Compute coverage denotes the fraction of arithmetic layers assigned to INT8, while precision shares include all plan layers.
Other data types, including INT64 and Boolean, are omitted from the displayed precision shares.
These statistics characterize compiled engine structure rather than device execution.}
\label{tab:layer-census}
\scriptsize
\renewcommand{\arraystretch}{0.9}
\setlength{\tabcolsep}{2.1pt}
\begin{tabular}{@{}l rrr r rrr r@{}}
\toprule
 & \multicolumn{3}{c}{layers} & & \multicolumn{3}{c}{plan layers (\%)} & \\
\cmidrule(lr){2-4}\cmidrule(lr){6-8}
Model & all & comp. & INT8 & cov.\ (\%) & INT8 & FP16 & FP32 & build (s) \\
\midrule
\quad CHF & 27 & 19 & 19 & 100.0 & 93 & 4 & 4 & 44 \\
\quad GCHF & 27 & 19 & 19 & 100.0 & 93 & 4 & 4 & 40 \\
\quad CCST & 349 & 63 & 63 & 100.0 & 43 & 47 & 10 & 136 \\
\quad CSRNet & 23 & 17 & 17 & 100.0 & 91 & 4 & 4 & 11 \\
\quad DM-Count & 34 & 19 & 19 & 100.0 & 75 & 21 & 4 & 14 \\
\quad P2PNet & 43 & 21 & 21 & 100.0 & 67 & 30 & 4 & 30 \\
\quad STEERER-HRNet & 634 & 310 & 309 & 99.7 & 60 & 39 & 0 & 75 \\
\quad FIDTM & 486 & 295 & 293 & 99.3 & 74 & 26 & 0 & 74 \\
\quad CLIP-EBC-RN50 & 74 & 60 & 59 & 98.3 & 87 & 10 & 3 & 20 \\
\quad STEERER-VGG19 & 224 & 54 & 53 & 98.2 & 39 & 60 & 1 & 41 \\
\quad Transcrowd-GAP & 107 & 38 & 37 & 97.4 & 59 & 38 & 3 & 22 \\
\quad Transcrowd-Token & 105 & 26 & 25 & 95.1 & 54 & 44 & 3 & 21 \\
\quad APGCC & 158 & 35 & 33 & 94.3 & 25 & 9 & 54 & 32 \\
\quad PET & 269 & 65 & 61 & 93.8 & 39 & 36 & 5 & 64 \\
\quad ZIP-Pico & 65 & 51 & 46 & 90.2 & 74 & 25 & 1 & 25 \\
\quad MPCount & 56 & 29 & 26 & 89.7 & 77 & 20 & 3 & 27 \\
\quad Gramformer & 123 & 33 & 29 & 88.2 & 47 & 37 & 8 & 67 \\
\quad ZIP-Tiny & 104 & 90 & 78 & 86.7 & 77 & 22 & 1 & 30 \\
\quad ZIP-Nano & 72 & 55 & 47 & 85.5 & 69 & 29 & 1 & 27 \\
\quad CCTrans-Full & 397 & 74 & 58 & 78.4 & 41 & 46 & 13 & 76 \\
\quad MCNN & 32 & 13 & 10 & 76.9 & 68 & 29 & 3 & 14 \\
\quad MobileCount & 124 & 62 & 45 & 72.6 & 40 & 60 & 1 & 39 \\
\quad CCTrans-Weak & 388 & 74 & 49 & 66.2 & 39 & 48 & 13 & 110 \\
\quad FFNet-ConvNeXt-T & 364 & 109 & 63 & 57.8 & 25 & 71 & 3 & 49 \\
\quad ZIP-Small & 390 & 193 & 101 & 52.3 & 31 & 68 & 1 & 73 \\
\quad FFNet-MobileNetV3 & 342 & 130 & 66 & 50.8 & 32 & 67 & 1 & 69 \\
\bottomrule
\end{tabular}
\vspace{-0.7em}
\end{table}

We additionally define five descriptive indicators of potential INT8 inefficiency, labeled R1--R5 in \cref{fig:coverage-risk}: (R1) conversion-bound ($\geq 50$ reformat layers per GMAC), (R2) misaligned GEMM ($<45\%$ of convolution and linear operations aligned to multiples of 32), (R3) unsupported channel-multiplier depthwise operations, (R4) overhead-bound ($<65\%$ GPU share of engine time), and (R5) low INT8 MAC coverage ($<50\%$). These criteria summarize structural characteristics of the compiled engines rather than serve as a predictive model of performance.

\fi

\end{document}